\documentclass[%
 reprint,
nofootinbib,
 amsmath,amssymb,
 aps,
]{revtex4-2}

\usepackage{graphicx}%
\usepackage{dcolumn}%
\usepackage{bm}%

\usepackage{todonotes}  %
\usepackage[most]{tcolorbox}
\usepackage[T1]{fontenc}
\usepackage[english]{babel} %
\usepackage{booktabs}
\usepackage{pifont}
\newcommand{\cmark}{\ding{51}}%
\newcommand{\xmark}{\ding{55}}%
\usepackage{tikz}
\usetikzlibrary{fit}
\usepackage{hyperref}
\newcommand{\hlcirc}[1]{%
\tikz[baseline=(X.base)]\node[draw=red, circle, line width=0.6pt, inner sep=1.2pt] (X) {#1};%
}

\makeatletter
\let\acknowledgmentswithtoc\acknowledgments
\let\endacknowledgmentswithtoc\endacknowledgments
\renewenvironment{acknowledgments}{%
  \begingroup
  \let\addcontentsline\@gobblethree
  \acknowledgmentswithtoc
}{%
  \endacknowledgmentswithtoc
  \endgroup
}
\AtBeginDocument{%
  \renewcommand{\bibsection}{%
    \begingroup
    \let\addcontentsline\@gobblethree
    \expandafter\section\expandafter*\expandafter{\refname}%
    \endgroup
    \@nobreaktrue
  }%
}
\makeatother

\begin{document}

\preprint{APS/123-QED}

\title{Rotational Equivariance in Machine Learning: A Comprehensive Tutorial}%

\author{Peter Lippmann}
 \email{peter.lippmann@iwr.uni-heidelberg.de}
\author{Fred A. Hamprecht}%
\affiliation{%
 Interdisciplinary Center for Scientific Computing (IWR), Heidelberg University,
69120 Heidelberg, Germany
}%

\date{\today}%

\begin{abstract}
Rotational symmetry is one of the most important structural principles in machine learning on 3D data. In applications ranging from physics and materials science to 3D computer vision, predictions should not depend on an arbitrary choice of coordinate frame. Rotational equivariance captures this requirement mathematically by enforcing that a rotation of the input induces a corresponding transformation of the model output. This tutorial provides a comprehensive introduction to rotational equivariance, starting from the physical and geometric intuition behind coordinate independence and building up the necessary machinery from geometric deep learning, group theory, and representation theory. We introduce message passing on Euclidean graphs, group actions and representations, spherical harmonics, Wigner matrices, tensor products, and Clebsch-Gordan decomposition, and explain how these ingredients give rise to modern equivariant architectures. We then survey the principal strategies for incorporating rotational equivariance in deep learning, including group convolutions, internal tensorial representations, and canonicalization-based methods, and discuss their practical strengths and limitations. The tutorial aims to lower the barrier to the subject by connecting the underlying mathematics to practical model design, by unifying ideas that are often expressed in different formal languages, and by helping practitioners choose among competing approaches through a clear discussion of their trade-offs.

\end{abstract}

\maketitle

\tableofcontents

\section{Introduction}
Symmetries play an essential role in the physical sciences. For instance, they restrict the form of admissible laws and dynamics: a symmetry rules out any equation of motion that would change under the corresponding transformation. Symmetries serve as guiding principles for formulating theories, e.g.~through symmetry-invariant Lagrangians in particle physics. More broadly, the laws of physics are formulated \emph{covariantly}, meaning that predictions do not depend on arbitrary choices of coordinates or reference frames. 

Modeling this coordinate independence in machine learning (ML) has become a central theme in geometric deep learning. Consider a molecule represented by atom positions in $\mathbb{R}^3$. The global coordinate system is arbitrary, so scalar observables such as the total energy must be invariant under global rotations and reflections, while vector- and tensor-valued quantities (e.g.~forces, dipole moments, stresses) must transform consistently between coordinate descriptions. However, a naive encoding of geometry by raw coordinates makes rotated or reflected configurations appear entirely different to a standard neural network. As a result, the model must effectively relearn the same physical patterns across many orientations.

A fundamental idea in geometric deep learning is therefore to incorporate symmetry as prior knowledge directly into the architecture by enforcing \emph{invariance} and \emph{equivariance}~\cite{cohen2016group,cohen2019general,bronstein2021geometric,duval2023hitchhiker}. These symmetry-based inductive biases improve generalization and data efficiency: the model need not spend model capacity on discovering symmetries from finite data and guarantees exact symmetries even out of distribution, that is, for regions of the input space not explored during training. Accordingly, rotationally (and often reflection-) equivariant models form the backbone of many state-of-the-art methods across disciplines~\cite{batatia2022mace,liao2023equiformerv2,lou2023crin,du2024new,aykent2025gotennet}. \\

The motivation for equivariant ML models is not only conceptual but practical as well. For many complex systems we lack closed-form equations, or solving the known underlying equations accurately is prohibitively expensive (e.g.~for quantum mechanics, weather, cosmology or high-energy physics). In such settings, data-driven surrogate models are essential for prediction and for comparing measurements to simulations~\cite{lam2023learning,spinner2025lorentz,remme2025stable}. When data-acquisition is costly, enforcing known symmetries can be decisive: equivariant networks often reach higher accuracy with fewer samples~\cite{geiger2022e3nn,lippmann2024tensor} and provide physically consistent predictions by construction, particularly important for the stability of long simulations~\cite{batzner20223,frank2024euclidean}. 
More broadly, scientific machine learning aims to combine the strengths of learning-based models with established physical principles, rather than treating neural networks as purely data-driven black boxes. 
In this sense, equivariant neural networks can be viewed as practical, data-driven surrogates that preserve core physical structure.\\

This tutorial is organized as follows. We begin by motivating symmetry as the mathematical expression of coordinate independence and introduce geometric deep learning on Euclidean graphs through the language of message passing. We then develop the group-theoretic framework needed to formalize equivariance, including group representations, which are key to describing how different physical quantities transform under symmetry operations. Building on this foundation, we introduce Cartesian tensors, spherical harmonics, Wigner-$D$ matrices, tensor products, and the Clebsch-Gordan decomposition. We then turn to architectural design and discuss the principal strategies for achieving rotational equivariance in deep learning, including group convolutions, internal tensorial representations, and canonicalization-based methods. Finally, we examine the practical limitations of these approaches and discuss the trade-offs between exact built-in equivariance and learned approximate equivariance achieved through data augmentation.\\

This tutorial is complementary to several excellent existing resources. The guide and overview by~\citet{duval2023hitchhiker} offers a broad and pedagogical overview of geometric deep learning for atomistic systems, while the article by~\citet{kondor2025principles} presents a more abstract and representation-theoretic perspective on equivariant networks. The survey by~\citet{han2022geometrically} provides a wider overview of equivariant GNNs across geometric domains. The value of the present tutorial lies in its tight focus: rather than surveying geometric deep learning broadly or treating equivariance in full generality, it concentrates specifically on rotational equivariance in 3D and develops the subject from the ground up. Its aim is to give readers a clear path from coordinate independence and symmetry principles to the concrete mathematical tools, such as $\mathrm{SO}(3)$-representations, spherical harmonics, Wigner matrices, and tensor products, that underlie modern equivariant architectures, while also helping practitioners navigate the different formalisms and design choices used in the literature.

\section{Symmetries and Coordinate Independence}
Intuitively, a symmetry of a physical system is a transformation that leaves the relevant properties of the system unchanged. In machine learning, the same idea appears in various forms. As illustrated above, the central reason why spatial symmetries matter in scientific machine learning is the arbitrariness of coordinate systems. We need coordinates to represent geometric objects \emph{numerically}, yet the choice of reference frame is not physical. Let us illustrate this point with a simple example: for two directions $\mathbf v_1,\mathbf v_2\in\mathbb R^3$, the angle can be computed from coordinates via
\begin{align}
    \cos\angle(\mathbf v_1,\mathbf v_2)=\frac{\mathbf v_1^{\mathrm T} \mathbf v_2}{\|\mathbf v_1\|_2\,\|\mathbf v_2\|_2}, 
    \qquad
    \|\mathbf v_i\|_2=\sqrt{\mathbf v_i^{\mathrm T} \mathbf v_i},
\end{align}
but the resulting value is independent of how the vectors are expressed in coordinates. \\

In some cases, an appropriate coordinate choice can substantially simplify a computation. For example, the volume of an axis-aligned box with side lengths $a,b,c$ can be written as
\begin{align}
    V = \int_{\mathbb R^3} \mathbb{I}_{\mathrm{Box}}(x,y,z)\,\mathrm dx\, \mathrm dy\, \mathrm dz = a\,b\,c,
\end{align}
where $\mathbb{I}_{\mathrm{Box}}$ is the indicator function of the box. In a coordinate system aligned with the box edges, the integral factorizes immediately; in a rotated frame, the same integral becomes more cumbersome even though the result must agree. Many fundamental physical theories, such as Maxwell's formulation of electromagnetism, exploits this redundancy in the coordinate description through suitable \emph{gauge} choices that simplify equations without changing physical predictions.

The requirement that results obtained in different coordinate systems must be consistent is naturally expressed as \emph{equivariance} with respect to Euclidean symmetries. Intuitively, a function or algorithm is equivariant if transforming its input induces a predictable, corresponding transformation of its output. In that way, equivariant models produce predictions that remain consistent under coordinate changes (or, equivalently, under global transformations of the input geometry).

Crucially, physical laws, formulated as coordinate-independent (covariant) equations, can be seen as \emph{equivariant algorithms}.\footnote{
To be exact, the notion of covariance is typically used in the context of physical laws. An equation is said to be covariant if it relates physical quantities that transform consistently under changes of reference frame, so that the form of the equation is preserved. Equivariance, by contrast, is typically used in the context of functions, and means that applying a transformation to the input is equivalent to first applying the function and then applying the corresponding transformation to the output.
} For instance, Newton's second law maps forces to acceleration,
\begin{align}
    \mathbf a(\mathbf F_1,\dots, \mathbf F_n,m)=\frac{1}{m}\sum_{i=1}^n \mathbf F_i,
\end{align}
and the predicted acceleration transforms as expected when the input is rotated by any rotation matrix $R$:
\begin{align*}
    \mathbf a(R \, \mathbf F_1,\dots,R \, \mathbf F_n,m)
    =\frac{1}{m}\sum_i R \, \mathbf F_i \\
    =R \, \Bigl(\frac{1}{m}\sum_i \mathbf F_i\Bigr) 
    =R\,\mathbf a(\mathbf F_1,\dots,\mathbf F_n,m).
\end{align*}

In contrast, standard neural networks operating on raw coordinate arrays are generally \emph{not} equivariant. As a toy example, suppose we concatenate two forces and a mass into an input vector
$x=(\mathbf F_1\Vert \mathbf F_2\Vert m)\in\mathbb R^{7}$
with $\mathbf F_i\in\mathbb R^{3}$ and scalar $m\in\mathbb R$, and predict the acceleration
$\hat{\mathbf a}\in\mathbb R^{3}$ using a single dense layer with weight $W \in \mathbb R^{3 \times 7}$ and bias $\mathbf b \in \mathbb R^3$, followed by a pointwise nonlinearity such as ReLU,
\begin{align*}
    \hat{\mathbf a} =\mathrm{ReLU}(\underbrace{Wx+ \mathbf b}_{=\mathbf z}) 
\end{align*}
with $\mathrm{ReLU}(\mathbf z)]_i = \max(0,z_i)$ for $i=1,2,3$.
An unconstrained matrix $W$ mixes scalar and vector components arbitrarily, as in $w_1F_{1,1}+w_2F_{1,2}+w_3F_{1,3}+w_4F_{2,1}+\cdots+w_7m$, which does not transform like a vector under rotations. Therefore, our predicted acceleration would not be consistent across different choices of reference frames. Furthermore, component-wise nonlinearities typically break equivariance. A simple example can be given in $\mathbb R^2$. Let $P=-I_2$ denote the $180^\circ$ rotation in 2D. Then,
\begin{align*}
    \mathrm{ReLU} \bigl(P(1,1)^{\mathrm T}\bigr)=\mathrm{ReLU}((-1,-1)^{\mathrm T})=(0,0)^{\mathrm T} \\
    \neq (-1,-1)^{\mathrm T} = P\,\mathrm{ReLU}((1,1)^{\mathrm T}),
\end{align*}
showing that $\mathrm{ReLU}$ does not map vectors to vectors in an equivariant way. These basic examples illustrate that equivariance is usually not obtained ``for free'' but often requires careful design of equivariant architectures.\\

\begin{tcolorbox}[
  enhanced,
  boxrule=0.6pt,
  colback=white,
  colframe=black!60,
  arc=2mm,
  left=3mm,right=3mm,top=2mm,bottom=2mm
]
\noindent\textbf{Key point.} Coordinate frames are needed for numerical computation, but results obtained in different coordinate systems must be consistent. This consistency is formalized via the concept of equivariance; physical laws can be viewed as equivariant algorithms.
\end{tcolorbox}

\section{Geometric Deep Learning and Message Passing} \label{sec:geometric_dl}
Data from various domains, such as scans of 3D scenes, molecular structures, detections in
particle collisions, or earth science measurements, can be viewed as a set of nodes positioned in
Euclidean space and equipped with geometric node features. Concretely, a \emph{point cloud} consists
of nodes $\{i\}$ with features $\{f_i\}$, each located at a position $x_i\in\mathbb R^d$. A \emph{Euclidean
graph} is a point cloud paired with edges $(i,j)$, for instance obtained from a point cloud by constructing a $k$-NN
graph or a radius graph.

A widely used deep learning approach for Euclidean graph is the class of \emph{message passing neural networks} (MPNNs), which
iteratively combine and abstract information by sending messages along edges. In a typical layer, below indexed by $k$,
every node $i$ receives messages $m_{ij}$ from each node $j$ in its neighborhood $\mathcal N(i)$ and aggregates them in a
permutation-invariant way to update its node feature:
\begin{align}
    f_i^{(k+1)}
    = \bigoplus_{j\in\mathcal N(i)} m_{ij} = 
    \bigoplus_{j\in\mathcal N(i)}
    h\!\left(f_i^{(k)},\, f_j^{(k)},\, e_{ij},\, x_i-x_j\right).
    \label{eq:mpnn_update}
\end{align}
Here, the message is constructed using a learnable message function (often an MLP) $h$. Besides the node feature vector of the sending node $f_j^{(k)}$, the message function may receive the feature of the receiving node $f_i^{(k)}$ as input, as well as the relative position vector $x_i-x_j$ and optional edge features $e_{ij}$. The $\bigoplus$ is a permutation-invariant aggregation operator such as sum or mean (or sometimes the component-wise max~\cite{qi2017pointnet++}). Using relative positions $x_i-x_j$ ensures translation invariance, since a global shift of all coordinates leaves all relative displacements unchanged. Since the node update given by Eq.~\eqref{eq:mpnn_update} is permutation invariant with respect to the ordering of neighbors, the overall layer applied to each node $i$ is permutation equivariant, i.e.~permuting input nodes permutes the output node
features accordingly.

Stacking multiple message passing layers in a neural network allows information to propagate over longer distances in the
graph and allows to combine node features into higher-level, more abstract representations. 
In the context of molecular data, for example, nodes usually correspond to atoms (located in 3D space) and input features can encode atom types;
message passing layers then combines local geometric information (which atoms are nearby and how are they positioned)
into increasingly abstract chemical representations used to predict the target property. For global
molecular properties, one typically applies a permutation-invariant readout (e.g.~sum, mean or max
pooling) over node features at the end of the network to produce a single prediction per molecule.

MPNNs form a very general family of models. As a special case of the above, MPNNs can also be applied to \emph{fully connected}
graphs, where each node is connected to all others, i.e.\ $\mathcal N(i)=\{1,\dots,N\}$ (including self-loops). From this perspective, the self-attention mechanism at the
core of the Transformer architecture~\cite{vaswani2017attention} can also be viewed as an instance of message passing with a learned,
``content-dependent'' aggregation. Given node features $f_i\in\mathbb R^{d}$, we obtain queries, keys, and
values from a linear embedding
\begin{align}
    q_i = W^Q f_i,\qquad k_j = W^K f_j,\qquad v_j = W^V f_j,
\end{align}
and attention weights
\begin{align}
    \alpha_{ij} = \mathrm{softmax}_j \bigl(q_i^{\mathrm T} k_j/\sqrt{d_k}\bigr)  :=\frac{\exp\bigl(q_i^{\mathrm T} k_j/\sqrt{d_k}\bigr)}{\sum_{\ell=1}^{N}\exp\bigl(q_i^{\mathrm T} k_\ell/\sqrt{d_k}\bigr)},
\end{align}
where $W^Q, W^K, W^V$ are learnable weight matrices and $d_k$ is the dimension of the key and query vectors.
The updated representation is then obtained by a normalized, weighted sum aggregation of the value vectors $v_j$,
\begin{align}
    f_i^{(k+1)} \;=\; \sum_{j=1}^{N} \alpha_{ij}\, v_j,
\end{align}
In a Transformer block, this self-attention operation is typically followed by an MLP, a residual connection, and a normalization layer. Thus, Transformers can be viewed as MPNNs on fully connected graphs with attention-based message weights, illustrating the versatility of the message passing framework.

\section{Groups and Spatial Symmetries} \label{sec:groups}
A symmetry of a physical system is a transformation that leaves certain properties of the system
unchanged. For example, in the absence of external fields, globally rotating a molecule changes its
orientation but not its energy. In mathematical terms, the set of symmetry transformations is typically
described as a \emph{group}, i.e.~a collection of transformations that can be composed and inverted.

Formally, a group is defined as a set $G$ equipped with a binary operation $\cdot:G\times G\to G$ that describes the composition of transformations and satisfies the following four axioms:
\begin{itemize}
    \item \textbf{Closure:} For all $a,b\in G$, the composition $a\cdot b$ is again an element of $G$.
    \item \textbf{Associativity:} For all $a,b,c\in G$, $(a\cdot b)\cdot c = a\cdot (b\cdot c)$.
    \item \textbf{Identity:} There exists an element $e\in G$ such that $e\cdot a = a\cdot e = a$ for all $a\in G$.
    \item \textbf{Inverse:} For every $a\in G$ there exists $a^{-1}\in G$ such that
    $a\cdot a^{-1} = a^{-1}\cdot a = e$.
\end{itemize}

In this tutorial we focus mostly on the Euclidean symmetries: translations, rotations, and reflections. The
orthogonal group $\mathrm O(d)$ combines rotations and reflections, while the special orthogonal subgroup $\mathrm{SO}(d) \subset \mathrm O(d)$ includes only (orientation-preserving) rotations:
\begin{align}
    \mathrm O(d) &:= \{R\in\mathbb R^{d\times d}\mid R^{\mathrm T} R=I\}, \\
    \mathrm{SO}(d) &:= \{R\in O(d)\mid \det(R)=1\}.
\end{align}
The condition $R^{\mathrm T} R=I$ is equivalent to preserving inner products, i.e.\ $(R \, u)^{\mathrm T}(R \, v)=u^{\mathrm T} v$ for all
$u,v\in\mathbb R^d$, and $\det(R)=-1$ corresponds to orientation-reversing transformations
(reflections). Translations $t \in \mathbb R^d$ (paired with addition) form a non-compact group, acting on points by $x\mapsto x+t$.
Combining rotations with translations yields the special Euclidean group of rigid motions $\mathrm{SE}(d)$. Concretely, we may
represent a rigid motion by a pair $(R,t)$ with $R\in \mathrm{SO}(d)$ and $t\in\mathbb R^d$, acting on points in $d$-dimensional space via
\begin{align}
    x \;\mapsto\; R \, x+t.
\end{align}
The set of all such pairs becomes a group under composition, with group product
\begin{align}
    (R_1,t_1)\cdot (R_2,t_2) \;:=\; (R_1R_2,\; t_1 + R_1 t_2),
    \label{eq:sed_group_law}
\end{align}
identity $(I,0)$, and inverse $(R,t)^{-1}=(R^{\mathrm T},\,-R^{\mathrm T} t)$. The Euclidean group $\mathrm E(d)$ also includes reflections, i.e.~$R\in \mathrm O(d)$.

\section{Group Representations and Equivariance} \label{sec:group_reps}
As we have seen before, physical equations are formulated in a \emph{covariant} (coordinate-independent)
way, that is, they retain the same form under a change of coordinates. Interpreting a physical law as a
map or algorithm from inputs to outputs, this covariance is precisely the requirement of \emph{equivariance}. Namely, the left-
and right-hand side of an equation must transform consistently under the same coordinate
transformation. Group representations provide the mathematical framework to formalize how different geometric objects 
behave under (symmetry) transformations.

As a basic example let us consider a velocity vector $\mathbf v\in\mathbb R^3$. It is a directed geometric quantity which under a rotation $R\in \mathrm{SO}(3)$ transforms
as a vector,
\begin{align}
    \mathbf v \;\mapsto\; \mathbf v' = R \, \mathbf v, \quad \text{i.e.}\quad v_i' = R_{ij} v_j.
\end{align}
Throughout this tutorial, we consistently use Einstein summation convention, i.e.~the summation over indices that are repeated twice is implied. While the components of the velocity vector change, its length remains invariant:
\begin{align}
    \|\mathbf v'\|_2^2
    = (R \, \mathbf v)^{\mathrm T}(R \, \mathbf v)
    = \mathbf v^{\mathrm T} R^{\mathrm T} R\,\mathbf v
    = \mathbf v^{\mathrm T} \mathbf v
    = \|\mathbf v\|_2^2,
\end{align}
using $R^{\mathrm T} R=I$ for orthogonal matrices. Beyond vectors, some geometric quantities transform as higher-order
tensors. For instance, the inertia tensor $\mathcal I \in\mathbb R^{3\times3}$ of a finite set of masses $\{m_k\}$  located at positions $\{\mathbf x_k \in \mathbb R^3\}$ is defined as
\begin{align}
    \mathcal I_{ij} := \sum_{k} m_k \bigl(\left \| \mathbf x_k \right \|_2^2 \delta_{ij} - x_{k,i} x_{k,j}\bigr),
\end{align}
and transforms under $R\in \mathrm{SO}(3)$ as
\begin{align}
    \mathcal I \;\mapsto\; \mathcal I' = R \, \mathcal I R^{\mathrm T},
    \qquad\text{i.e.}\qquad
    \mathcal I'_{ij} = R_{ik} \mathcal I_{kl} R_{jl},
\end{align}
which follows directly from the transformation of positions $\mathbf x_k\mapsto R \, \mathbf x_k$. \\

\begin{tcolorbox}[
  enhanced,
  boxrule=0.6pt,
  colback=white,
  colframe=black!60,
  arc=2mm,
  left=3mm,right=3mm,top=2mm,bottom=2mm
]
\noindent\textbf{Key point.} These examples
illustrate that different geometric objects live in different vector spaces and transform differently
under the same symmetry transformation: scalars are invariant, vectors transform linearly, and higher-order Cartesian tensors
transform by contracting each index with the rotation.
\end{tcolorbox}

\paragraph{Group representation.} Given a group $G$ and a vector space $V$, a \textit{group representation} $\rho$ is a homomorphism from $G$ to the invertible matrices $\mathrm{GL}(V)$, i.e.~a mapping that fulfills
\begin{align} \label{eq:rep}
    \rho(g_1 g_2) = \rho(g_1) \rho(g_2) , \quad \forall g_1 , g_2 \in G ,
\end{align}
where $g_1 g_2$ is the group product and $\rho(g_1) \rho(g_2)$ the standard matrix product. The representation specifies how elements of the group act on vectors $v \in V$, i.e.~in components $(\rho(g) v)_i = \rho(g)_{ij} v_j$. The dimension of the representation $\rho$ is defined to be the dimension of $V$. Condition~(\ref{eq:rep}) implies that $\rho(g^{-1}) = (\rho(g))^{-1}$, see e.g.~\cite{jeevanjee2011introduction} for details.

\paragraph{Cartesian tensor representations.}
Starting from the defining action of $R\in \mathrm O(d)$ on $v\in\mathbb R^d$, $(Rv)_i=R_{ij}v_j$,
one can obtain higher-order Cartesian tensor representations by acting on each index. For an order-$n$
tensor $T_{i_1\ldots i_n}$, we have
\begin{align}
    T'_{i_1\ldots i_n}
    \;=\;
    R_{i_1 j_1}\cdots R_{i_n j_n}\, T_{j_1\ldots j_n}.
    \label{eq:tensor_rep}
\end{align}
For instance, the outer product of two vectors $T_{ij} = (vw^\mathrm T)_{ij} =v_i w_j$ transforms as
$T'_{ij}=(R \, v)_i(R \, w)_j=R_{ik}R_{jl}T_{kl}$. One can easily verify that
Eq.~\eqref{eq:tensor_rep} indeed satisfies the representation property Eq.~\eqref{eq:rep}. To avoid cumbersome indexing, let us illustrate the proof just for $T_{ij}$. Given $R_1,R_2\in \mathrm O(d)$, in components we have
\begin{align*}
    [\rho(R_1 &R_2)T]_{ij}
    = (R_1 R_2)_{ik} (R_1 R_2)_{jl} T_{kl} \\ &= R_{1,im} R_{1,jn} \underbrace{R_{2,mk} R_{2,nl} T_{kl}}_{= [\rho(R_2)T]_{mn}} = [\rho(R_1) \rho(R_2) T]_{ij}.
\end{align*}
where $\rho(R)$ denotes the Cartesian tensor representation of $R$ acting on $T \in \mathbb R^{d\times d}$. Consequently, order-$n$ (or ``rank-$n$'')
Cartesian tensors form a $d^n$-dimensional representation space of $\mathrm O(d)$.

\begin{figure}[t]
\centering
\includegraphics[width=0.8\linewidth]{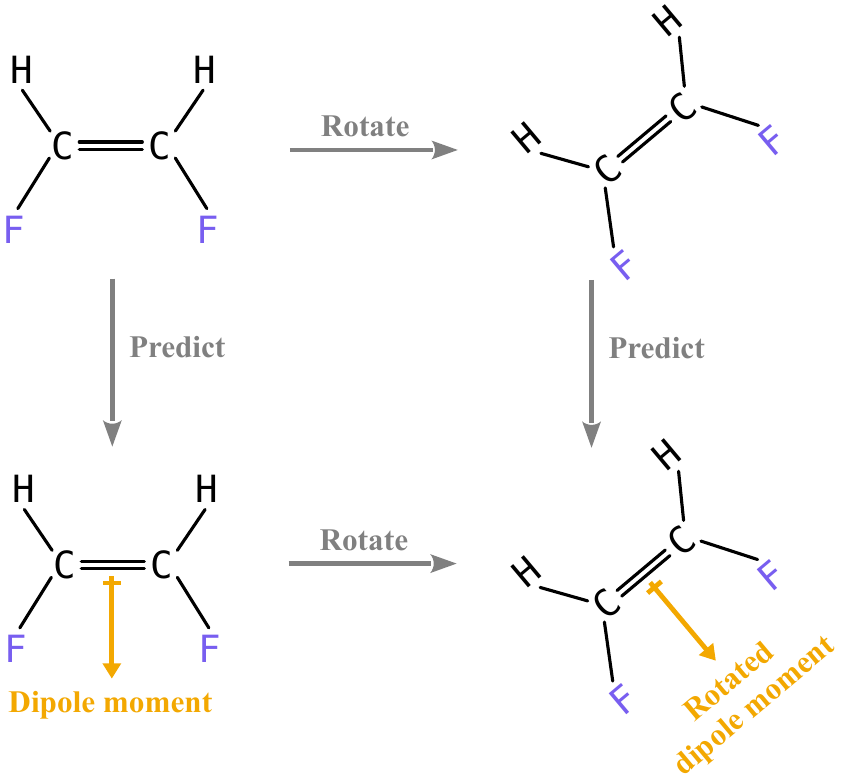}
\caption{\textbf{Equivariant prediction of a molecular dipole moment.} Applying a global
transformation to the molecular geometry before predicting the dipole moment yields the same result as first predicting the dipole moment and then applying the same transformation to the predicted vector.}
\label{fig:equi_dipole}
\end{figure}

\paragraph{Equivariance.} Using the notion of group representations, we can now formalize the concept of equivariance. Let $G$ be a group and $V, W$ two vector spaces equipped with group representations $\rho_{\mathrm{in}}$ and $\rho_{\mathrm{out}}$ respectively. A function $\varphi: V \to W$ is said to be \textit{equivariant} under the group $G$ if the following holds:
\begin{align*}
    \rho_{\mathrm{out}}(g)\varphi(x)=\varphi(\rho_{\mathrm{in}}(g)x), \quad \forall g \in G, \, \forall x \in V
\end{align*}
where the input to $\varphi$ transforms under the representation $\rho_{\mathrm{in}}: V \to \mathrm{GL}(V)$ and its output under the representation $\rho_{\mathrm{out}} : W \to \mathrm{GL}(W)$. If $\rho_{\mathrm{out}}(g) = I$ (identity) for all $g\in G$, the function $\varphi$ is said to be \textit{invariant.}  
Intuitively, equivariance means that $\varphi$ \emph{commutes} with the group action, i.e.~transforming the input and
then applying $\varphi$ is equivalent to first applying~$\varphi$ and then transforming the output, as illustrated in Fig.~\ref{fig:equi_dipole}. 

As a direct consequence of the definition, an equivariant function $\varphi$ is symmetry preserving, i.e.~if the input $x$ has a certain symmetry, e.g.~invariance under a subgroup $H\subset G$, then the output $\varphi(x)$ will have the same symmetry, since for all $h\in H$ we have 
\begin{align}
\varphi(x) = \varphi(\rho_{\mathrm{in}}(h)x) = \rho_{\mathrm{out}}(h)\varphi(x).
\end{align}
Therefore, $\varphi(x)$ is also invariant under $H$.

\paragraph{Example: the cross product is \texorpdfstring{$\boldsymbol{\mathrm{O}(3)}$}{O(3)}-equivariant.}
A well-known yet non-trivial example of an $\mathrm{O}(3)$-equivariant map is the cross product (or vector product) $\mathbf u \times \mathbf v$. It is an insightful exercise to show that indeed, 
\begin{align}
    (R \, \mathbf u)\times(R \, \mathbf v) \;=\; R \, (\mathbf  u\times \mathbf  v)
\end{align}
for any $R \in \mathrm{SO}(3) \quad \text{ and } \quad \mathbf u, \mathbf v \in \mathbb R^3$.
We start by writing the cross product in components using the totally antisymmetric Levi-Civita symbol $\epsilon_{ijk}$, 
\begin{align}
    (\mathbf u \times \mathbf v)_i \;=\; \epsilon_{ijk}\,u_j v_k,
\end{align}
with $\epsilon_{ijk}=1$ if $(i,j,k)$ is an even permutation of $(1,2,3)$, $\epsilon_{ijk}=-1$ if it is an odd permutation, and $\epsilon_{ijk}=0$ otherwise.
Now, inserting $u_j \to R_{jl}u_l$ and $v_k \to R_{km}v_m$, we obtain
\begin{align}
    \bigl((R \,  \mathbf u)\times(R \, \mathbf v)\bigr)_i
    &= \epsilon_{ijk} (R \, \mathbf u)_j (R \, \mathbf v)_k
     = \epsilon_{ijk} R_{jl}R_{km}\,u_l v_m. \label{eq:cross_prod_2R}
\end{align}
To eliminate one of the matrices $R$ on the right-hand side, we use the following general identity that connects the Levi-Civita symbol to the matrix determinant~\cite{lipschutz2009linear}:
\begin{align}
    \epsilon_{ijk} R_{jl} R_{km} R_{in} \;=\; \det(R)\,\epsilon_{lmn}.
    \label{eq:eps_det_identity}
\end{align}   
Contracting both sides with $R_{no}^{\mathrm T}$ and using $R^{\mathrm T} R=I$ (i.e.~$R_{in} R_{no}^{\mathrm T}  = \delta_{io}$) yields
\begin{align}
    \epsilon_{ojk} R_{jl} R_{km} \;=\; \det(R)\,\epsilon_{lmn} R_{on}.
    \label{eq:eps_det_identity2}
\end{align}
Renaming the free index $o \to i$, inserting Eq.~\eqref{eq:eps_det_identity2} into Eq.~\eqref{eq:cross_prod_2R} and using that $\det(R) = 1$ for rotation matrices, we can confirm the equivariance by
\begin{align}
    \bigl((R \, \mathbf  u)\times(R \, \mathbf v)\bigr)_i
    &= \det(R) \epsilon_{lmn} R_{in} u_l v_m = R_{in} \epsilon_{nlm} u_l v_m \nonumber \\&= \bigl(R \, (\mathbf u \times \mathbf v)\bigr)_i
\end{align}
For reflections (orientation-reversing transformations) with
$\det(R)=-1$, the cross product picks up a minus sign in the last step of the derivation. For example, for parity
$P=-I\in \mathrm O(3) \setminus \mathrm{SO}(3)$,
\begin{align}
    (P \, \mathbf u)\times(P \, \mathbf v) \;=\; \det(P)\,P \, (\mathbf u\times \mathbf v) \;=\; -\,P \,(\mathbf u\times \mathbf v).
\end{align}
Thus $\mathbf u\times \mathbf v$ is an \emph{axial vector} (or pseudovector). Under a general orthogonal transformation
$R\in \mathrm O(3)$ it transforms as
\begin{align}
    v_i' \;=\; \det(R)\,R_{ij}v_j.
    \label{eq:pseudovector_rep}
\end{align}

\paragraph{Cartesian pseudotensor representation.}
As we have seen from the example of pseudovectors, the fact that the orthogonal matrices also include reflections can be used to distinguish the transformation behavior of geometric objects with respect to orientation-reversing transformations (with determinant $-1$). 
The following transformation behavior defines another representation, namely the one of \textit{pseudotensors}: 
\begin{align} \label{eq:pseudo}
    P'_{i_1 ... i_n} = \det(R) R_{i_1, j_1} \ ... \ R_{i_n, j_n}  P_{j_1 ... j_n} .
\end{align}
The proof that the pseudotensor representation forms a group representation (i.e.~satisfies Eq.~\eqref{eq:rep}) works exactly as the one for the tensor representation above, additionally using the fact that the determinant is multiplicative: $\det (AB)=\det (A) \det (B)$. \\

\begin{tcolorbox}[
  enhanced,
  boxrule=0.6pt,
  colback=white,
  colframe=black!60,
  arc=2mm,
  left=3mm,right=3mm,top=2mm,bottom=2mm
]
\noindent\textbf{Key point.} Group representations specify how symmetry transformations act on different geometric objects living in separate vector spaces. They form the basis for transforming features between different (local) reference frames in our tensorial message passing approach. 
\end{tcolorbox}

\section{Spherical Harmonics and the Irreducible Representations of \texorpdfstring{$\boldsymbol{\mathrm{SO}(3)}$}{SO(3)}}
\label{sec:decomposition}

In this section, we demonstrate how the Cartesian tensor representations of Eq.~\eqref{eq:tensor_rep}
can be decomposed into irreducible subrepresentations of $\mathrm{SO}(3)$, realized by the so-called\linebreak Wigner-$D$ matrices. We further illustrate
that spherical harmonics, which form a complete orthonormal basis for (square-integrable) functions on the
sphere $S^2$, transform precisely under these irreducible representation of $\mathrm{SO}(3)$. Based on that, we introduce the notion of \emph{spherical tensors}.

\paragraph{Irreducible subrepresentations.} Formally, a subspace $W\subset V$ is called
\emph{$G$-invariant}, with respect to a representation $\rho:G\to \mathrm{GL}(V)$, if
$\rho(g)w\in W$ for all $g\in G$ and $w\in W$. In this case, the restriction $\rho|_W: G \to \mathrm{GL}(W), \, g \mapsto \rho(g)$ defines a
\emph{subrepresentation}. A representation is called \emph{irreducible} if it has no non-trivial
$G$-invariant subspaces, i.e.~if the only invariant subspaces are $\{0\}$ and $V$ itself~\cite{jeevanjee2011introduction}.

\paragraph{Example: rank-2 Cartesian tensors are reducible.}
Consider the rank-2 Cartesian tensor representation of $\mathrm{SO}(3)$ acting on $T\in\mathbb R^{3\times 3}$ by
\begin{align}
    T \;\mapsto\; T' = R \, T R^{\mathrm T},
    \qquad\text{i.e.}\qquad
    T'_{ij}=R_{ik}R_{jl}T_{kl}.
\end{align}
This representation is reducible because it contains the one-dimensional invariant subspace spanned by the identity matrix 
$\mathrm{span}\{I\}$. Indeed, for $T=aI \in \mathrm{span}\{I\}$ we have
$T' = aRIR^{\mathrm T}=aI \in \mathrm{span}\{I\}$. This component is often called the \emph{trace part} because every rank-2 tensor admits
the canonical decomposition
\begin{align}
    T \label{eq:trace_decomposition}
    \;=\;
    \underbrace{\Bigl(T-\tfrac{\mathrm{tr}(T)}{3}\,I\Bigr)}_{\text{traceless part}}
    \;+\;
    \underbrace{\tfrac{\mathrm{tr}(T)}{3}\,I}_{\text{(isotropic) trace part}},
\end{align}
and the coefficient of the isotropic basis tensor $I$ is fully determined by $\mathrm{tr}(T)$. We will see how to further decompose a rank-2 Cartesian tensor in Sec.~\ref{sec:tensor_prod} below.

\paragraph{Spherical Harmonics.}
Spherical harmonics are scalar functions on the sphere
$S^2=\{\hat{\mathbf r}\in\mathbb R^3:\|\hat{\mathbf r}\|_2=1\}$, indexed by an integer \emph{degree}
$l\in\{0,1,2,\dots\}$ and an \emph{order} $m\in\{-l,\dots,l\}$. One may work either with the
\emph{complex} spherical harmonics $Y^{\mathbb C}_{lm}:S^2\to\mathbb C$ or with an equivalent
\emph{real} spherical harmonics basis $Y^{\mathbb R}_{lm}:S^2\to\mathbb R$. In either case, the
set of functions $\{Y_{lm}\}_{l,m}$ forms a complete orthonormal basis of $L^2(S^2)$, that is, any square-integrable
real-valued function $f:S^2\to\mathbb R$ can be expanded into spherical harmonics as
\begin{align}
    f(\hat{\mathbf r}) \;=\; \sum_{l=0}^{\infty}\sum_{m=-l}^{l} c_{lm}\,Y^{\mathbb C}_{lm}(\hat{\mathbf r}),
    \label{eq:sh_expansion}
\end{align}
with coefficients $c_{lm}\in\mathbb C$. Importantly, for $f$ to be real-valued, the coefficients in the complex basis are not independent but satisfy the following constraint\footnote{This assumes the usual Condon-Shortley phase convention
$Y^{\mathbb C}_{l,-m} = (-1)^m \overline{Y^{\mathbb C}_{lm}}$~\cite{arfken1967mathematical}.}
\begin{align}
    c_{l,-m} \;=\; (-1)^m\,\overline{c_{lm}}
    \qquad (m=1,\dots,l),
    \label{eq:sh_real_constraint}
\end{align}
where $\overline{c_{lm}}$ denotes the complex conjugate.

For given $l$, we may change between the complex spherical harmonics $Y^{\mathbb C}$ and the real ones $Y^{\mathbb R}$ using the following conversion~\cite{edmonds1996angular}: 
\begin{align}
    Y^{\mathbb R}_{lm}(\hat{\mathbf r}) &:= \sqrt{2} \, (-1)^{\vert m \vert} \, \mathrm{Im}\bigl(Y^{\mathbb C}_{l \vert m \vert}(\hat{\mathbf r})\bigr), \qquad m < 0, \\
    Y^{\mathbb R}_{l0}(\hat{\mathbf r}) &:= Y^{\mathbb C}_{l0}(\hat{\mathbf r}), \qquad  \hspace{3.1cm} \, m = 0,  \\
    Y^{\mathbb R}_{lm}(\hat{\mathbf r}) &:= \sqrt{2} \,(-1)^m \, \mathrm{Re}\bigl(Y^{\mathbb C}_{lm}(\hat{\mathbf r})\bigr), \qquad  \hspace{0.3cm} m > 0. 
\end{align}
The general form of the complex spherical harmonics of degree $l$ is given by
\begin{align}
    Y^{\mathbb C}_{lm}(\theta,\phi) &\propto P_l^m(\cos\theta)\,\mathrm e^{\mathrm i m\phi},
\end{align}
where $P_l^m$ are the associated Legendre polynomials. In spherical coordinates $(\theta,\phi)$, $\phi\in[0,2\pi)$ denotes the azimuthal angle in the $xy$-plane from the positive $x$-axis and $\theta\in[0,\pi]$ is the polar angle from the positive $z$-axis.
Since, in the complex basis, the azimuthal dependence of $Y^{\mathbb C}_{lm}$ is given by $\mathrm e^{\mathrm i m\phi}$, in the real basis, $Y^{\mathbb R}_{lm}$ correspond to $\cos(m\phi)$- and $\sin(m\phi)$-type angular dependences, respectively.

\begin{figure*}[t]
    \centering
    \includegraphics[width=.85\linewidth]{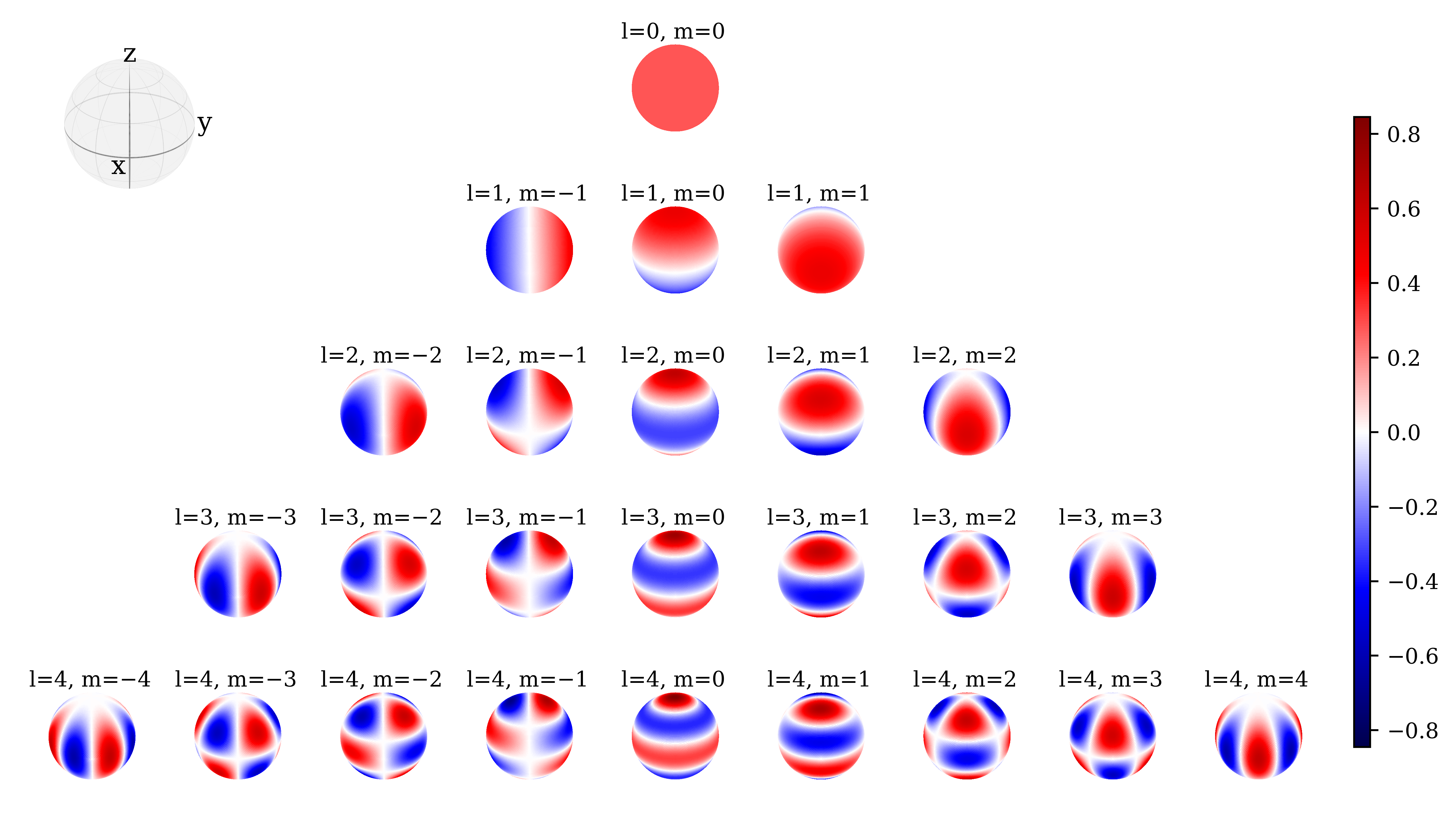}
    \caption{Visualization of the lowest-degree real spherical harmonics $Y^{\mathbb R}_{lm}$ as functions on the sphere. For $m=0$ the harmonics are axially symmetric around the $z$-axis, while for $m\neq 0$ they have a characteristic $\cos(m\phi)$- or $\sin(m\phi)$-type angular dependence in the $xy$-plane. The degree $l$ determines the number of nodal lines (circles where the function changes sign).}
    \label{fig:spherical_harmonics}
\end{figure*}

Up to normalization, the lowest-degree complex harmonics are given by 
\begin{align*}
    Y^{\mathbb C}_{00}(\theta,\phi) &\propto 1, \qquad 
    Y^{\mathbb C}_{10}(\theta,\phi) \propto \cos\theta, \\
    &Y^{\mathbb C}_{1,\pm 1}(\theta,\phi) \propto \sin\theta\,\mathrm e^{\pm \mathrm i\phi},
\end{align*}
and the corresponding real harmonics (for $l=1$) are
\begin{align}
    \label{eq:real_sh_l1}
    Y^{\mathbb R}_{1,-1}(\theta,\phi) &\propto \sin\theta\sin\phi,\qquad
    Y^{\mathbb R}_{10}(\theta,\phi) \propto \cos\theta, \\
    &Y^{\mathbb R}_{11}(\theta,\phi) \propto \sin\theta\cos\phi.
\end{align}
In machine learning one typically uses the real basis so that
all features and activations remain real-valued. Figure~\ref{fig:spherical_harmonics} shows a visualization of the lowest-degree real spherical harmonics as functions on the sphere.

For us, the key property is that, under any rotation $R\in \mathrm{SO}(3)$, spherical harmonics of a given degree $l$ mix \emph{only among themselves}. Concretely, stacking all orders into a vector-valued map
\begin{align}
    Y^{(l)}:S^2\to \mathbb R^{2l+1},\qquad
    Y^{(l)}(\hat{\mathbf r}):=\bigl(Y^{\mathbb R}_{l,-l}(\hat{\mathbf r}),\dots,Y^{\mathbb R}_{l,l}(\hat{\mathbf r})\bigr)^{\mathrm T}
\end{align}
(here in the real basis) one obtains a linear transformation law\footnote{The transformation here should be understood in the following way: we fix the coordinate system, and with it the spherical harmonics functions, and simply evaluate these functions at the (physically) rotated unit vector
$R \, \hat{\mathbf r}$.
}
\begin{align}
    Y^{(l)}(R \, \hat{\mathbf r}) \;=\; D^{(l)}(R)\,Y^{(l)}(\hat{\mathbf r}),
    \qquad R\in SO(3),
    \label{eq:sh_mix}
\end{align}
where $D^{(l)}(R)\in\mathbb R^{(2l+1)\times(2l+1)}$ is the so-called \emph{Wigner-$D$ matrix} in that basis. 
We will discuss in Sec.~\ref{sec:compute_wigner_d} how to compute $D^{(l)}(R)$ efficiently.

Let us explicitly study the transformation behavior of the spherical harmonics of the lowest degrees. For $l=0$, we have $Y_{00}\propto 1$,
hence $Y^{(0)}(R \, \hat{\mathbf r})=Y^{(0)}(\hat{\mathbf r})$, corresponding to the trivial (invariant) representation. For $l=1$, we recognize that the real harmonics are (up to permutations) equal standard spherical coordinates with radius $r=1$. For that compare, Eq.~\eqref{eq:real_sh_l1} with
\begin{align}
    x = r \sin\theta\cos\phi,\qquad
    y = r \sin\theta\sin\phi,\qquad
    z = r \cos\theta.
\end{align}
Consequently, one can choose a suitable real basis such that
\begin{align}
    Y^{(1)}(\hat{\mathbf r}) \;\propto\; \hat{\mathbf r}.
\end{align}
In this basis the Wigner-$D$ matrices for $l=1$ coincide with the standard rotation matrices,
\begin{align}
    D^{(1)}(R) = R.
    \label{eq:l1_is_vector}
\end{align}
This convention (used e.g.~in the equivariant ML library \texttt{e3nn}~\cite{geiger2022e3nn}) emphasizes that degree-$1$ harmonics transform like ordinary vectors.

\paragraph{Irreducible representations and spherical tensors.}
A standard classification result in representation theory tells us that the finite-dimensional irreducible representations (irreps)
of $\mathrm{SO}(3)$ are unique, can be indexed by $l\in\{0,1,2,\dots\}$ and have dimension $2l+1$~\cite{fulton2013representation}.
Since for each $l$ the functions $\{Y_{lm}\}_{m=-l}^{l}$ span a $(2l+1)$-dimensional subspace that is
invariant under rotations, the induced actions of $R\in\mathrm{SO}(3)$ on these subspaces are (up to a change of basis) exactly the unique irreducible representation of of $R\in\mathrm{SO}(3)$.
This motivates the definition of a \emph{spherical tensor} of degree $l$ as a vector $x\in\mathbb C^{2l+1}$ (or in a real
basis $x\in\mathbb R^{2l+1}$) that transforms according to
\begin{align}
    (D^{(l)}(R)x^{(l)})_m \;=\; \sum_{m'=-l}^{l} D^{(l)}_{mm'}(R)\,x^{(l)}_{m'}.
    \label{eq:spherical_tensor_rep}
\end{align}

Further, we can distinguish between even and odd parity spherical tensors, which transform under the same $\mathrm{SO}(3)$ Wigner-$D$ matrices but differ by a sign under reflections: We use that any orientation-reversing transformation $R \in \mathrm O(3)\setminus \mathrm{SO}(3)$ can be written as
$R = P R'$ with $P=-I$ and $R' := PR \in \mathrm{SO}(3)$ to differentiate between an even-parity spherical tensor $x^{(l)}_{+}$ and an odd-parity spherical tensor $x^{(l)}_{-}$, which transform as
\begin{align}
    x^{(l)}_{+} \;\mapsto\; D^{(l)}(R')\,x^{(l)}_{+},
     \qquad
    x^{(l)}_{-} \;\mapsto\; -\,D^{(l)}(R')\,x^{(l)}_{-}.
    \label{eq:parity_trafo}
\end{align}
for $R=PR'\in \mathrm O(3)\setminus \mathrm{SO}(3)$.

\begin{tcolorbox}[
  enhanced,
  boxrule=0.6pt,
  colback=white,
  colframe=black!60,
  arc=2mm,
  left=3mm,right=3mm,top=2mm,bottom=2mm
]
\noindent\textbf{Key point.} Cartesian tensors, Eq.~(\ref{eq:tensor_rep}), transform under a reducible representation, and can therefore be decomposed into spherical tensors, Eq.~\eqref{eq:spherical_tensor_rep}, which transform via Wigner-$D$ matrices, i.e.~under the irreducible representations of $\mathrm{SO}(3)$. In Sec.~\ref{sec:tensor_prod}, we see explicitly how to perform this decomposition. 
\end{tcolorbox}

\subsection{Computing Wigner-\texorpdfstring{$\boldsymbol D$}{D} matrices} \label{sec:compute_wigner_d}

A fast and practical method for computing $D^{(l)}(R)$ for a given $R\in \mathrm{SO}(3)$ is described by~\citet{pinchon2007rotation}. Their construction works entirely in the \emph{real} basis and evaluates $D^{(l)}(R)$ in $\mathcal O(l^3)$ time using precomputed,
degree-dependent building blocks.
Since the Wigner-$D$ matrices are tied to the transformation behavior of the spherical harmonics by Eq.~\eqref{eq:sh_mix}, we construct the representation matrices by studying how the spherical harmonics transform under rotations.
Following~\cite{pinchon2007rotation}, we order the real spherical harmonics for fixed $l$ as
\begin{align}
    \bigl(Y^{\mathbb R}_{l0},\,Y^{\mathbb R}_{l1},\,Y^{\mathbb R}_{l,-1},\,Y^{\mathbb R}_{l2},\,Y^{\mathbb R}_{l,-2},\,\dots,\,Y^{\mathbb R}_{ll},\,Y^{\mathbb R}_{l,-l}\bigr)^{\mathrm T}.
    \label{eq:ph_ordering}
\end{align}
As we have seen before, in this convention, $Y^{\mathbb R}_{lm}$ and $Y^{\mathbb R}_{l,-m}$ (for $m>0$) carry the azimuthal
dependence $\cos(m\phi)$ and $\sin(m\phi)$, respectively. A rotation about the $z$-axis by angle $\alpha$
acts as $\phi\mapsto \phi+\alpha$ and therefore mixes each pair $(\cos(m\phi),\sin(m\phi))$ by a planar
rotation of angle $m\alpha$. Consequently, the representation matrix for $R_z(\alpha)$ is block-diagonal:
\begin{align}
    \begin{split}
    X^{(l)}(\alpha)
    \;=\;
    \mathrm{diag}\Bigl(1,\;R(\alpha),\;R(2\alpha),\;\dots,\;R(l\alpha)\Bigr),
    \\ \text{where} \quad
    R(\vartheta)=
    \begin{pmatrix}
        \cos\vartheta & -\sin\vartheta\\
        \sin\vartheta & \cos\vartheta
    \end{pmatrix},
    \end{split}
    \label{eq:Xl_alpha}
\end{align}
which corresponds to Eq.~(12) in~\cite{pinchon2007rotation}.
To leverage that rotations around the $z$-axis take this particularly simple form, we decompose any rotation $R \in \mathrm{SO}(3)$ via a $z$-$y$-$z$ Euler decomposition
\begin{align} \label{eq:zyz_decomposition}
    R \;=\; R_z(\alpha)\,R_y(\beta)\,R_z(\gamma),
\end{align}
where $R_z(\cdot)$ and $R_y(\cdot)$ are rotations around the $z$- and $y$-axes, respectively, and $\alpha,\beta,\gamma$ are the corresponding Euler angles. The Wigner-$D$ matrix for $R$ can then be computed from the homomorphism property of representations (cf. Eq.~\eqref{eq:rep}):
\begin{align}
    D^{(l)}(R)
    \;=\;
    X^{(l)}(\alpha)\,D^{(l)}(R_y(\beta))\,X^{(l)}(\gamma).
    \label{eq:zyz_factor}
\end{align}
The key idea by~\citet{pinchon2007rotation} is to reduce the remaining, more complicated $y$-rotation to a $z$-rotation by conjugation with the (orthogonal)
coordinate swap
\begin{align}
    J:\ (x,y,z)\mapsto(x,z,y),
    \qquad
    J\in \mathrm O(3) \setminus \mathrm{SO}(3),\quad J^2=I.
\end{align}
Based on the transformation rules of Eq.~\eqref{eq:parity_trafo} for orientation-reversing orthogonal transformations, $J$ induces a well-defined linear operator $J^{(l)}\in\mathbb R^{(2l+1)\times(2l+1)}$ on space of degree-$l$ spherical tensors. The induced operator similarly satisfies
\begin{align}
    (J^{(l)})^2=I
    \qquad\Rightarrow\qquad
    (J^{(l)})^{-1}=J^{(l)}.
\end{align}
Using once more the homomorphism property of the Wigner-$D$ matrices, we can write the more complicated $y$-rotation as
\begin{align}
    D^{(l)}(R_y(\beta)) \;=\; J^{(l)}\,X^{(l)}(\beta)\,J^{(l)}.
    \label{eq:Dy_via_J}
\end{align}
The matrices $J^{(l)}$ can be precomputed once per degree $l$. In~\cite{pinchon2007rotation} they are
constructed recursively by propagating the known action of $J$ at $l=1$ (i.e.\ on $(x,y,z)$) to higher
degrees. Combining Eq.~\eqref{eq:zyz_decomposition} and~\eqref{eq:Dy_via_J} yields the efficient Pinchon-Hoggan factorization
\begin{align}
    D^{(l)}(R)
    \;=\;
    X^{(l)}(\alpha)\,J^{(l)}\,X^{(l)}(\beta)\,J^{(l)}\,X^{(l)}(\gamma),
    \label{eq:PH_factorization}
\end{align}
where each $X^{(l)}(\cdot)$ is cheap to evaluate (independent $2\times2$ blocks $R(m\,\cdot)$), and
$J^{(l)}$ is fixed for a given $l$ and can be reused across all rotations. This yields an accurate and
practical $\mathcal O(l^3)$ scheme for computing Wigner-$D$ matrices in the real basis.

\section{Decomposition of Cartesian Tensors and the Tensor Product} \label{sec:tensor_prod}
We have established that Cartesian tensor representations (cf.~Eq.~\eqref{eq:tensor_rep}) are in general
\emph{reducible}, while the irreducible $\mathrm{SO}(3)$ representations are realized by the Wigner-$D$
matrices $D^{(l)}(R)\in\mathbb R^{(2l+1)\times(2l+1)}$. In this section, we connect both representation types and
introduce the tensor product and its irreducible (Clebsch-Gordan) decomposition. The latter is one of
the key building blocks in state-of-the-art $\mathrm{SO}(3)$-equivariant neural networks~\cite{batatia2022mace,musaelian2023learning,liao2024equiformerv2}.

\paragraph{Tensor products of representations.}
Given two tensors $A_{i_1\ldots i_n}$ and $B_{j_1\ldots j_m}$, their tensor (outer) product is defined component-wise by
\begin{align}
    (A\otimes B)_{i_1\ldots i_n\, j_1\ldots j_m} \;:=\; A_{i_1\ldots i_n}\,B_{j_1\ldots j_m}.
\end{align}
If $A$ and $B$ are Cartesian tensors and transform under rotations as in Eq.~\eqref{eq:tensor_rep}, then $A\otimes B$ transforms as
\begin{align}
    (&A\otimes B)'_{i_1\ldots i_n\, j_1\ldots j_m} \nonumber \\
    &=\;
    R_{i_1k_1}\cdots R_{i_nk_n}\,R_{j_1\ell_1}\cdots R_{j_m\ell_m}\,
    (A\otimes B)_{k_1\ldots k_n\,\ell_1\ldots \ell_m},
\end{align}
so the tensor product again forms a tensor representation. The same statement holds if $x^{(l_1)}$ and $y^{(l_2)}$ are spherical tensors, then $x^{(l_1)}\otimes y^{(l_2)}$ transforms under the
\emph{product representation} $D^{(l_1)}\otimes D^{(l_2)}$. While the tensor (outer) product yields an equivariant output tensor by construction, stacking outer products in a neural network would quickly lead to a combinatorial blow-up in the feature dimensionality. This motivates the decomposing into irreps, that is, a set of smaller, more manageable building blocks with well-defined transformation behavior.\\

\paragraph{Decomposition of Cartesian tensors.}
Cartesian tensors of rank $0$ (scalars) and rank $1$ (vectors) are already irreducible under $\mathrm{SO}(3)$. 
The first non-trivial case is a rank-2 Cartesian tensor. To keep track of the dimensions in a tensor decomposition, we introduce the shorthand $\underline{d}$ to denote an irrep of dimension $d$, that is,
$\underline{2l+1}$ refers to the unique $(2l+1)$-dimensional irrep labeled by $l$ (cf.~Sec.~\ref{sec:decomposition}).
Since a rank-2 Cartesian tensor has the same transformation behavior, $T'_{ij}=R_{ik}R_{j\ell}T_{k\ell}$, as the tensor (outer) product of two vectors, we associate $T_{ij}$ with $\underline{3}\otimes \underline{3}$.   
First, we define the symmetric and antisymmetric parts of $T_{ij}$ by
\begin{align}
    S_{ij}=\tfrac12(T_{ij}+T_{ji}),
    \qquad
    A_{ij}=\tfrac12(T_{ij}-T_{ji}).
\end{align}
Both span invariant subspaces (subrepresentations), since for example
\begin{align}
    S'_{ij}
    =\tfrac12(T'_{ij}+T'_{ji})
    =R_{ik}R_{j\ell}\,\tfrac12(T_{k\ell}+T_{\ell k})
    =R_{ik}R_{j\ell}\,S_{k\ell},
\end{align}
and similarly for $A_{ij}$. The antisymmetric subspace is $3$-dimensional and is isomorphic to a
\emph{pseudovector} by use of the Levi-Civita symbol:
\begin{align}
    a_k \;:=\; \tfrac12\,\epsilon_{kij}\,A_{ij},
    \qquad
    A_{ij}=\epsilon_{ijk}\,a_k.
\end{align}
In fact, it is easy to confirm\footnote{In the case where $T$ is given by an outer product $T_{ij} = v_i w_j$, plugging in the definition of $A_{ij}$ reveals that $\mathbf a$ is actually proportional to the cross product $\mathbf v \times \mathbf w$.} that in this form $\mathbf a$ transforms as $\mathbf a' =R \, \mathbf a$ for $R\in\mathrm{SO}(3)$ acting on $A_{ij}$ and that $\mathbf a'$ picks up an additional sign under reflections,
cf.~Eq.~\eqref{eq:pseudovector_rep}. Thus, the antisymmetric part realizes the same $\underline{3}$ irrep as an ordinary
vector, but with different parity under $\mathrm{O}(3)$.

The symmetric part further decomposes into trace and traceless parts. As we have seen before
(cf.~Eq.~\eqref{eq:trace_decomposition}), the trace part forms a closed subspace equipped with the trivial (invariant)
representation ($l=0$). The remaining symmetric traceless part
\begin{align}
    \tilde S_{ij} \;:=\; S_{ij}-\delta_{ij}\tfrac{\mathrm{Tr}(S)}{3}
\end{align}
is $5$-dimensional and can be shown to correspond to the $l=2$ irrep. Altogether, in dimension notation,
\begin{align}
    \underline{3}\otimes \underline{3}
    \;=\;
    \underline{1}\oplus \underline{3}\oplus \underline{5}.
    \label{eq:3x3_decomp}
\end{align}

\paragraph{Clebsch-Gordan decomposition and the coupled tensor product.}
More generally, the tensor product of two $\mathrm{SO}(3)$ irreps decomposes as
\begin{align}
    \underline{2l_1+1}\otimes \underline{2l_2+1}
    \;=\;
    \bigoplus_{L=\lvert l_1-l_2\rvert}^{l_1+l_2}\underline{2L+1}.
    \label{eq:CG_dim_rule}
\end{align}
Using Gauss's summation formula, one may easily verify that the dimensions match by showing that
\begin{align}
    \sum_{L=\lvert l_1-l_2\rvert}^{l_1+l_2}(2L+1)=(2l_1+1)(2l_2+1).
\end{align}
As an example, we can use Eq.~\eqref{eq:CG_dim_rule} to examine how a rank-$3$ Cartesian tensor decomposes into irreps. A rank-$3$ Cartesian tensor corresponds to $\underline{3}\otimes\underline{3}\otimes\underline{3}$. Using Eq.~\eqref{eq:3x3_decomp} iteratively yields
\begin{align}
    \underline{3}\otimes\underline{3}\otimes\underline{3}
    \;&=\;
    (\underline{1}\oplus \underline{3}\oplus \underline{5})\otimes \underline{3} \nonumber \\
    \;&=\;
    \underline{1} \otimes \underline{3} \oplus \underline{3} \otimes \underline{3} \oplus \underline{5} \otimes \underline{3} \nonumber \\
    \;&=\;
    \underline{3} \oplus (1\oplus \underline{3} \oplus \underline{5})  \oplus (\underline{3}\oplus \underline{5} \oplus \underline{7}) \nonumber \\
    \;&=\;
    \underline{1} \oplus \underline{3} \oplus \underline{3} \oplus \underline{3} \oplus \underline{5} \oplus \underline{5} \oplus \underline{7}
\end{align}
i.e.\ one scalar, three vector-type components, two $l=2$ components, and one $l=3$ component.

While Eq.~\eqref{eq:CG_dim_rule} provides a decomposition in terms of subspace dimensions, the numerical irreps decomposition of spherical tensors is implemented by so-called Clebsch-Gordan (CG) coefficients,
which define a change of basis from the \emph{product basis} (irrep$_1$ $\otimes$ irrep$_2$) to a \emph{coupled} basis (direct sum of irreps): Let $x^{(l_1)}$ and $y^{(l_2)}$ be spherical tensors transforming as
\begin{align*}
    (D^{(l_1)}(R)x^{(l_1)})_{m_1}=\sum_{m_1'=-l_1}^{l_1} D^{(l_1)}_{m_1m_1'}(R)\,x^{(l_1)}_{m_1'},
    \\
    (D^{(l_2)}(R)y^{(l_2)})_{m_2}=\sum_{m_2'=-l_2}^{l_2}D^{(l_2)}_{m_2m_2'}(R)\,y^{(l_2)}_{m_2'}.
\end{align*}
Their tensor product transforms under the product representation:
\begin{align}
    \bigl((D^{(l_1)}(R)\otimes D^{(l_2)}(R))(x^{(l_1)}\otimes y^{(l_2)})\bigr)_{m_1m_2}
    \nonumber \\=
    \sum_{m_1',m_2'}
    D^{(l_1)}_{m_1m_1'}(R)\,D^{(l_2)}_{m_2m_2'}(R)\,x^{(l_1)}_{m_1'}y^{(l_2)}_{m_2'}.
\end{align}
To extract the $L$-irrep component of the product representation, we define the \emph{Clebsch-Gordan (coupled) product}
\begin{align}
    z^{(L)} \;&=\; x^{(l_1)}\otimes_{\mathrm{CG}} y^{(l_2)},
    \nonumber \\
    z^{(L)}_M
    &:=
    \sum_{m_1,m_2}
    C^{LM}_{l_1m_1,\,l_2m_2}\,x^{(l_1)}_{m_1}y^{(l_2)}_{m_2},
    \label{eq:cg_product_def}
\end{align}
with $|l_1-l_2|\le L\le l_1+l_2$ and $-L\le M\le L$. The famous Clebsch-Gordan coefficients are denoted as $C^{LM}_{l_1m_1,\,l_2m_2}$. In the literature, the Clebsch-Gordan product is often referred to simply as ``tensor product'', which can be confusing since the CG product combines the usual tensor (outer) product with an additional change of basis afterwards. We use the notation $\otimes_{\mathrm{CG}}$ vs.~$\otimes$ to make this distinction explicit. 

The defining property of the CG coefficients is that the resulting coupled tensor
$z^{(L)}=x^{(l_1)}\otimes_{\mathrm{CG}}y^{(l_2)}$ transforms as an $L$-type spherical tensor under $R\in \mathrm{SO}(3)$. This condition is equivalent to imposing equivariance of the CG tensor product. Concretely, equivariance of the CG decomposition requires that, for all $R\in\mathrm{SO}(3)$,
\begin{align}
    z^{(L)}\!\bigl(D^{(l_1)}(R)x^{(l_1)},\,D^{(l_2)}(R)y^{(l_2)}\bigr)
    \nonumber \\ \;=\;
    D^{(L)}(R)\,z^{(L)}\!\bigl(x^{(l_1)},y^{(l_2)}\bigr),
    \label{eq:cg_equiv_req}
\end{align}
i.e.~we equate ``transform then couple'' with ``couple then transform''.
Expanding both sides and using that this equation must hold for any tensors of $x^{(l_1)}, y^{(l_2)}$, we obtain the \emph{three-$D$} equation. In components, we require that for any $R\in\mathrm{SO}(3)$
\begin{align}
    \sum_{m_1=-l_1}^{l_1}\sum_{m_2=-l_2}^{l_2}
    C^{LM}_{l_1m_1,\,l_2m_2}\,
    D^{(l_1)}_{m_1 m_1'}(R)\,D^{(l_2)}_{m_2 m_2'}(R) \nonumber \\
    \;=\;
    \sum_{M'=-L}^{L}
    D^{(L)}_{MM'}(R)\,C^{LM'}_{l_1m_1',\,l_2m_2'}.
    \label{eq:three_D_equation_final}
\end{align}
Together with a normalization and a phase convention, this constraint determines the
coefficients uniquely. 
As discussed in Sec.~\ref{sec:compute_wigner_d}, in the complex spherical-harmonics convention, a $z$-rotation acts diagonally on spherical tensors, 
i.e.~$D^{(l)}_{mm'}(R_z(\alpha))=e^{-im\alpha}\delta_{mm'}$.
Plugging $R=R_z(\alpha)$ into
Eq.~\eqref{eq:three_D_equation_final} yields, for all $\alpha$,
\begin{align}
    e^{-i(m_1'+m_2')\alpha}\,C^{LM}_{l_1m_1',\,l_2m_2'}
    \;=\;
    e^{-iM\alpha}\,C^{LM}_{l_1m_1',\,l_2m_2'}
    \nonumber \\ \Rightarrow \ \ 
    C^{LM}_{l_1m_1',\,l_2m_2'}=0 \ \text{ unless }\ M=m_1'+m_2',
    \label{eq:M_selection_rule_final}
\end{align}
which is one of the standard Clebsch-Gordan selection rules. 
The same sparsity structure holds for the CG coefficients in the real basis.
In practice, CG coefficients are rarely computed by solving Eq.~\eqref{eq:three_D_equation_final} directly. Instead, one
evaluates so-called Wigner $3j$ symbols using fast recursion schemes~\cite{schulten1975exact,johansson2016fast}. \\

\paragraph{Excursion: angular-momentum coupling as intuition for CG rules.}
In fact, there is a deep mathematical connection between the coupling of spherical tensors and angular
momentum coupling in quantum mechanical two-spin systems. In quantum mechanics, spatial rotations act on
state vectors via unitary operators; mathematically, this action is formulated via the special unitary group $\mathrm{SU}(2)$. Intuitively, this means that the complex vectors, on which these unitary matrices act, return to themselves only after a $720^\circ$ rotation. However, there is a one-to-one correspondence between integer-$l$ representations of $\mathrm{SU}(2)$ and the irreps of $\mathrm{SO}(3)$. In this correspondence, $l$ and $m$ can be interpreted as the total angular momentum and its $z$-component, respectively. 
In the coupling of two spins, the $z$-components simply add up, which explains the selection rule $M=m_1+m_2$, while the allowed total angular momenta $L$ ranges from $|l_1-l_2|$ (maximal anti-alignment of the individual angular momenta) to $l_1+l_2$ (maximal alignment).~\cite{edmonds1996angular} \\

\paragraph{Computational scaling of the CG product.}
The coupled product Eq.~\eqref{eq:cg_product_def} is a structured contraction with a rank-3 CG tensor. Treating this
contraction naively, the cost of one CG product for a single $(l_1,l_2,L)$ path is often reported as $\mathcal O(\ell^3)$
when $l_1,l_2,L\sim \ell$.\footnote{This matches the common convention to count the full contraction cost without exploiting
sparsity.}
In many equivariant architectures one carries all irreps up to a maximal degree $L$~\cite{passaro2023reducing,liao2024equiformerv2,fu2025learning}. Then there are $\mathcal O(L^2)$ input
pairs $(l_1,l_2)$, each producing $\mathcal O(L)$ many output spherical tensors. Hence, there are $\mathcal O(L^3)$ input to output paths in total. Combining this
with $\mathcal O(L^3)$ per path yields a naive overall scaling of $\mathcal O(L^6)$ for evaluating all tensor product couplings up to degree $L$.

This cost can become a practical bottleneck: in many equivariant neural networks CG tensor products are executed at essentially every layer. Recent work by~\citet{passaro2023reducing} reduce the cost by converting $\mathrm{SO}(3)$-convolutions to mathematically equivalent $\mathrm{SO}(2)$-convolutions, bringing down the total complexity from $\mathcal O(L^6)$ to $\mathcal O(L^3)$. \\

\begin{tcolorbox}[
  enhanced,
  boxrule=0.6pt,
  colback=white,
  colframe=black!60,
  arc=2mm,
  left=3mm,right=3mm,top=2mm,bottom=2mm
]
\noindent\textbf{Key point.} The outer product of tensors is equivariant, but the output representation is reducible. To keep the representations of feature vectors in neural networks compact, we therefore additionally apply an equivariant Clebsch-Gordan decomposition into irreducible representations. In practice, the decomposition is realized by a change of basis using the Clebsch-Gordan coefficients.
\end{tcolorbox}

\section{Rotational Equivariance in Deep Learning}
With the representation-theoretic background in place, we can now illustrate how equivariance can be achieved in neural networks. In this chapter we focus on \emph{rotational} equivariance, as
this is the most involved part of Euclidean symmetry handling; including reflections as well is mostly straightforward. \emph{Permutation} equivariance is typically ensured by construction in message passing networks, as the aggregation over
neighbors is a set operation, and \emph{translation} invariance is typically achieved by using only relative coordinates $\mathbf x_i- \mathbf x_j$ and distances $\|\mathbf x_i- \mathbf x_j\|_2$ (cf.\ Sec.~\ref{sec:geometric_dl}). Rotational equivariance, in contrast, requires a well-defined transformation behavior of all geometric features under $\mathrm{SO}(3)$ (at least for some approaches). Below we give a brief overview of the most common techniques; for a broader discussion of equivariant message passing we refer the reader to~\cite{duval2023hitchhiker}.\\

\paragraph{Data augmentation (learned, approximate equivariance).}
In general, the simplest way to achieve (approximate) equivariance with respect to a set of transformations
is to augment the training data, i.e.~present the model with randomly transformed samples during training.
This strategy is completely independent of the model and works with any architecture and any group.
However, the equivariance must be learned and is therefore not guaranteed (in particular, it may fail
out-of-distribution), and the learned equivariance is not exact. Still, data augmentation can be very
effective in practice and is, for instance, successfully used in large-scale systems such as
AlphaFold~3~\cite{abramson2024accurate}.\\

\paragraph{Equivariance using tensorial internal representations.}
Historically the first approach to exact rotational \emph{invariance} (as special case of equivariance) was
to feed only invariant geometric inputs, such as distances, angles, or dihedral angles, into a standard (non-equivariant) network~\cite{schutt2018schnet,gasteiger2020directional,li2021rotation,gasteiger2021gemnet,liu2022spherical}. 
While this allows the use of typical deep learning building blocks (linear layers, activations, norm layers, etc.), these approaches are not able to communicate non-scalar geometric information (such as directions) during message passing. To overcome this limitation, several approaches for built-in rotational equivariance have been developed that leverage \emph{tensorial} internal features that transform under specified group representations. Tracking the representation (i.e.~transformation behavior) of internal features enables these models to predict equivariant outputs. Tensorial internal features have also shown to improve performance even when the final target is invariant~\cite{fuchs2020se,brandstetter2021geometric}. In particular, higher-order tensor representations are helpful in tasks where angular information matters, e.g.~for predicting forces in molecules~\cite{zitnick2022spherical,batzner20223}. Exact equivariance in these architectures is ensured by requiring that \emph{every layer} is equivariant.
Two widely used families of such architectures are (i) group-convolutional networks based on the regular representation (see Sec.~\ref{sec:equi_by_group_conv} below), and (ii) tensor field networks based on irreducible representations (see Sec.~\ref{sec:tensor_field_networks_bg}).

\subsection{Equivariance via Group Convolutions}\label{sec:equi_by_group_conv}
For discrete groups, a conceptually clean approach to equivariance is to represent features in the network as \emph{functions on the group}.
In practice, this is achieved by a \emph{lifting} step: an input signal $f:X\to\mathbb{R}^C$ (e.g.\ an
image on $X=\mathbb{Z}^2$) with $C$ feature channels is mapped to a group-indexed feature map
$F:G\to\mathbb{R}^C$ by evaluating the signal in all possible transformed frames (indexed by elements
$g\in G$). 
Subsequent layers then process $F$ via convolutions over $G$. We will see that, in this way,
equivariance is realized as an indexing symmetry of the lifted feature map~\cite{cohen2016group,bekkers2018roto,bekkers2023fast}.%

To explicitly follow the construction, we first introduce yet another group representation: the (left) \emph{regular representation}. Concretely, let $G$ be a finite group that acts on an input domain $X$ (e.g.~planar rotations by $90^\circ$ acting on $X=\mathbb{Z}^2$). Then, we can define the \emph{left action} on signals $f:X\to\mathbb{R}^C$,
\begin{align}
    [L_u f](x) \;:=\; f(u^{-1}\!\cdot x), \qquad u\in G,\ x\in X,
    \label{eq:signal_left_action}
\end{align}
which acts on a function $f$ by evaluating it in transformed coordinates. We again take the viewpoint that the function itself remains unchanged while the argument at which it is evaluated is transformed.
On a group-indexed feature map $F:G\to\mathbb{R}^C$, the analogous left regular action is given by
\begin{align}
    [L_u F](g) \;:=\; F(u^{-1}g), \qquad u,g\in G.
    \label{eq:left_regular_action_cw}
\end{align}
This action defines a group representation, since the homomorphism condition $L_{u_1}L_{u_2}=L_{u_1u_2}$ (compare to Eq.~\eqref{eq:rep}) follows directly from the associativity of the group $G$.

\paragraph{Lifting as a \texorpdfstring{$\boldsymbol G$}{G}-correlation.}
To move from a signal $f:X\to\mathbb{R}^C$ to a group-indexed feature map, one applies a \emph{lifting} layer.
Following \cite{cohen2016group}, the lift is conveniently written as a cross-correlation with a learnable filter $\psi:X\to\mathbb{R}^{C\times C'}$:
\begin{align}
    F(g) \;=\; [f \star \psi](g)
    \;:=\;
    \sum_{x\in X} f(x)\,\psi(g^{-1}\!\cdot x),
    \qquad g\in G.
    \label{eq:lift_correlation}
\end{align}
That is, the filter $\psi(g^{-1}\!\cdot x)$ is applied on the input signal $f$ ``transformed'' by $g$, exactly as in the usual
$\mathbb{Z}^d$ correlation $[f\star\psi](t)=\sum_x f(x)\psi(x-t)$).\footnote{Most papers use the term
``convolution'' also for cross-correlations; the distinction is whether the filter is flipped, which is practically irrelevant for learned filters.}
The output of the lift is now a function on $G$, and transforms under the left regular action.
Indeed, the lifting defined by Eq.~\eqref{eq:lift_correlation} is $G$-equivariant in the sense that transforming the input signal and then lifting is equivalent to lifting and then applying the left regular action on the group index:
\begin{align}
    [(L_u f)\star \psi](g) \;=\; [L_u(f\star \psi)](g), \qquad \forall u,g\in G,
    \label{eq:lift_equivariance_cw}
\end{align}
which can be shown as follows:
\begin{align}
    &[(L_u f)\star \psi](g) \nonumber \\
    &\overset{\phantom{x=u\cdot y}}{=} \sum_{x\in X} (L_u f)(x)\,\psi(g^{-1}\!\cdot x)
     = \sum_{x\in X} f(u^{-1}\!\cdot x)\,\psi(g^{-1}\!\cdot x) \nonumber\\
    &\overset{x=u\cdot y}{=} \sum_{y\in X} f(y)\,\psi(g^{-1}\!\cdot (u\cdot y))
     = \sum_{y\in X} f(y)\,\psi((u^{-1}g)^{-1}\!\cdot y) \nonumber\\
    &\overset{\phantom{x=u\cdot y}}{=} [f\star\psi](u^{-1}g)
     = [L_u(f\star\psi)](g),
\end{align}
where we used that $x\mapsto u\cdot x$ is a bijection on $X$ (e.g.\ a permutation of pixels for discrete
rotations). Thus, the lift turns an input transformation ($[L_u f](x) = f(u^{-1}\!\cdot x)$) into a re-indexing of the group coordinate ($[L_u F](g) = F(u^{-1}g)$).

\paragraph{Group convolution on \texorpdfstring{$\boldsymbol G$}{G}.}
Once features live on $G$, the natural analogue of translation correlation is correlation over the group.
Given $F:G\to\mathbb{R}^C$ and a filter $\kappa:G\to\mathbb{R}^{C\times C'}$, we define a group convolution as
\begin{align}
    [F \star \kappa](g)
    \;:=\;
    \sum_{h\in G} F(h)\,\kappa(g^{-1}h),
    \qquad g\in G.
    \label{eq:group_correlation}
\end{align}
This is the direct generalization of $\mathbb{Z}^d$ convolution: the kernel depends only on the
\emph{relative group element} $g^{-1}h$, i.e.\ on the displacement from $g$ to $h$ measured in group
coordinates.
The group convolution \eqref{eq:group_correlation} is equivariant with respect to the left regular action:
\begin{align}
    [(L_uF)\star \kappa](g) \;=\; [L_u(F\star \kappa)](g), \qquad \forall u,g\in G.
    \label{eq:groupconv_equivariance_cw}
\end{align}
The proof uses again a change of variables, completely analogous to the one above.

In practice, e.g.~for images, one typically retains the spatial index in the lifting and correlates jointly over space and group. A convolution over the group combined with a kernel that is local in $X$ (as in ordinary CNNs) then takes the following form:
\begin{align}
    (F \star \kappa)(x,g)
    \;:=\;
    \sum_{y\in X}\;\sum_{h\in G}
    F(y,h)\,\kappa(x-y,\,g^{-1}h),
    \label{eq:space_group_correlation}
\end{align}
so the kernel depends on the relative spatial displacement $x-y$ and the relative group displacement $g^{-1}h$. This setting is used, for example, in roto-translation equivariant $\mathrm{SE}(2)$ CNNs~\cite{bekkers2018roto}.

\paragraph{Nonlinearities and readout.}
A practical benefit of this construction is that the left regular action \eqref{eq:left_regular_action_cw}
acts by re-indexing the group coordinate and does not mix the channel values in $\mathbb{R}^C$. Therefore,
channel-wise nonlinearities commute with the action: for any scalar nonlinearity $\sigma$, we have $\sigma(L_uF)=L_u\sigma(F)$. This allows standard deep learning layers of alternating
group-correlations and channel-wise nonlinearities. In the end, the group index is typically removed by pooling (e.g.~averaging) over $g$, which is invariant under re-indexing to obtain an invariant or equivariant output (e.g.~a segmentation that is equivariant under global rotations by $90^\circ$~\cite{worrall2018cubenet}). 

Since the feature maps carry an explicit group index, the cost scales with the size $|G|$ of the finite group. In 2D, discretizing $\mathrm{SO}(2)$ is straightforward by sampling angles equidistant on $[0, 2\pi]$, forming the dihedral rotation group. In 3D, however, using a subgroup with exact closure under composition becomes restrictive: beyond the dihedral rotation group (which is only planar), the only finite closed rotation groups correspond to the rotational symmetries of the Platonic solids (tetrahedral, octahedral, icosahedral)~\cite{beisert2015symmetries}. While finer discretizations improve the approximation of the full group, they increase runtime and memory, making the choice of resolution a practical accuracy-efficiency trade-off~\cite{islam2025platonic}.

\subsection{Equivariance via Internal Tensorial Representations} \label{sec:tensor_field_networks_bg}
Probably the most widely used approach for exact $\mathrm{SO}(3)$ equivariance is the method of tensor field networks\footnote{Here, we use the notion of tensor field networks not only to refer to the initial approach presented by~\citet{thomas2018tensor} but more broadly for the class of models described in this section.}, also referred to as e3nn-style networks, which are often based on the Python package \texttt{e3nn}~\cite{geiger2022e3nn}.
In this approach, internal features transform as direct sums of irreducible representations of
$\mathrm{SO}(3)$ and equivariance is enforced by composing equivariant building blocks
\cite{thomas2018tensor,fuchs2020se,geiger2022e3nn,batatia2022mace,liao2023equiformerv2,musaelian2023learning}.
In contrast to group convolution networks, these architectures can no longer treat every internal activation as an individual number: the coordinates of vectors (and higher-order tensors) must be processed jointly to maintain equivariance. As a result, tensor field networks rely on representation-aware building blocks, such as equivariant linear maps, equivariant nonlinearities, and norm layers, as well as specialized tensor operations (the Clebsch-Gordan product introduced in Sec.~\ref{sec:tensor_prod}) to mix the information of spherical tensors of different $l$.

Concretely, node features consist of spherical tensors $f_i^{(l)}\in\mathbb{R}^{2l+1}$ transforming under Wigner-$D$ matrices. Messages are constructed from tensorial node features and tensorial convolution filters coupled through the Clebsch-Gordan product. The convolution filters in a typical $\mathrm{SO}(3)$-convolution consist of a learnable (scalar) radial function, and an angular part given by spherical harmonics~\cite{thomas2018tensor,geiger2022e3nn,fuchs2020se}:
\begin{align}
    \Bigl[f_j^{(l_1)} \otimes_{\mathrm{CG}} \bigl(\mathcal R(r_{ij})\,Y^{(l_2)}(\hat{\mathbf r}_{ij})\bigr)\Bigr]^{(l_3)}_{m_3}&
    \nonumber \\ \;=\;
    \sum_{m_1=-l_1}^{l_1}\sum_{m_2=-l_2}^{l_2}
    C^{l_3 m_3}_{l_1 m_1,\,l_2 m_2}\,
    &f^{(l_1)}_{j,m_1}\,\mathcal R(r_{ij})\,Y^{(l_2)}_{m_2}(\hat{\mathbf r}_{ij}),
    \label{eq:tensor_product_message_bg}
\end{align}
where $r_{ij}=\|\mathbf x_i-\mathbf x_j\|_2$ and $\hat{\mathbf r}_{ij}=(\mathbf x_i- \mathbf x_j)/\|\mathbf x_i- \mathbf x_j\|$, and $\otimes_{\mathrm{CG}}$ denotes the
tensor product followed by a decomposition into spherical tensors (cf.\
Sec.~\ref{sec:tensor_prod}).

\paragraph{Equivariant linear layers, normalization, and nonlinearities.}
Equivariant linear layers respect the representation structure: for each degree $l$, one may freely mix
channels \emph{within} the $l$-type, but not between different degrees. 
Concretely, if a node carries $C_l$ channels of degree-$l$ features, we collect them as
$f^{(l)}\in\mathbb{R}^{C_l\times(2l+1)}$ with components $f^{(l)}_{cm}$, where $c\in\{1,\dots,C_l\}$ indexes the channel and
$m\in\{-l,\dots,l\}$ the $(2l+1)$ irrep components. Then, an equivariant linear map may mix channels
but must act identically on all irrep components, i.e. in components
\begin{align}
    f^{(l)\prime}_{c'm}
    \;=\;
    \sum_{c=1}^{C_l} W^{(l)}_{c'c}\, f^{(l)}_{cm},
    \qquad
    W^{(l)}\in\mathbb{R}^{C_l'\times C_l}.
\end{align}
That is, the same weight matrix $W^{(l)}$ is applied for every $m$. This preserves equivariance because the
$\mathrm{SO}(3)$ action, in contrast, acts on the $m$-index.

For nonlinearities, scalar ($l=0$) features can use standard pointwise activations. For $l>0$, a popular choice are \emph{gated nonlinearities}~\cite{weiler20183d}: in each feature vector, we reserve one scalar feature $s_c$ per tensorial feature $f^{(l)}_{c}$. Then, we
apply a nonlinearity $\sigma$ to $s_c$ to obtain a scalar prefactor, and scale the tensor, i.e.~all its components, by
\begin{align}
    f^{(l)}_{cm} \;\mapsto\; \sigma(s_c)\,f^{(l)}_{cm} \quad \text{( no sum over } c \text{ )}.
    \label{eq:gating}
\end{align}
Since $\sigma(s_c)$ is a scalar, the scaling commutes with rotations and hence preserves equivariance. In a similar way, normalization layers are typically defined through invariant statistics so that they do not introduce direction-dependent biases~\cite{liao2023equiformerv2}. 

A very related line of work are steerable CNNs developed specifically for data on regular Cartesian grids (images and 3D volumes)~\cite{weiler20183d,e2cnn}. 
Steerable CNNs and $\mathrm{SO}(3)$-equivariant message passing on point clouds share the
same principle: feature channels carry representations and convolution kernels are constrained so that each layer is
equivariant. 
The main difference, besides the data domain, is the operational form of the kernels. 
Steerable CNNs implement equivariance via constrained filter banks and standard discrete convolutions, whereas in tensor field message passing kernels are continuous functions. 
In both cases, equivariance is obtained by composing equivariant operations. \\

Related to the idea of bundling node features in spherical tensors transforming under $\mathrm{SO}(3)$ irreps, specialized architectures exist which use elements of the projective geometric (or Clifford) algebra to achieve Euclidean rotation and translation equivariance~\citep{brehmer2023geometric,ruhe2024clifford}. Notably, there is also an alternative line of work that uses Cartesian tensor representations instead of irreps and that avoids the use of costly CG tensor products~\cite{deng2021vector,satorras2021n,cheng2024cartesian,simeon2024tensornet}.

\subsection{Canonicalization-Based Approaches} \label{sec:canonicalization_bg}
Nonetheless, many popular message passing architectures operating on Euclidean graphs are not equivariant per se \cite{qi2017pointnet++,liu2019relation,wang2019dynamic}. Canonicalization offers an alternative route to exact equivariance that can be built around non-equivariant backbone architectures and avoids specialized architectural building blocks. \\

\paragraph{Group averaging.}
In principle, a simple construction to guarantee invariance (or equivariance) is \emph{group averaging}. For that, one applies a set of transformations to the input (similar to the lifting described in Sec.~\ref{sec:equi_by_group_conv}), runs the same backbone network for each transformed input, and average the outputs. While this provides a principled guarantee, it can be computationally expensive when the group is large or continuous (and one must approximate the average by a finite set). Practical variants include averaging over a small set predicted from the input, as in frame-averaging methods for rotation equivariance~\cite{puny2021frame,pozdnyakov2024smooth}. \\

\paragraph{Global canonicalization.} Instead of averaging over multiple group elements, one may also pick one \emph{canonical pose} in an equivariant way, transform the input into that canonical frame, and then apply a standard backbone. Concretely, a small network (or heuristic) predicts a global orientation, and one canonicalizes the geometric input before processing it with the main model~\cite{kaba2023equivariance}. This factors out the global orientation of the input so that the model output will be invariant. Thus, canonicalization approaches avoid the need for specialized equivariant layers and can provide exact equivariance when the canonical pose is well-defined. The canonical orientation can be estimated in various ways depending on the data modality by using subnetworks~\cite{qi2017pointnet}, PCA-like heuristics~\cite{li2021closer}, asymmetric units~\cite{baker2024explicit}, anchor points~\cite{lou2023crin}, or as a combination of local orientations~\cite{zhao2020quaternion,zhao2022rotation}.
Importantly, if the downstream task requires an \emph{equivariant} (non-scalar) output, one can ``undo'' the canonicalization at the end of the pipeline. That is, one predicts a geometric quantity in the canonical frame and rotates the prediction back to the original frame using the inverse of the estimated pose. \\

\begin{figure}[t]
\centering
\includegraphics[width=0.8\linewidth]{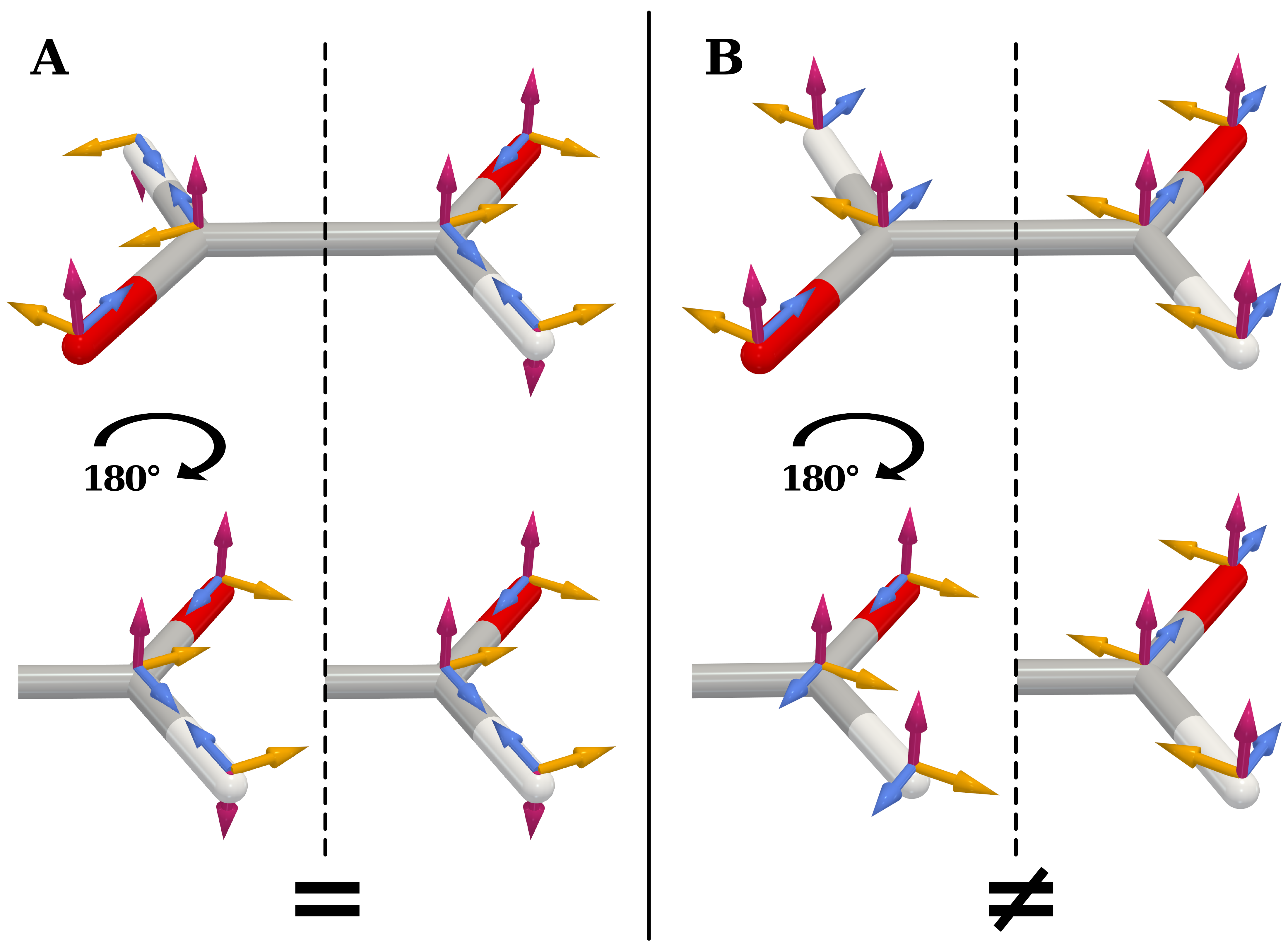} \\
(a) Local Canonicalization
    \hfill (b) Global Canonicalization
\caption{\textbf{Local vs.~global canonicalization of a molecular geometry.}
Left: Local canonicalization with distinct local frames constructed from nearest-neighbor directions. The left and right molecular substructures, which differ only by a $180^\circ$ rotation, are equivalent under local canonicalization.
Right: Under global canonicalization, all ``local'' frames are identical. Consequently, identical local substructures appearing in different orientations are not treated equivalently.
While both approaches achieve global equivariance, only local canonicalization also realizes \emph{local} equivariance.
}
\label{fig:local_vs_global_canonicalization}
\end{figure}

\begin{table*}[t]
\centering
\small
\setlength{\tabcolsep}{6pt}
\renewcommand{\arraystretch}{1.15}

\caption{\textbf{Qualitative comparison of common approaches to rotational equivariance in geometric deep learning.}
\cmark\ indicates the property typically holds, \xmark\ indicates it typically does not hold.
\emph{Continuous $G$} means the method supports continuous (infinite) groups without discretizing the group itself.
\emph{Architectural flexibility} indicates whether equivariance is achieved without enforcing architecural constraints, such as representation-aware nonlinearities.
\emph{Non-scalar messages} refers to whether the method allows to communicate directed geometric information during message passing. The limitations of each approach are highlighted in red circles.
}

\begin{tabular}{lccccccc}
\toprule
\textbf{Approach} &
\textbf{Exact} &
\textbf{Contin.} $\boldsymbol{G}$ &
\textbf{Non-scalar} &
\textbf{Local} &
\textbf{Architectural} &
\textbf{Continuous} \\[-0.15cm]
&\textbf{equiv.} &  & \textbf{messages} & \textbf{equiv.} & \textbf{flexibility} & \textbf{predictions} \\
\midrule
Data augmentation
& \hlcirc{\xmark} & \cmark & \cmark & \hlcirc{\xmark} & \cmark & \cmark \\
Group CNNs
& \cmark & \hlcirc{\xmark} & \cmark & \cmark & \cmark & \cmark \\
Tensor field networks
& \cmark & \cmark & \cmark & \cmark & \hlcirc{\xmark} & \cmark \\
Global canonicalization
& \cmark & \cmark & \cmark & \hlcirc{\xmark} & \cmark & \hlcirc{\xmark} \\
Local canonicalization
& \cmark & \cmark & \hlcirc{\xmark} & \cmark & \cmark & \hlcirc{\xmark} \\
Local can. + tensor msg.
& \cmark & \cmark & \cmark & \cmark & \cmark & \hlcirc{\xmark} \\
\bottomrule
\end{tabular}

\label{tab:equi_methods}
\end{table*}

\paragraph{Local canonicalization.}
Similar to equivariance by global orientation estimation, one can achieve equivariance using \textit{local}
canonicalization~\citep{zhang2020global,wang2022graph,luo2022equivariant,du2022se,du2024new}. In local canonicalization, one equivariantly predicts a local
coordinate frame for each node, into which the geometric input features are transformed. As for global canonicalization, this ensures that the local coordinates of the node features are independent of the global orientation of the input and are thus invariant. 
One advantage of using local reference frames rather than a single global frame is that geometrically similar substructures are described by similar local features. This may be viewed as a form of \emph{local} equivariance, as illustrated in Fig.~\ref{fig:local_vs_global_canonicalization}. More precisely, under local canonicalization, the same local substructure appearing in different orientations is represented by the same local features, because the local reference frame is constructed from the local geometry and therefore rotates together with the substructure. In other words, when expressed in its local frame, the geometry ``looks'' the same irrespective of its global orientation. By contrast, under global canonicalization all local features are expressed in a single global frame, so identical local substructures appearing in different orientations are not represented equivalently. Local canonicalization can thus capture local symmetries that global canonicalization cannot, which may improve learning and generalization.

However, a practical hurdle for local canonicalization is the following \emph{communication barrier}: the same geometric object expressed in different local frames has different coordinates. Therefore, naive message passing between nodes with different frames does not communicate direct geometric information faithfully. Many existing approaches for local canonicalization do not address this issue, which substantially limits the communication between nodes that have different local frames~\cite{lippmann2024tensor}. The authors of~\cite{wang2022graph} and~\cite{du2024new} try to learn an approximate frame-to-frame transition, and in a recent line of work, exact frame-to-frame transitions are used to enable faithful communication of directed, tensor-valued information~\cite{lippmann2024tensor,gerhartz2025equivariance,spinner2025lloca}.

One conceptual, and potentially practical, caveat of (local) canonicalization approaches is the possible \emph{discontinuity} of the predicted local reference frames used to canonicalize geometric features. From a theoretical perspective, there are impossibility results showing that globally continuous canonicalization maps cannot exist in full generality~\cite{dym2024equivariant}. Intuitively, discontinuities arise whenever the input geometry does not uniquely determine an orientation. For perfectly symmetric objects (e.g.\ a sphere), the choice of frame is irrelevant: any reference frame ``sees'' the same input. The more delicate regime is \emph{approximate} symmetry. Consider an almost spherical shape with a tiny dent. Even if this asymmetry is infinitesimal, it singles out a preferred direction, e.g.~the $x$-axis of our predicted frame. However, removing the infinitesimal dent and making a dent in another place, which overall amounts to an infinitesimal deformation, can therefore induce a \emph{finite} change in the predicted $x$-direction. In other words, a near-symmetric input can lead to a locally ill-conditioned orientation estimate.

In practice, one can show that local frames are typically robust to random perturbations of the input geometry, in particular when trained with noisy data~\cite{lippmann2024tensor}. Another promising mitigation strategy is \emph{frame averaging}, e.g. in the form proposed by~\citet{pozdnyakov2024smooth}. The idea is to predict multiple local frames per node and downweigh contributions from those that are close to being discontinuous with respect to the input geometry.\\

\begin{tcolorbox}[
  enhanced,
  boxrule=0.6pt,
  colback=white,
  colframe=black!60,
  arc=2mm,
  left=3mm,right=3mm,top=2mm,bottom=2mm
]
\noindent\textbf{Key point.} Data augmentation offers a straightforward approach to learn approximate equivariances but exact equivariance is typically achieved by stacking specialized layers that are equivariant by design, such as group convolutions or $\mathrm{SO}(3)$-convolutions based on the Clebsch-Gordon tensor product. Canonicalization approaches avoid architectural constraints but rely on the prediction of a well-defined reference pose to canonicalize geometric inputs.  
\end{tcolorbox}

In summary, we have discussed five main approaches to achieve rotational equivariance in geometric deep learning: data augmentation, group convolutions (group CNNs), tensor field networks, global canonicalization, and local canonicalization. Each of these approaches has its own advantages and limitations. Data augmentation is simple but only provides approximate equivariance. Group convolution and internal tensorial representations achieve exact equivariance but require specialized architectural designs and can be computationally expensive. Global canonicalization avoids architectural constraints but may not capture local symmetries effectively. Local canonicalization can additionally capture local symmetries. A qualitative comparison of the different approaches to rotational equivariance is given in Table~\ref{tab:equi_methods}.

\subsection{Conceptual Links between Different Approaches}
\paragraph{Data augmentation and canonicalization.}
Clearly, a global change of reference frame is equivalent to a global rotation of a geometric structure. Using global canonicalization, where the reference frame is chosen canonically and in an equivariant way, the resulting representation becomes independent of the original global orientation (as is shown further below). If, however, the frame is chosen randomly rather than by a principled construction, global ``canonicalization'' effectively reduces to data augmentation by random global rotations.

A similar perspective applies to local canonicalization. First we note that global canonicalization can in fact be viewed as a special case of local canonicalization: if all local frames are chosen identically, this is equivalent to selecting a single global reference frame (see Fig.~\ref{fig:local_vs_global_canonicalization}). However, with local frames constructed equivariantly from the local geometry, the model can achieve local equivariance, as described above. This provides a useful inductive bias, since local motifs that appear in different orientations can then be recognized as rotated versions of the same underlying pattern. This raises the question of what happens when local frames are chosen at random. In that case, the model should not rely on information encoded in the specific frame choice; its predictions should be independent of the chosen gauge, that is, of the particular set of local frames. This viewpoint connects local canonicalization to gauge-equivariant neural networks~\cite{cohen2019gauge,de2020gauge}, which likewise make use of local frames but construct their operations so that predictions are independent of the specific frame choice.

An ablation study in~\cite{lippmann2024tensor} compares geometrically constructed local frames with randomly chosen ones. The results suggest that informative, geometry-adapted local frames are beneficial when the model is allowed to exploit them explicitly. This highlights an important trade-off between robustness to frame choice and the additional geometric signal that can be extracted from informative canonical frames. \\

\paragraph{A scalarization and tensorization perspective on canonicalization.} \label{sec:scalarization_tensorization}
At the core of (local) canonicalization stands the idea that by transforming geometric features into \emph{equivariantly predicted} (local) reference frames, the coordinates of the geometric features become invariant. Concretely, each local frame is represented by a rotation matrix $R\in \mathrm{SO}(3)$ whose rows $\mathbf n_1 ^\mathrm T, \mathbf n_2 ^\mathrm T, \mathbf n_3 ^\mathrm T$ are geometric basis vectors that transform equivariantly under global rotations. Concretely,
\begin{align} \label{eq:lframe_from_basis}
    R_i = (\mathbf n_{i,1}, \mathbf n_{i,2}, \mathbf n_{i,3})^\mathrm{T} = \begin{pmatrix}
    \rule[0.5ex]{0.5cm}{0.55pt} \ \mathbf n_{i,1} \ \rule[0.5ex]{0.5cm}{0.55pt}\\
    \rule[0.5ex]{0.5cm}{0.55pt} \ \mathbf n_{i,2} \ \rule[0.5ex]{0.5cm}{0.55pt}\\
    \rule[0.5ex]{0.5cm}{0.55pt} \ \mathbf n_{i,3} \ \rule[0.5ex]{0.5cm}{0.55pt}\\
    \end{pmatrix} .
\end{align}
In the literature, this change of coordinates, which amounts to a projection onto the basis vectors $\mathbf n_i$, is sometimes referred to as \emph{scalarization}~\cite{du2024new}. After processing the resulting invariant quantities with standard (non-equivariant) building blocks, the output is transformed back to the global frame, a step that can be viewed as \emph{tensorization}. 

Let us briefly illustrate the perspective of scalarization and tensorization in formulae.
Let $\mathbf x\in\mathbb R^3$ be a vector-valued feature in the global frame and let $R\in\mathrm{SO}(3)$ denote the local frame with $R_{ij} = (\mathbf n_i)_j$, i.e. $R_{ij}$ is given by the $j$-th component of the $i$-th basis vector. When transforming $\mathbf x$ into the local frame $R$, we obtain a \emph{scalarized} representation of $\mathbf x$ in the form of its local coordinates,
\begin{align}
\mathbf s = R \mathbf x \in \mathbb R^3, 
\qquad\text{with}\qquad
s_i \;=\; \mathbf n_i ^\mathrm T \mathbf x,\quad i=1,2,3.
\label{eq:scalarize_vec}
\end{align}
That is, in the transformation into the local frame the vector $\mathbf x$ is projected onto each basis vector $\mathbf n_i$ to obtain scalar coordinates $s_i$.
If the input is rotated by a global rotation $Q\in\mathrm{SO}(3)$, the vector transforms as $\mathbf x\mapsto \mathbf x' = Q \mathbf x$. Furthermore, equivariant frame prediction implies that 
$R\mapsto R' = RQ ^\mathrm T$, since  $\mathbf n_i \mapsto \mathbf n_i' = Q \mathbf n_i$ for each $i$ (cf.~Eq.~\ref{eq:lframe_from_basis}). Consequently,
\begin{align}
\mathbf s' \;=\; R' \mathbf x' \;=\; (RQ ^\mathrm T)(Q \mathbf x) \;=\; R \mathbf x \;=\; \mathbf s,
\end{align}
so the coordinates $s_i$ are indeed invariant under global rotations. After processing $s_i$ with an arbitrary (non-equivariant) function, we obtain new scalar coordinates. Let us group three of these into $\tilde s_i, \ i=1,2,3$. Then, to produce an equivariant vector output $\tilde{\mathbf x}$, we \emph{tensorize} $\tilde s_i$ by rotating back into the global frame via
\begin{align}
\tilde{\mathbf x} = R^\mathrm T\, \tilde{\mathbf s}
\quad \text{or equivalently} \quad \tilde{\mathbf x} = \sum_{i=1}^3 \tilde s_i \mathbf n_i .
\label{eq:tensorize_vec}
\end{align}
That is, the transformation back into the global frame of reference amounts to a linear combination of the basis vectors $\mathbf n_i$ with the scalar coefficients $\tilde s_i$.
Indeed, since the basis vectors are predicted equivariantly, it follows directly that $\tilde{\mathbf x}$ transforms equivariantly under a global rotation $Q$, i.e.~$\tilde{\mathbf x}' = Q\tilde{\mathbf x}$.

The same viewpoint extends to rank-$2$ Cartesian tensors, which are transformed into a local frame, i.e.~scalarized, via
\begin{align}
S \;=\; R \, T R ^\mathrm T \in \mathbb R^{3\times 3},
\qquad\text{with}\qquad
S_{ij} \;=\; \mathbf n_i ^\mathrm T T \mathbf n_j.
\label{eq:scalarize_mat}
\end{align}
Equivalently, nine invariant local features grouped into $\tilde S \in \mathbb R^{3\times 3}$ are tensorized by transforming them back into the global frame of reference via
\begin{align}
\tilde T \;=\; R ^\mathrm T\,\tilde S\,R, 
\quad \text{or equivalently} \quad \tilde T = \sum_{i,j} \tilde S_{ij} \mathbf n_i \mathbf n_j ^\mathrm T.
\label{eq:tensorize_mat}
\end{align}
Clearly, as a sum of outer products of the basis vectors $\mathbf n_i$, the output $\tilde T$ transforms equivariantly under global rotations. Higher-order Cartesian tensors follow analogously: one scalarizes by contracting each index with a basis vector of the local frame, and tensorizes by expanding back with the same basis. \\

\paragraph{Connection to tensor field networks and irreducible representations.}
This perspective of scalarization and tensorization can be used to connect (local) canonicalization to e3nn-style architectures and equivariant message passing based on $\mathrm{SO}(3)$-convolutions as discussed in Sec.~\ref{sec:tensor_field_networks_bg}. The coupled Clebsch-Gordan product between spherical tensor features $f^{(l_1)}$ and the spherical harmonics $Y^{(l_2)}(\hat{\mathbf r}_{ij})$, cf.~Eq.~\eqref{eq:tensor_product_message_bg}, contains a scalar output path exactly if $l_1 = l_2 = l$ due to the composition rule Eq.~\eqref{eq:CG_dim_rule}. In that case, we may interpret the scalar output channel
\begin{align} 
f^{(0)}:= \big(f^{(l)} \otimes_{\mathrm{CG}} Y^{(l)}(\hat{\mathbf r}_{ij}) \big)^{(0)} = \sum_{m=-l}^l f^{(l)}_m Y_{lm}(\hat{\mathbf r}_{ij})
\end{align}
as a scalarization, in which we project the feature $f^{(l)}$ onto the basis given by the spherical harmonics $Y_{lm}(\hat{\mathbf r}_{ij})$. The resulting scalar feature $f^{(0)}$ is invariant under global rotations and is typically processed with unconstrained building blocks in e3nn-style architectures (as in canonicalization-based approaches). Afterwards, a CG-tensor product with spherical harmonics $Y^{(l)}(\hat{\mathbf r}_{ij})$ can be used to tensorize the scalar features $\tilde f^{(0)}$ through the only available path:
\begin{align}
\big(\tilde f^{(0)} \otimes_{\mathrm{CG}} Y^{(l)}(\hat{\mathbf r}_{ij}) \big)^{(l)} = \tilde f^{(0)} Y^{(l)}(\hat{\mathbf r}_{ij}) =: \tilde f^{(l)},
\label{eq:tensorize_sph}
\end{align}
i.e.~the scalar multiplication of the vector of spherical harmonics $Y^{(l)}(\hat{\mathbf r}_{ij})$,
resulting in a spherical tensor $\tilde f^{(l)}$.

Focusing only on the scalar paths of the Clebsch-Gordan tensor product in the beginning of the pipeline (scalarization) and on the scalar multiplication of spherical features provided by the spherical harmonics in the end (tensorization) reveals an interesting connection between canonicalization and e3nn-style architectures. One key difference (besides the other tensorial channels in e3nn-style architectures) is that in (local) canonicalization, we project and expand tensorial features with a complete basis, i.e. three orthonormal vectors, instead of the projection only defined by the relative distance vector $\hat{\mathbf r}_{ij}$ in the case of e3nn-style architectures.

\section{Problems with Rotational Equivariance in Practice} \label{sec:problems_with_equi}
As we have seen, equivariant architectures are conceptually well motivated and have become a standard tool in geometric deep learning for good reasons. In practice, however, the specialized building blocks are usually computationally more demanding or do not align as well with highly optimized hardware
primitives, such as dense matrix multiplication, leading to substantial runtime overheads~\cite{passaro2023reducing}. Further, the constraints imposed by exact equivariance can
complicate optimization: initialization and training recipes are less standardized, and several works report that such models require careful tuning and can be difficult to train successfully at scale~\cite{wang2023swallowing,pertigkiozoglou2024improving,elhag2024relaxed,qu2024importance}.
For example, \citet{pertigkiozoglou2024improving} emphasize that equivariant models ``can be difficult to optimize and require careful hyperparameter tuning to train successfully''.
These issues matter in different ways across contexts: while large-scale industrial settings may
afford substantial compute, the dominant bottleneck is often human time for model development, and debugging; in academic settings, both compute budgets and the added
engineering complexity can become limiting factors. 

As a consequence, while exact equivariance remains desirable in principle, relaxing or even dropping built-in symmetries has become a (re-)emerging pattern in practice, particularly in molecular machine learning~\cite{wang2023swallowing,langer2024probing,eissler2025simple,manolache2025learning,berndt2026approximate,bigi2026pushing}. A prominent example is AlphaFold~3, where the authors state that ``we find that no invariance or equivariance with respect to global rotations and translation of the molecule are required in the architecture and so we omit them to simplify the machine learning architecture''~\cite{abramson2024accurate}. This trend highlights a central tradeoff: we would like the \emph{physical guarantees} of exact equivariance, but without sacrificing the \emph{architectural freedom} and efficiency of standard deep learning components.

In the context of rotational equivariance, some recent works even suggest that enforcing exact symmetries may hinder optimal scaling behavior as more data becomes available~\cite{wang2023swallowing,qu2024importance}. This hypothesis resonates with the tough message of Richard Sutton's ``bitter lesson''~\cite{sutton2019bitter}: while incorporating prior knowledge into architectures can yield short-term gains and is intellectually satisfying, Richard Sutton postulates that it inhibits the long-run progress of deep learning that comes from scaling general-purpose architectures.
One may argue that we have indeed observed Sutton's bitter lesson in practice in two prominent examples. First, deep learning models in general apply end-to-end representation learning, where the network discovers its own features from data, instead of using hand-crafted input features. Second, transformer-based Large Language Models (LLMs) are immensely successful at scale and rely on the general-purpose self-attention mechanism (see Sec.~\ref{sec:geometric_dl}), which can be interpreted as a content-dependent dynamic routing of information; in this sense, the model learns content-dependent patterns (or biases) from data rather than enforcing these. Relatedly, modern LLMs utilize content-dependent routing mechanisms such as mixture-of-experts~\cite{jacobs1991adaptive,fedus2022switch}, another example of data-driven specialization. At the same time, even transformers incorporate inductive biases. As discussed in Sec.~\ref{sec:geometric_dl}, they can be seen as an instance of message passing networks and as such enforce exact permutation equivariance. At least for such a large symmetry group, enforcing equivariance is indispensable, right? 

Regarding rotational equivariance, for instance,~\citet{batzner20223} reported improved data efficiency for models with built-in symmetry, arguing that such models need not allocate network capacity to learning the symmetry from data. The least we can say today is that the current picture is less clear~\cite{brehmer2024does} and that there is conflicting empirical evidence.
This connects to the recent finding that built-in equivariance via local canonicalization can be more data efficient -- in the sense that model performance improves faster as more data becomes available -- while data augmentation can yield better accuracies in the low-data regime~\cite{lippmann2024tensor,gerhartz2025equivariance,spinner2025lloca}. Put differently, equivariance often corresponds to a steeper scaling curve, but a steeper slope does not necessarily imply better absolute accuracy for small datasets. For these reasons, across-the-board prescriptions to either always enforce equivariance, or never enforce equivariance, seem unhelpful. In practice, it is rarely clear \emph{a priori} how important exact symmetry is for a given problem, and how this interacts with dataset size, target quantity, and the optimization landscape. We believe that conducting more systematic side-by-side comparisons of approximate equivariance via data augmentation and built-in equivariance is an interesting avenue for future research.

\section{Conclusion}
Rotational equivariance provides a principled way to encode coordinate independence in machine learning models for 3D data, ensuring that models respond to rotations in a structured and physically meaningful way. This tutorial has developed the subject from first principles to practical constructions, beginning with symmetry and coordinate independence, and building through geometric deep learning, group representations, spherical harmonics, Wigner matrices, tensor products, and Clebsch-Gordan decomposition to the main architectural strategies used in modern equivariant learning. By relating these ingredients to one another and making explicit the conceptual links between different formalisms, the tutorial aims to give readers a coherent understanding of a field whose central ideas are often introduced in different mathematical and architectural languages.

Understanding rotational equivariance is useful not only because it clarifies the theoretical foundations of modern 3D machine learning, but also because it informs practical model design. In applications, one must choose between different ways of incorporating geometric structure, each with its own trade-offs in expressivity, computational efficiency, implementation complexity, and numerical stability. It is therefore important to understand when exact symmetry should be built into a model, when approximate methods may be sufficient, and how different architectural choices reflect different assumptions about the task and the data. This is particularly relevant in scientific and engineering settings, where high-quality 3D data are often limited and simulation data can be costly to obtain, so that appropriate symmetry-aware inductive biases can play an important role in improving data efficiency and generalization.

\begin{acknowledgments}
This work has received funding from the Klaus Tschira Stiftung gGmbH (Simplaix project) and the Wildcard program from the Carl Zeiss Stiftung. Further, it is supported by Deutsche Forschungsgemeinschaft (DFG) under Germany's Excellence Strategy EXC-2181/1 -- 390900948 (the Heidelberg STRUCTURES Excellence Cluster) as well as under project number 240245660 -- SFB 1129.
The authors acknowledge support by the state of Baden-Württemberg through bwHPC
and the German Research Foundation (DFG) through grant INST 35/1597-1 FUGG.
\end{acknowledgments}

\bibliography{apssamp}

\providecommand{\noopsort}[1]{}\providecommand{\singleletter}[1]{#1}%
\begin{thebibliography}{87}%
\makeatletter
\providecommand \@ifxundefined [1]{%
 \@ifx{#1\undefined}
}%
\providecommand \@ifnum [1]{%
 \ifnum #1\expandafter \@firstoftwo
 \else \expandafter \@secondoftwo
 \fi
}%
\providecommand \@ifx [1]{%
 \ifx #1\expandafter \@firstoftwo
 \else \expandafter \@secondoftwo
 \fi
}%
\providecommand \natexlab [1]{#1}%
\providecommand \enquote  [1]{``#1''}%
\providecommand \bibnamefont  [1]{#1}%
\providecommand \bibfnamefont [1]{#1}%
\providecommand \citenamefont [1]{#1}%
\providecommand \href@noop [0]{\@secondoftwo}%
\providecommand \href [0]{\begingroup \@sanitize@url \@href}%
\providecommand \@href[1]{\@@startlink{#1}\@@href}%
\providecommand \@@href[1]{\endgroup#1\@@endlink}%
\providecommand \@sanitize@url [0]{\catcode `\\12\catcode `\$12\catcode
  `\&12\catcode `\#12\catcode `\^12\catcode `\_12\catcode `\%12\relax}%
\providecommand \@@startlink[1]{}%
\providecommand \@@endlink[0]{}%
\providecommand \url  [0]{\begingroup\@sanitize@url \@url }%
\providecommand \@url [1]{\endgroup\@href {#1}{\urlprefix }}%
\providecommand \urlprefix  [0]{URL }%
\providecommand \Eprint [0]{\href }%
\providecommand \doibase [0]{https://doi.org/}%
\providecommand \selectlanguage [0]{\@gobble}%
\providecommand \bibinfo  [0]{\@secondoftwo}%
\providecommand \bibfield  [0]{\@secondoftwo}%
\providecommand \translation [1]{[#1]}%
\providecommand \BibitemOpen [0]{}%
\providecommand \bibitemStop [0]{}%
\providecommand \bibitemNoStop [0]{.\EOS\space}%
\providecommand \EOS [0]{\spacefactor3000\relax}%
\providecommand \BibitemShut  [1]{\csname bibitem#1\endcsname}%
\let\auto@bib@innerbib\@empty
\bibitem [{\citenamefont {Cohen}\ and\ \citenamefont
  {Welling}(2016)}]{cohen2016group}%
  \BibitemOpen
  \bibfield  {author} {\bibinfo {author} {\bibfnamefont {T.}~\bibnamefont
  {Cohen}}\ and\ \bibinfo {author} {\bibfnamefont {M.}~\bibnamefont
  {Welling}},\ }\bibfield  {title} {\bibinfo {title} {Group equivariant
  convolutional networks},\ }in\ \href@noop {} {\emph {\bibinfo {booktitle}
  {International conference on machine learning}}}\ (\bibinfo {organization}
  {PMLR},\ \bibinfo {year} {2016})\ pp.\ \bibinfo {pages}
  {2990--2999}\BibitemShut {NoStop}%
\bibitem [{\citenamefont {Cohen}\ \emph
  {et~al.}(2019{\natexlab{a}})\citenamefont {Cohen}, \citenamefont {Geiger},\
  and\ \citenamefont {Weiler}}]{cohen2019general}%
  \BibitemOpen
  \bibfield  {author} {\bibinfo {author} {\bibfnamefont {T.~S.}\ \bibnamefont
  {Cohen}}, \bibinfo {author} {\bibfnamefont {M.}~\bibnamefont {Geiger}},\ and\
  \bibinfo {author} {\bibfnamefont {M.}~\bibnamefont {Weiler}},\ }\bibfield
  {title} {\bibinfo {title} {A general theory of equivariant cnns on
  homogeneous spaces},\ }\href@noop {} {\bibfield  {journal} {\bibinfo
  {journal} {Advances in neural information processing systems}\ }\textbf
  {\bibinfo {volume} {32}} (\bibinfo {year} {2019}{\natexlab{a}})}\BibitemShut
  {NoStop}%
\bibitem [{\citenamefont {Bronstein}\ \emph {et~al.}(2021)\citenamefont
  {Bronstein}, \citenamefont {Bruna}, \citenamefont {Cohen},\ and\
  \citenamefont {Veli{\v{c}}kovi{\'c}}}]{bronstein2021geometric}%
  \BibitemOpen
  \bibfield  {author} {\bibinfo {author} {\bibfnamefont {M.~M.}\ \bibnamefont
  {Bronstein}}, \bibinfo {author} {\bibfnamefont {J.}~\bibnamefont {Bruna}},
  \bibinfo {author} {\bibfnamefont {T.}~\bibnamefont {Cohen}},\ and\ \bibinfo
  {author} {\bibfnamefont {P.}~\bibnamefont {Veli{\v{c}}kovi{\'c}}},\
  }\bibfield  {title} {\bibinfo {title} {Geometric deep learning: Grids,
  groups, graphs, geodesics, and gauges},\ }\href@noop {} {\bibfield  {journal}
  {\bibinfo  {journal} {arXiv preprint arXiv:2104.13478}\ } (\bibinfo {year}
  {2021})}\BibitemShut {NoStop}%
\bibitem [{\citenamefont {Duval}\ \emph {et~al.}(2024)\citenamefont {Duval},
  \citenamefont {Mathis}, \citenamefont {Joshi}, \citenamefont {Schmidt},
  \citenamefont {Miret}, \citenamefont {Malliaros}, \citenamefont {Cohen},
  \citenamefont {Liò}, \citenamefont {Bengio},\ and\ \citenamefont
  {Bronstein}}]{duval2023hitchhiker}%
  \BibitemOpen
  \bibfield  {author} {\bibinfo {author} {\bibfnamefont {A.}~\bibnamefont
  {Duval}}, \bibinfo {author} {\bibfnamefont {S.~V.}\ \bibnamefont {Mathis}},
  \bibinfo {author} {\bibfnamefont {C.~K.}\ \bibnamefont {Joshi}}, \bibinfo
  {author} {\bibfnamefont {V.}~\bibnamefont {Schmidt}}, \bibinfo {author}
  {\bibfnamefont {S.}~\bibnamefont {Miret}}, \bibinfo {author} {\bibfnamefont
  {F.~D.}\ \bibnamefont {Malliaros}}, \bibinfo {author} {\bibfnamefont
  {T.}~\bibnamefont {Cohen}}, \bibinfo {author} {\bibfnamefont
  {P.}~\bibnamefont {Liò}}, \bibinfo {author} {\bibfnamefont {Y.}~\bibnamefont
  {Bengio}},\ and\ \bibinfo {author} {\bibfnamefont {M.}~\bibnamefont
  {Bronstein}},\ }\href@noop {} {\bibinfo {title} {A hitchhiker's guide to
  geometric gnns for 3d atomic systems}},\ \bibinfo {howpublished}
  {arXiv:2312.07511 [cs.LG]} (\bibinfo {year} {2024}),\ \Eprint
  {https://arxiv.org/abs/arXiv:2312.07511} {arXiv:2312.07511 [cs.LG]}
  \BibitemShut {NoStop}%
\bibitem [{\citenamefont {Batatia}\ \emph {et~al.}(2022)\citenamefont
  {Batatia}, \citenamefont {Kovacs}, \citenamefont {Simm}, \citenamefont
  {Ortner},\ and\ \citenamefont {Csanyi}}]{batatia2022mace}%
  \BibitemOpen
  \bibfield  {author} {\bibinfo {author} {\bibfnamefont {I.}~\bibnamefont
  {Batatia}}, \bibinfo {author} {\bibfnamefont {D.~P.}\ \bibnamefont {Kovacs}},
  \bibinfo {author} {\bibfnamefont {G.}~\bibnamefont {Simm}}, \bibinfo {author}
  {\bibfnamefont {C.}~\bibnamefont {Ortner}},\ and\ \bibinfo {author}
  {\bibfnamefont {G.}~\bibnamefont {Csanyi}},\ }\bibfield  {title} {\bibinfo
  {title} {{MACE}: Higher order equivariant message passing neural networks for
  fast and accurate force fields},\ }in\ \href@noop {} {\emph {\bibinfo
  {booktitle} {Advances in Neural Information Processing Systems}}},\
  Vol.~\bibinfo {volume} {35}\ (\bibinfo {year} {2022})\ pp.\ \bibinfo {pages}
  {11423--11436}\BibitemShut {NoStop}%
\bibitem [{\citenamefont {Liao}\ \emph
  {et~al.}(2024{\natexlab{a}})\citenamefont {Liao}, \citenamefont {Wood},
  \citenamefont {Das},\ and\ \citenamefont {Smidt}}]{liao2023equiformerv2}%
  \BibitemOpen
  \bibfield  {author} {\bibinfo {author} {\bibfnamefont {Y.-L.}\ \bibnamefont
  {Liao}}, \bibinfo {author} {\bibfnamefont {B.~M.}\ \bibnamefont {Wood}},
  \bibinfo {author} {\bibfnamefont {A.}~\bibnamefont {Das}},\ and\ \bibinfo
  {author} {\bibfnamefont {T.}~\bibnamefont {Smidt}},\ }\bibfield  {title}
  {\bibinfo {title} {{EquiformerV2}: Improved equivariant transformer for
  scaling to higher-degree representations},\ }in\ \href
  {https://openreview.net/forum?id=mCOBKZmrzD} {\emph {\bibinfo {booktitle}
  {The Twelfth International Conference on Learning Representations}}}\
  (\bibinfo {year} {2024})\BibitemShut {NoStop}%
\bibitem [{\citenamefont {Lou}\ \emph {et~al.}(2023)\citenamefont {Lou},
  \citenamefont {Ye}, \citenamefont {You}, \citenamefont {Jiang}, \citenamefont
  {Lu}, \citenamefont {Wang}, \citenamefont {Ma},\ and\ \citenamefont
  {Lu}}]{lou2023crin}%
  \BibitemOpen
  \bibfield  {author} {\bibinfo {author} {\bibfnamefont {Y.}~\bibnamefont
  {Lou}}, \bibinfo {author} {\bibfnamefont {Z.}~\bibnamefont {Ye}}, \bibinfo
  {author} {\bibfnamefont {Y.}~\bibnamefont {You}}, \bibinfo {author}
  {\bibfnamefont {N.}~\bibnamefont {Jiang}}, \bibinfo {author} {\bibfnamefont
  {J.}~\bibnamefont {Lu}}, \bibinfo {author} {\bibfnamefont {W.}~\bibnamefont
  {Wang}}, \bibinfo {author} {\bibfnamefont {L.}~\bibnamefont {Ma}},\ and\
  \bibinfo {author} {\bibfnamefont {C.}~\bibnamefont {Lu}},\ }\bibfield
  {title} {\bibinfo {title} {Crin: rotation-invariant point cloud analysis and
  rotation estimation via centrifugal reference frame},\ }in\ \href@noop {}
  {\emph {\bibinfo {booktitle} {Proceedings of the AAAI Conference on
  Artificial Intelligence}}},\ Vol.~\bibinfo {volume} {37}\ (\bibinfo {year}
  {2023})\ pp.\ \bibinfo {pages} {1817--1825}\BibitemShut {NoStop}%
\bibitem [{\citenamefont {Du}\ \emph {et~al.}(2024)\citenamefont {Du},
  \citenamefont {Wang}, \citenamefont {Feng}, \citenamefont {Wang},
  \citenamefont {Ji}, \citenamefont {Gomes}, \citenamefont {Ma} \emph
  {et~al.}}]{du2024new}%
  \BibitemOpen
  \bibfield  {author} {\bibinfo {author} {\bibfnamefont {Y.}~\bibnamefont
  {Du}}, \bibinfo {author} {\bibfnamefont {L.}~\bibnamefont {Wang}}, \bibinfo
  {author} {\bibfnamefont {D.}~\bibnamefont {Feng}}, \bibinfo {author}
  {\bibfnamefont {G.}~\bibnamefont {Wang}}, \bibinfo {author} {\bibfnamefont
  {S.}~\bibnamefont {Ji}}, \bibinfo {author} {\bibfnamefont {C.~P.}\
  \bibnamefont {Gomes}}, \bibinfo {author} {\bibfnamefont {Z.-M.}\ \bibnamefont
  {Ma}}, \emph {et~al.},\ }\bibfield  {title} {\bibinfo {title} {A new
  perspective on building efficient and expressive {3D} equivariant graph
  neural networks},\ }\href@noop {} {\bibfield  {journal} {\bibinfo  {journal}
  {Advances in Neural Information Processing Systems}\ }\textbf {\bibinfo
  {volume} {36}} (\bibinfo {year} {2024})}\BibitemShut {NoStop}%
\bibitem [{\citenamefont {Aykent}\ and\ \citenamefont
  {Xia}(2025)}]{aykent2025gotennet}%
  \BibitemOpen
  \bibfield  {author} {\bibinfo {author} {\bibfnamefont {S.}~\bibnamefont
  {Aykent}}\ and\ \bibinfo {author} {\bibfnamefont {T.}~\bibnamefont {Xia}},\
  }\bibfield  {title} {\bibinfo {title} {Gotennet: Rethinking efficient 3d
  equivariant graph neural networks},\ }in\ \href@noop {} {\emph {\bibinfo
  {booktitle} {The Thirteenth International Conference on Learning
  Representations}}}\ (\bibinfo {year} {2025})\BibitemShut {NoStop}%
\bibitem [{\citenamefont {Lam}\ \emph {et~al.}(2023)\citenamefont {Lam},
  \citenamefont {Sanchez-Gonzalez}, \citenamefont {Willson}, \citenamefont
  {Wirnsberger}, \citenamefont {Fortunato}, \citenamefont {Alet}, \citenamefont
  {Ravuri}, \citenamefont {Ewalds}, \citenamefont {Eaton-Rosen}, \citenamefont
  {Hu} \emph {et~al.}}]{lam2023learning}%
  \BibitemOpen
  \bibfield  {author} {\bibinfo {author} {\bibfnamefont {R.}~\bibnamefont
  {Lam}}, \bibinfo {author} {\bibfnamefont {A.}~\bibnamefont
  {Sanchez-Gonzalez}}, \bibinfo {author} {\bibfnamefont {M.}~\bibnamefont
  {Willson}}, \bibinfo {author} {\bibfnamefont {P.}~\bibnamefont
  {Wirnsberger}}, \bibinfo {author} {\bibfnamefont {M.}~\bibnamefont
  {Fortunato}}, \bibinfo {author} {\bibfnamefont {F.}~\bibnamefont {Alet}},
  \bibinfo {author} {\bibfnamefont {S.}~\bibnamefont {Ravuri}}, \bibinfo
  {author} {\bibfnamefont {T.}~\bibnamefont {Ewalds}}, \bibinfo {author}
  {\bibfnamefont {Z.}~\bibnamefont {Eaton-Rosen}}, \bibinfo {author}
  {\bibfnamefont {W.}~\bibnamefont {Hu}}, \emph {et~al.},\ }\bibfield  {title}
  {\bibinfo {title} {Learning skillful medium-range global weather
  forecasting},\ }\href@noop {} {\bibfield  {journal} {\bibinfo  {journal}
  {Science}\ }\textbf {\bibinfo {volume} {382}},\ \bibinfo {pages} {1416}
  (\bibinfo {year} {2023})}\BibitemShut {NoStop}%
\bibitem [{\citenamefont {Spinner}\ \emph {et~al.}(2024)\citenamefont
  {Spinner}, \citenamefont {Bres{\'o}}, \citenamefont {De~Haan}, \citenamefont
  {Plehn}, \citenamefont {Thaler},\ and\ \citenamefont
  {Brehmer}}]{spinner2025lorentz}%
  \BibitemOpen
  \bibfield  {author} {\bibinfo {author} {\bibfnamefont {J.}~\bibnamefont
  {Spinner}}, \bibinfo {author} {\bibfnamefont {V.}~\bibnamefont {Bres{\'o}}},
  \bibinfo {author} {\bibfnamefont {P.}~\bibnamefont {De~Haan}}, \bibinfo
  {author} {\bibfnamefont {T.}~\bibnamefont {Plehn}}, \bibinfo {author}
  {\bibfnamefont {J.}~\bibnamefont {Thaler}},\ and\ \bibinfo {author}
  {\bibfnamefont {J.}~\bibnamefont {Brehmer}},\ }\bibfield  {title} {\bibinfo
  {title} {Lorentz-equivariant geometric algebra transformers for high-energy
  physics},\ }in\ \href {https://arxiv.org/abs/2405.14806} {\emph {\bibinfo
  {booktitle} {Advances in Neural Information Processing Systems}}},\
  Vol.~\bibinfo {volume} {37}\ (\bibinfo {year} {2024})\ \Eprint
  {https://arxiv.org/abs/2405.14806} {2405.14806} \BibitemShut {NoStop}%
\bibitem [{\citenamefont {Remme}\ \emph {et~al.}(2025)\citenamefont {Remme},
  \citenamefont {Kaczun}, \citenamefont {Ebert}, \citenamefont {Gehrig},
  \citenamefont {Geng}, \citenamefont {Gerhartz}, \citenamefont {Ickler},
  \citenamefont {Klockow}, \citenamefont {Lippmann}, \citenamefont {Schmidt},
  \citenamefont {Wagner}, \citenamefont {Dreuw},\ and\ \citenamefont
  {Hamprecht}}]{remme2025stable}%
  \BibitemOpen
  \bibfield  {author} {\bibinfo {author} {\bibfnamefont {R.}~\bibnamefont
  {Remme}}, \bibinfo {author} {\bibfnamefont {T.}~\bibnamefont {Kaczun}},
  \bibinfo {author} {\bibfnamefont {T.}~\bibnamefont {Ebert}}, \bibinfo
  {author} {\bibfnamefont {C.~A.}\ \bibnamefont {Gehrig}}, \bibinfo {author}
  {\bibfnamefont {D.}~\bibnamefont {Geng}}, \bibinfo {author} {\bibfnamefont
  {G.}~\bibnamefont {Gerhartz}}, \bibinfo {author} {\bibfnamefont {M.~K.}\
  \bibnamefont {Ickler}}, \bibinfo {author} {\bibfnamefont {M.~V.}\
  \bibnamefont {Klockow}}, \bibinfo {author} {\bibfnamefont {P.}~\bibnamefont
  {Lippmann}}, \bibinfo {author} {\bibfnamefont {J.~S.}\ \bibnamefont
  {Schmidt}}, \bibinfo {author} {\bibfnamefont {S.}~\bibnamefont {Wagner}},
  \bibinfo {author} {\bibfnamefont {A.}~\bibnamefont {Dreuw}},\ and\ \bibinfo
  {author} {\bibfnamefont {F.~A.}\ \bibnamefont {Hamprecht}},\ }\bibfield
  {title} {\bibinfo {title} {Stable and accurate orbital-free density
  functional theory powered by machine learning},\ }\href@noop {} {\bibfield
  {journal} {\bibinfo  {journal} {Journal of the American Chemical Society}\
  }\textbf {\bibinfo {volume} {147}},\ \bibinfo {pages} {28851} (\bibinfo
  {year} {2025})}\BibitemShut {NoStop}%
\bibitem [{\citenamefont {Geiger}\ and\ \citenamefont
  {Smidt}(2022)}]{geiger2022e3nn}%
  \BibitemOpen
  \bibfield  {author} {\bibinfo {author} {\bibfnamefont {M.}~\bibnamefont
  {Geiger}}\ and\ \bibinfo {author} {\bibfnamefont {T.}~\bibnamefont {Smidt}},\
  }\bibfield  {title} {\bibinfo {title} {e3nn: Euclidean neural networks},\
  }\href@noop {} {\bibfield  {journal} {\bibinfo  {journal} {arXiv preprint
  arXiv:2207.09453}\ } (\bibinfo {year} {2022})}\BibitemShut {NoStop}%
\bibitem [{\citenamefont {Lippmann}\ \emph {et~al.}(2025)\citenamefont
  {Lippmann}, \citenamefont {Gerhartz}, \citenamefont {Remme},\ and\
  \citenamefont {Hamprecht}}]{lippmann2024tensor}%
  \BibitemOpen
  \bibfield  {author} {\bibinfo {author} {\bibfnamefont {P.}~\bibnamefont
  {Lippmann}}, \bibinfo {author} {\bibfnamefont {G.}~\bibnamefont {Gerhartz}},
  \bibinfo {author} {\bibfnamefont {R.}~\bibnamefont {Remme}},\ and\ \bibinfo
  {author} {\bibfnamefont {F.~A.}\ \bibnamefont {Hamprecht}},\ }\bibfield
  {title} {\bibinfo {title} {Beyond canonicalization: How tensorial messages
  improve equivariant message passing},\ }in\ \href
  {https://proceedings.iclr.cc/paper_files/paper/2025/file/db7534a06ace69f4ec95bc89e91d5dbb-Paper-Conference.pdf}
  {\emph {\bibinfo {booktitle} {International Conference on Representation
  Learning}}},\ Vol.\ \bibinfo {volume} {2025},\ \bibinfo {editor} {edited by\
  \bibinfo {editor} {\bibfnamefont {Y.}~\bibnamefont {Yue}}, \bibinfo {editor}
  {\bibfnamefont {A.}~\bibnamefont {Garg}}, \bibinfo {editor} {\bibfnamefont
  {N.}~\bibnamefont {Peng}}, \bibinfo {editor} {\bibfnamefont {F.}~\bibnamefont
  {Sha}},\ and\ \bibinfo {editor} {\bibfnamefont {R.}~\bibnamefont {Yu}}}\
  (\bibinfo {year} {2025})\ pp.\ \bibinfo {pages} {88067--88087}\BibitemShut
  {NoStop}%
\bibitem [{\citenamefont {Batzner}\ \emph {et~al.}(2022)\citenamefont
  {Batzner}, \citenamefont {Musaelian}, \citenamefont {Sun}, \citenamefont
  {Geiger}, \citenamefont {Mailoa}, \citenamefont {Kornbluth}, \citenamefont
  {Molinari}, \citenamefont {Smidt},\ and\ \citenamefont
  {Kozinsky}}]{batzner20223}%
  \BibitemOpen
  \bibfield  {author} {\bibinfo {author} {\bibfnamefont {S.}~\bibnamefont
  {Batzner}}, \bibinfo {author} {\bibfnamefont {A.}~\bibnamefont {Musaelian}},
  \bibinfo {author} {\bibfnamefont {L.}~\bibnamefont {Sun}}, \bibinfo {author}
  {\bibfnamefont {M.}~\bibnamefont {Geiger}}, \bibinfo {author} {\bibfnamefont
  {J.~P.}\ \bibnamefont {Mailoa}}, \bibinfo {author} {\bibfnamefont
  {M.}~\bibnamefont {Kornbluth}}, \bibinfo {author} {\bibfnamefont
  {N.}~\bibnamefont {Molinari}}, \bibinfo {author} {\bibfnamefont {T.~E.}\
  \bibnamefont {Smidt}},\ and\ \bibinfo {author} {\bibfnamefont
  {B.}~\bibnamefont {Kozinsky}},\ }\bibfield  {title} {\bibinfo {title}
  {E(3)-equivariant graph neural networks for data-efficient and accurate
  interatomic potentials},\ }\href@noop {} {\bibfield  {journal} {\bibinfo
  {journal} {Nature communications}\ }\textbf {\bibinfo {volume} {13}},\
  \bibinfo {pages} {2453} (\bibinfo {year} {2022})}\BibitemShut {NoStop}%
\bibitem [{\citenamefont {Frank}\ \emph {et~al.}(2024)\citenamefont {Frank},
  \citenamefont {Unke}, \citenamefont {M{\"u}ller},\ and\ \citenamefont
  {Chmiela}}]{frank2024euclidean}%
  \BibitemOpen
  \bibfield  {author} {\bibinfo {author} {\bibfnamefont {J.~T.}\ \bibnamefont
  {Frank}}, \bibinfo {author} {\bibfnamefont {O.~T.}\ \bibnamefont {Unke}},
  \bibinfo {author} {\bibfnamefont {K.-R.}\ \bibnamefont {M{\"u}ller}},\ and\
  \bibinfo {author} {\bibfnamefont {S.}~\bibnamefont {Chmiela}},\ }\bibfield
  {title} {\bibinfo {title} {A euclidean transformer for fast and stable
  machine learned force fields},\ }\href@noop {} {\bibfield  {journal}
  {\bibinfo  {journal} {Nature Communications}\ }\textbf {\bibinfo {volume}
  {15}},\ \bibinfo {pages} {6539} (\bibinfo {year} {2024})}\BibitemShut
  {NoStop}%
\bibitem [{\citenamefont {Kondor}(2025)}]{kondor2025principles}%
  \BibitemOpen
  \bibfield  {author} {\bibinfo {author} {\bibfnamefont {R.}~\bibnamefont
  {Kondor}},\ }\bibfield  {title} {\bibinfo {title} {The principles behind
  equivariant neural networks for physics and chemistry},\ }\href@noop {}
  {\bibfield  {journal} {\bibinfo  {journal} {Proceedings of the National
  Academy of Sciences}\ }\textbf {\bibinfo {volume} {122}},\ \bibinfo {pages}
  {e2415656122} (\bibinfo {year} {2025})}\BibitemShut {NoStop}%
\bibitem [{\citenamefont {Han}\ \emph {et~al.}(2022)\citenamefont {Han},
  \citenamefont {Rong}, \citenamefont {Xu},\ and\ \citenamefont
  {Huang}}]{han2022geometrically}%
  \BibitemOpen
  \bibfield  {author} {\bibinfo {author} {\bibfnamefont {J.}~\bibnamefont
  {Han}}, \bibinfo {author} {\bibfnamefont {Y.}~\bibnamefont {Rong}}, \bibinfo
  {author} {\bibfnamefont {T.}~\bibnamefont {Xu}},\ and\ \bibinfo {author}
  {\bibfnamefont {W.}~\bibnamefont {Huang}},\ }\bibfield  {title} {\bibinfo
  {title} {Geometrically equivariant graph neural networks: A survey},\
  }\href@noop {} {\bibfield  {journal} {\bibinfo  {journal} {arXiv preprint
  arXiv:2202.07230}\ } (\bibinfo {year} {2022})}\BibitemShut {NoStop}%
\bibitem [{\citenamefont {Qi}\ \emph {et~al.}(2017{\natexlab{a}})\citenamefont
  {Qi}, \citenamefont {Yi}, \citenamefont {Su},\ and\ \citenamefont
  {Guibas}}]{qi2017pointnet++}%
  \BibitemOpen
  \bibfield  {author} {\bibinfo {author} {\bibfnamefont {C.~R.}\ \bibnamefont
  {Qi}}, \bibinfo {author} {\bibfnamefont {L.}~\bibnamefont {Yi}}, \bibinfo
  {author} {\bibfnamefont {H.}~\bibnamefont {Su}},\ and\ \bibinfo {author}
  {\bibfnamefont {L.~J.}\ \bibnamefont {Guibas}},\ }\bibfield  {title}
  {\bibinfo {title} {Pointnet++: Deep hierarchical feature learning on point
  sets in a metric space},\ }\href@noop {} {\bibfield  {journal} {\bibinfo
  {journal} {Advances in neural information processing systems}\ }\textbf
  {\bibinfo {volume} {30}} (\bibinfo {year} {2017}{\natexlab{a}})}\BibitemShut
  {NoStop}%
\bibitem [{\citenamefont {Vaswani}\ \emph {et~al.}(2017)\citenamefont
  {Vaswani}, \citenamefont {Shazeer}, \citenamefont {Parmar}, \citenamefont
  {Uszkoreit}, \citenamefont {Jones}, \citenamefont {Gomez}, \citenamefont
  {Kaiser},\ and\ \citenamefont {Polosukhin}}]{vaswani2017attention}%
  \BibitemOpen
  \bibfield  {author} {\bibinfo {author} {\bibfnamefont {A.}~\bibnamefont
  {Vaswani}}, \bibinfo {author} {\bibfnamefont {N.}~\bibnamefont {Shazeer}},
  \bibinfo {author} {\bibfnamefont {N.}~\bibnamefont {Parmar}}, \bibinfo
  {author} {\bibfnamefont {J.}~\bibnamefont {Uszkoreit}}, \bibinfo {author}
  {\bibfnamefont {L.}~\bibnamefont {Jones}}, \bibinfo {author} {\bibfnamefont
  {A.~N.}\ \bibnamefont {Gomez}}, \bibinfo {author} {\bibfnamefont
  {{\L}.}~\bibnamefont {Kaiser}},\ and\ \bibinfo {author} {\bibfnamefont
  {I.}~\bibnamefont {Polosukhin}},\ }\bibfield  {title} {\bibinfo {title}
  {Attention is all you need},\ }\href@noop {} {\bibfield  {journal} {\bibinfo
  {journal} {{Advances in Neural Information Processing Systems}}\ }\textbf
  {\bibinfo {volume} {30}} (\bibinfo {year} {2017})}\BibitemShut {NoStop}%
\bibitem [{\citenamefont {Jeevanjee}(2011)}]{jeevanjee2011introduction}%
  \BibitemOpen
  \bibfield  {author} {\bibinfo {author} {\bibfnamefont {N.}~\bibnamefont
  {Jeevanjee}},\ }\href@noop {} {\emph {\bibinfo {title} {An introduction to
  tensors and group theory for physicists}}}\ (\bibinfo  {publisher}
  {Springer},\ \bibinfo {year} {2011})\BibitemShut {NoStop}%
\bibitem [{\citenamefont {Lipschutz}\ and\ \citenamefont
  {Lipson}(2009)}]{lipschutz2009linear}%
  \BibitemOpen
  \bibfield  {author} {\bibinfo {author} {\bibfnamefont {S.}~\bibnamefont
  {Lipschutz}}\ and\ \bibinfo {author} {\bibfnamefont {M.~L.}\ \bibnamefont
  {Lipson}},\ }\href@noop {} {\emph {\bibinfo {title} {Linear algebra}}}\
  (\bibinfo  {publisher} {MacGraw-Hill},\ \bibinfo {year} {2009})\BibitemShut
  {NoStop}%
\bibitem [{\citenamefont {Arfken}\ and\ \citenamefont
  {Romain}(1967)}]{arfken1967mathematical}%
  \BibitemOpen
  \bibfield  {author} {\bibinfo {author} {\bibfnamefont {G.}~\bibnamefont
  {Arfken}}\ and\ \bibinfo {author} {\bibfnamefont {J.}~\bibnamefont
  {Romain}},\ }\href@noop {} {\bibinfo {title} {Mathematical methods for
  physicists}} (\bibinfo {year} {1967})\BibitemShut {NoStop}%
\bibitem [{\citenamefont {Edmonds}(1996)}]{edmonds1996angular}%
  \BibitemOpen
  \bibfield  {author} {\bibinfo {author} {\bibfnamefont {A.~R.}\ \bibnamefont
  {Edmonds}},\ }\href@noop {} {\emph {\bibinfo {title} {Angular momentum in
  quantum mechanics}}},\ Vol.~\bibinfo {volume} {4}\ (\bibinfo  {publisher}
  {Princeton university press},\ \bibinfo {year} {1996})\BibitemShut {NoStop}%
\bibitem [{\citenamefont {Fulton}\ and\ \citenamefont
  {Harris}(2013)}]{fulton2013representation}%
  \BibitemOpen
  \bibfield  {author} {\bibinfo {author} {\bibfnamefont {W.}~\bibnamefont
  {Fulton}}\ and\ \bibinfo {author} {\bibfnamefont {J.}~\bibnamefont
  {Harris}},\ }\href@noop {} {\emph {\bibinfo {title} {Representation theory: a
  first course}}}\ (\bibinfo  {publisher} {Springer Science \& Business
  Media},\ \bibinfo {year} {2013})\BibitemShut {NoStop}%
\bibitem [{\citenamefont {Pinchon}\ and\ \citenamefont
  {Hoggan}(2007)}]{pinchon2007rotation}%
  \BibitemOpen
  \bibfield  {author} {\bibinfo {author} {\bibfnamefont {D.}~\bibnamefont
  {Pinchon}}\ and\ \bibinfo {author} {\bibfnamefont {P.~E.}\ \bibnamefont
  {Hoggan}},\ }\bibfield  {title} {\bibinfo {title} {Rotation matrices for real
  spherical harmonics: general rotations of atomic orbitals in space-fixed
  axes},\ }\href@noop {} {\bibfield  {journal} {\bibinfo  {journal} {Journal of
  Physics A: Mathematical and Theoretical}\ }\textbf {\bibinfo {volume} {40}},\
  \bibinfo {pages} {1597} (\bibinfo {year} {2007})}\BibitemShut {NoStop}%
\bibitem [{\citenamefont {Musaelian}\ \emph {et~al.}(2023)\citenamefont
  {Musaelian}, \citenamefont {Batzner}, \citenamefont {Johansson},
  \citenamefont {Sun}, \citenamefont {Owen}, \citenamefont {Kornbluth},\ and\
  \citenamefont {Kozinsky}}]{musaelian2023learning}%
  \BibitemOpen
  \bibfield  {author} {\bibinfo {author} {\bibfnamefont {A.}~\bibnamefont
  {Musaelian}}, \bibinfo {author} {\bibfnamefont {S.}~\bibnamefont {Batzner}},
  \bibinfo {author} {\bibfnamefont {A.}~\bibnamefont {Johansson}}, \bibinfo
  {author} {\bibfnamefont {L.}~\bibnamefont {Sun}}, \bibinfo {author}
  {\bibfnamefont {C.~J.}\ \bibnamefont {Owen}}, \bibinfo {author}
  {\bibfnamefont {M.}~\bibnamefont {Kornbluth}},\ and\ \bibinfo {author}
  {\bibfnamefont {B.}~\bibnamefont {Kozinsky}},\ }\bibfield  {title} {\bibinfo
  {title} {Learning local equivariant representations for large-scale atomistic
  dynamics},\ }\href@noop {} {\bibfield  {journal} {\bibinfo  {journal} {Nature
  Communications}\ }\textbf {\bibinfo {volume} {14}},\ \bibinfo {pages} {579}
  (\bibinfo {year} {2023})}\BibitemShut {NoStop}%
\bibitem [{\citenamefont {Liao}\ \emph
  {et~al.}(2024{\natexlab{b}})\citenamefont {Liao}, \citenamefont {Wood},
  \citenamefont {Das},\ and\ \citenamefont {Smidt}}]{liao2024equiformerv2}%
  \BibitemOpen
  \bibfield  {author} {\bibinfo {author} {\bibfnamefont {Y.-L.}\ \bibnamefont
  {Liao}}, \bibinfo {author} {\bibfnamefont {B.~M.}\ \bibnamefont {Wood}},
  \bibinfo {author} {\bibfnamefont {A.}~\bibnamefont {Das}},\ and\ \bibinfo
  {author} {\bibfnamefont {T.}~\bibnamefont {Smidt}},\ }\bibfield  {title}
  {\bibinfo {title} {Equiformerv2: Improved equivariant transformer for scaling
  to higher-degree representations},\ }in\ \href@noop {} {\emph {\bibinfo
  {booktitle} {The Twelfth International Conference on Learning
  Representations}}}\ (\bibinfo {year} {2024})\BibitemShut {NoStop}%
\bibitem [{\citenamefont {Schulten}\ and\ \citenamefont
  {Gordon}(1975)}]{schulten1975exact}%
  \BibitemOpen
  \bibfield  {author} {\bibinfo {author} {\bibfnamefont {K.}~\bibnamefont
  {Schulten}}\ and\ \bibinfo {author} {\bibfnamefont {R.~G.}\ \bibnamefont
  {Gordon}},\ }\bibfield  {title} {\bibinfo {title} {Exact recursive evaluation
  of 3 j-and 6 j-coefficients for quantum-mechanical coupling of angular
  momenta},\ }\href@noop {} {\bibfield  {journal} {\bibinfo  {journal} {Journal
  of Mathematical Physics}\ }\textbf {\bibinfo {volume} {16}},\ \bibinfo
  {pages} {1961} (\bibinfo {year} {1975})}\BibitemShut {NoStop}%
\bibitem [{\citenamefont {Johansson}\ and\ \citenamefont
  {Forss{\'e}n}(2016)}]{johansson2016fast}%
  \BibitemOpen
  \bibfield  {author} {\bibinfo {author} {\bibfnamefont {H.~T.}\ \bibnamefont
  {Johansson}}\ and\ \bibinfo {author} {\bibfnamefont {C.}~\bibnamefont
  {Forss{\'e}n}},\ }\bibfield  {title} {\bibinfo {title} {Fast and accurate
  evaluation of wigner 3 j, 6 j, and 9 j symbols using prime factorization and
  multiword integer arithmetic},\ }\href@noop {} {\bibfield  {journal}
  {\bibinfo  {journal} {SIAM Journal on Scientific Computing}\ }\textbf
  {\bibinfo {volume} {38}},\ \bibinfo {pages} {A376} (\bibinfo {year}
  {2016})}\BibitemShut {NoStop}%
\bibitem [{\citenamefont {Passaro}\ and\ \citenamefont
  {Zitnick}(2023)}]{passaro2023reducing}%
  \BibitemOpen
  \bibfield  {author} {\bibinfo {author} {\bibfnamefont {S.}~\bibnamefont
  {Passaro}}\ and\ \bibinfo {author} {\bibfnamefont {C.~L.}\ \bibnamefont
  {Zitnick}},\ }\bibfield  {title} {\bibinfo {title} {Reducing {SO(3)}
  convolutions to {SO(2)} for efficient equivariant gnns},\ }in\ \href@noop {}
  {\emph {\bibinfo {booktitle} {International Conference on Machine
  Learning}}}\ (\bibinfo {organization} {PMLR},\ \bibinfo {year} {2023})\ pp.\
  \bibinfo {pages} {27420--27438}\BibitemShut {NoStop}%
\bibitem [{\citenamefont {Fu}\ \emph {et~al.}(2025)\citenamefont {Fu},
  \citenamefont {Wood}, \citenamefont {Barroso-Luque}, \citenamefont {Levine},
  \citenamefont {Gao}, \citenamefont {Dzamba},\ and\ \citenamefont
  {Zitnick}}]{fu2025learning}%
  \BibitemOpen
  \bibfield  {author} {\bibinfo {author} {\bibfnamefont {X.}~\bibnamefont
  {Fu}}, \bibinfo {author} {\bibfnamefont {B.~M.}\ \bibnamefont {Wood}},
  \bibinfo {author} {\bibfnamefont {L.}~\bibnamefont {Barroso-Luque}}, \bibinfo
  {author} {\bibfnamefont {D.~S.}\ \bibnamefont {Levine}}, \bibinfo {author}
  {\bibfnamefont {M.}~\bibnamefont {Gao}}, \bibinfo {author} {\bibfnamefont
  {M.}~\bibnamefont {Dzamba}},\ and\ \bibinfo {author} {\bibfnamefont {C.~L.}\
  \bibnamefont {Zitnick}},\ }\bibfield  {title} {\bibinfo {title} {Learning
  smooth and expressive interatomic potentials for physical property
  prediction},\ }\href@noop {} {\bibfield  {journal} {\bibinfo  {journal}
  {arXiv preprint arXiv:2502.12147}\ } (\bibinfo {year} {2025})}\BibitemShut
  {NoStop}%
\bibitem [{\citenamefont {Abramson}\ \emph {et~al.}(2024)\citenamefont
  {Abramson}, \citenamefont {Adler}, \citenamefont {Dunger}, \citenamefont
  {Evans}, \citenamefont {Green}, \citenamefont {Pritzel}, \citenamefont
  {Ronneberger}, \citenamefont {Willmore}, \citenamefont {Ballard},
  \citenamefont {Bambrick} \emph {et~al.}}]{abramson2024accurate}%
  \BibitemOpen
  \bibfield  {author} {\bibinfo {author} {\bibfnamefont {J.}~\bibnamefont
  {Abramson}}, \bibinfo {author} {\bibfnamefont {J.}~\bibnamefont {Adler}},
  \bibinfo {author} {\bibfnamefont {J.}~\bibnamefont {Dunger}}, \bibinfo
  {author} {\bibfnamefont {R.}~\bibnamefont {Evans}}, \bibinfo {author}
  {\bibfnamefont {T.}~\bibnamefont {Green}}, \bibinfo {author} {\bibfnamefont
  {A.}~\bibnamefont {Pritzel}}, \bibinfo {author} {\bibfnamefont
  {O.}~\bibnamefont {Ronneberger}}, \bibinfo {author} {\bibfnamefont
  {L.}~\bibnamefont {Willmore}}, \bibinfo {author} {\bibfnamefont {A.~J.}\
  \bibnamefont {Ballard}}, \bibinfo {author} {\bibfnamefont {J.}~\bibnamefont
  {Bambrick}}, \emph {et~al.},\ }\bibfield  {title} {\bibinfo {title} {Accurate
  structure prediction of biomolecular interactions with alphafold 3},\
  }\href@noop {} {\bibfield  {journal} {\bibinfo  {journal} {Nature}\ }\textbf
  {\bibinfo {volume} {630}},\ \bibinfo {pages} {493} (\bibinfo {year}
  {2024})}\BibitemShut {NoStop}%
\bibitem [{\citenamefont {Sch{\"u}tt}\ \emph {et~al.}(2018)\citenamefont
  {Sch{\"u}tt}, \citenamefont {Sauceda}, \citenamefont {Kindermans},
  \citenamefont {Tkatchenko},\ and\ \citenamefont
  {M{\"u}ller}}]{schutt2018schnet}%
  \BibitemOpen
  \bibfield  {author} {\bibinfo {author} {\bibfnamefont {K.~T.}\ \bibnamefont
  {Sch{\"u}tt}}, \bibinfo {author} {\bibfnamefont {H.~E.}\ \bibnamefont
  {Sauceda}}, \bibinfo {author} {\bibfnamefont {P.-J.}\ \bibnamefont
  {Kindermans}}, \bibinfo {author} {\bibfnamefont {A.}~\bibnamefont
  {Tkatchenko}},\ and\ \bibinfo {author} {\bibfnamefont {K.-R.}\ \bibnamefont
  {M{\"u}ller}},\ }\bibfield  {title} {\bibinfo {title} {Schnet--a deep
  learning architecture for molecules and materials},\ }\href@noop {}
  {\bibfield  {journal} {\bibinfo  {journal} {The Journal of Chemical Physics}\
  }\textbf {\bibinfo {volume} {148}} (\bibinfo {year} {2018})}\BibitemShut
  {NoStop}%
\bibitem [{\citenamefont {Gasteiger}\ \emph {et~al.}(2020)\citenamefont
  {Gasteiger}, \citenamefont {Groß},\ and\ \citenamefont
  {Günnemann}}]{gasteiger2020directional}%
  \BibitemOpen
  \bibfield  {author} {\bibinfo {author} {\bibfnamefont {J.}~\bibnamefont
  {Gasteiger}}, \bibinfo {author} {\bibfnamefont {J.}~\bibnamefont {Groß}},\
  and\ \bibinfo {author} {\bibfnamefont {S.}~\bibnamefont {Günnemann}},\
  }\bibfield  {title} {\bibinfo {title} {Directional message passing for
  molecular graphs},\ }in\ \href {https://openreview.net/forum?id=B1eWbxStPH}
  {\emph {\bibinfo {booktitle} {International Conference on Learning
  Representations}}}\ (\bibinfo {year} {2020})\BibitemShut {NoStop}%
\bibitem [{\citenamefont {Li}\ \emph {et~al.}(2021{\natexlab{a}})\citenamefont
  {Li}, \citenamefont {Li}, \citenamefont {Chen}, \citenamefont {Fu},
  \citenamefont {Cohen-Or},\ and\ \citenamefont {Heng}}]{li2021rotation}%
  \BibitemOpen
  \bibfield  {author} {\bibinfo {author} {\bibfnamefont {X.}~\bibnamefont
  {Li}}, \bibinfo {author} {\bibfnamefont {R.}~\bibnamefont {Li}}, \bibinfo
  {author} {\bibfnamefont {G.}~\bibnamefont {Chen}}, \bibinfo {author}
  {\bibfnamefont {C.-W.}\ \bibnamefont {Fu}}, \bibinfo {author} {\bibfnamefont
  {D.}~\bibnamefont {Cohen-Or}},\ and\ \bibinfo {author} {\bibfnamefont
  {P.-A.}\ \bibnamefont {Heng}},\ }\bibfield  {title} {\bibinfo {title} {A
  rotation-invariant framework for deep point cloud analysis},\ }\href@noop {}
  {\bibfield  {journal} {\bibinfo  {journal} {IEEE transactions on
  visualization and computer graphics}\ }\textbf {\bibinfo {volume} {28}},\
  \bibinfo {pages} {4503} (\bibinfo {year} {2021}{\natexlab{a}})}\BibitemShut
  {NoStop}%
\bibitem [{\citenamefont {Gasteiger}\ \emph {et~al.}(2021)\citenamefont
  {Gasteiger}, \citenamefont {Becker},\ and\ \citenamefont
  {G{\"u}nnemann}}]{gasteiger2021gemnet}%
  \BibitemOpen
  \bibfield  {author} {\bibinfo {author} {\bibfnamefont {J.}~\bibnamefont
  {Gasteiger}}, \bibinfo {author} {\bibfnamefont {F.}~\bibnamefont {Becker}},\
  and\ \bibinfo {author} {\bibfnamefont {S.}~\bibnamefont {G{\"u}nnemann}},\
  }\bibfield  {title} {\bibinfo {title} {Gemnet: Universal directional graph
  neural networks for molecules},\ }\href@noop {} {\bibfield  {journal}
  {\bibinfo  {journal} {Advances in neural information processing systems}\
  }\textbf {\bibinfo {volume} {34}},\ \bibinfo {pages} {6790} (\bibinfo {year}
  {2021})}\BibitemShut {NoStop}%
\bibitem [{\citenamefont {Liu}\ \emph {et~al.}(2022)\citenamefont {Liu},
  \citenamefont {Wang}, \citenamefont {Liu}, \citenamefont {Lin}, \citenamefont
  {Zhang}, \citenamefont {Oztekin},\ and\ \citenamefont
  {Ji}}]{liu2022spherical}%
  \BibitemOpen
  \bibfield  {author} {\bibinfo {author} {\bibfnamefont {Y.}~\bibnamefont
  {Liu}}, \bibinfo {author} {\bibfnamefont {L.}~\bibnamefont {Wang}}, \bibinfo
  {author} {\bibfnamefont {M.}~\bibnamefont {Liu}}, \bibinfo {author}
  {\bibfnamefont {Y.}~\bibnamefont {Lin}}, \bibinfo {author} {\bibfnamefont
  {X.}~\bibnamefont {Zhang}}, \bibinfo {author} {\bibfnamefont
  {B.}~\bibnamefont {Oztekin}},\ and\ \bibinfo {author} {\bibfnamefont
  {S.}~\bibnamefont {Ji}},\ }\bibfield  {title} {\bibinfo {title} {Spherical
  message passing for 3d molecular graphs},\ }in\ \href@noop {} {\emph
  {\bibinfo {booktitle} {International conference on learning
  representations}}}\ (\bibinfo {year} {2022})\BibitemShut {NoStop}%
\bibitem [{\citenamefont {Fuchs}\ \emph {et~al.}(2020)\citenamefont {Fuchs},
  \citenamefont {Worrall}, \citenamefont {Fischer},\ and\ \citenamefont
  {Welling}}]{fuchs2020se}%
  \BibitemOpen
  \bibfield  {author} {\bibinfo {author} {\bibfnamefont {F.}~\bibnamefont
  {Fuchs}}, \bibinfo {author} {\bibfnamefont {D.}~\bibnamefont {Worrall}},
  \bibinfo {author} {\bibfnamefont {V.}~\bibnamefont {Fischer}},\ and\ \bibinfo
  {author} {\bibfnamefont {M.}~\bibnamefont {Welling}},\ }\bibfield  {title}
  {\bibinfo {title} {{SE(3)}-transformers: {3D} roto-translation equivariant
  attention networks},\ }\href@noop {} {\bibfield  {journal} {\bibinfo
  {journal} {Advances in neural information processing systems}\ }\textbf
  {\bibinfo {volume} {33}},\ \bibinfo {pages} {1970} (\bibinfo {year}
  {2020})}\BibitemShut {NoStop}%
\bibitem [{\citenamefont {Brandstetter}\ \emph {et~al.}(2022)\citenamefont
  {Brandstetter}, \citenamefont {Hesselink}, \citenamefont {van~der Pol},
  \citenamefont {Bekkers},\ and\ \citenamefont
  {Welling}}]{brandstetter2021geometric}%
  \BibitemOpen
  \bibfield  {author} {\bibinfo {author} {\bibfnamefont {J.}~\bibnamefont
  {Brandstetter}}, \bibinfo {author} {\bibfnamefont {R.}~\bibnamefont
  {Hesselink}}, \bibinfo {author} {\bibfnamefont {E.}~\bibnamefont {van~der
  Pol}}, \bibinfo {author} {\bibfnamefont {E.~J.}\ \bibnamefont {Bekkers}},\
  and\ \bibinfo {author} {\bibfnamefont {M.}~\bibnamefont {Welling}},\
  }\bibfield  {title} {\bibinfo {title} {Geometric and physical quantities
  improve {E(3)} equivariant message passing},\ }in\ \href
  {https://openreview.net/forum?id=_xwr8gOBeV1} {\emph {\bibinfo {booktitle}
  {International Conference on Learning Representations}}}\ (\bibinfo {year}
  {2022})\BibitemShut {NoStop}%
\bibitem [{\citenamefont {Zitnick}\ \emph {et~al.}(2022)\citenamefont
  {Zitnick}, \citenamefont {Das}, \citenamefont {Kolluru}, \citenamefont {Lan},
  \citenamefont {Shuaibi}, \citenamefont {Sriram}, \citenamefont {Ulissi},\
  and\ \citenamefont {Wood}}]{zitnick2022spherical}%
  \BibitemOpen
  \bibfield  {author} {\bibinfo {author} {\bibfnamefont {L.}~\bibnamefont
  {Zitnick}}, \bibinfo {author} {\bibfnamefont {A.}~\bibnamefont {Das}},
  \bibinfo {author} {\bibfnamefont {A.}~\bibnamefont {Kolluru}}, \bibinfo
  {author} {\bibfnamefont {J.}~\bibnamefont {Lan}}, \bibinfo {author}
  {\bibfnamefont {M.}~\bibnamefont {Shuaibi}}, \bibinfo {author} {\bibfnamefont
  {A.}~\bibnamefont {Sriram}}, \bibinfo {author} {\bibfnamefont
  {Z.}~\bibnamefont {Ulissi}},\ and\ \bibinfo {author} {\bibfnamefont
  {B.}~\bibnamefont {Wood}},\ }\bibfield  {title} {\bibinfo {title} {Spherical
  channels for modeling atomic interactions},\ }\href@noop {} {\bibfield
  {journal} {\bibinfo  {journal} {Advances in Neural Information Processing
  Systems}\ }\textbf {\bibinfo {volume} {35}},\ \bibinfo {pages} {8054}
  (\bibinfo {year} {2022})}\BibitemShut {NoStop}%
\bibitem [{\citenamefont {Bekkers}\ \emph {et~al.}(2018)\citenamefont
  {Bekkers}, \citenamefont {Lafarge}, \citenamefont {Veta}, \citenamefont
  {Eppenhof}, \citenamefont {Pluim},\ and\ \citenamefont
  {Duits}}]{bekkers2018roto}%
  \BibitemOpen
  \bibfield  {author} {\bibinfo {author} {\bibfnamefont {E.~J.}\ \bibnamefont
  {Bekkers}}, \bibinfo {author} {\bibfnamefont {M.~W.}\ \bibnamefont
  {Lafarge}}, \bibinfo {author} {\bibfnamefont {M.}~\bibnamefont {Veta}},
  \bibinfo {author} {\bibfnamefont {K.~A.}\ \bibnamefont {Eppenhof}}, \bibinfo
  {author} {\bibfnamefont {J.~P.}\ \bibnamefont {Pluim}},\ and\ \bibinfo
  {author} {\bibfnamefont {R.}~\bibnamefont {Duits}},\ }\bibfield  {title}
  {\bibinfo {title} {Roto-translation covariant convolutional networks for
  medical image analysis},\ }in\ \href@noop {} {\emph {\bibinfo {booktitle}
  {International conference on medical image computing and computer-assisted
  intervention}}}\ (\bibinfo {organization} {Springer},\ \bibinfo {year}
  {2018})\ pp.\ \bibinfo {pages} {440--448}\BibitemShut {NoStop}%
\bibitem [{\citenamefont {Bekkers}\ \emph {et~al.}(2023)\citenamefont
  {Bekkers}, \citenamefont {Vadgama}, \citenamefont {Hesselink}, \citenamefont
  {Van~der Linden},\ and\ \citenamefont {Romero}}]{bekkers2023fast}%
  \BibitemOpen
  \bibfield  {author} {\bibinfo {author} {\bibfnamefont {E.~J.}\ \bibnamefont
  {Bekkers}}, \bibinfo {author} {\bibfnamefont {S.}~\bibnamefont {Vadgama}},
  \bibinfo {author} {\bibfnamefont {R.~D.}\ \bibnamefont {Hesselink}}, \bibinfo
  {author} {\bibfnamefont {P.~A.}\ \bibnamefont {Van~der Linden}},\ and\
  \bibinfo {author} {\bibfnamefont {D.~W.}\ \bibnamefont {Romero}},\ }\bibfield
   {title} {\bibinfo {title} {Fast, expressive se $(n) $ equivariant networks
  through weight-sharing in position-orientation space},\ }\href@noop {}
  {\bibfield  {journal} {\bibinfo  {journal} {arXiv preprint arXiv:2310.02970}\
  } (\bibinfo {year} {2023})}\BibitemShut {NoStop}%
\bibitem [{\citenamefont {Worrall}\ and\ \citenamefont
  {Brostow}(2018)}]{worrall2018cubenet}%
  \BibitemOpen
  \bibfield  {author} {\bibinfo {author} {\bibfnamefont {D.}~\bibnamefont
  {Worrall}}\ and\ \bibinfo {author} {\bibfnamefont {G.}~\bibnamefont
  {Brostow}},\ }\bibfield  {title} {\bibinfo {title} {Cubenet: Equivariance to
  3d rotation and translation},\ }in\ \href@noop {} {\emph {\bibinfo
  {booktitle} {Proceedings of the European Conference on Computer Vision
  (ECCV)}}}\ (\bibinfo {year} {2018})\ pp.\ \bibinfo {pages}
  {567--584}\BibitemShut {NoStop}%
\bibitem [{\citenamefont {Beisert}(2015)}]{beisert2015symmetries}%
  \BibitemOpen
  \bibfield  {author} {\bibinfo {author} {\bibfnamefont {N.}~\bibnamefont
  {Beisert}},\ }\bibfield  {title} {\bibinfo {title} {Symmetries in physics},\
  }\href@noop {} {\bibfield  {journal} {\bibinfo  {journal} {ETH Zurich}\ }
  (\bibinfo {year} {2015})}\BibitemShut {NoStop}%
\bibitem [{\citenamefont {Islam}\ \emph {et~al.}(2025)\citenamefont {Islam},
  \citenamefont {Anand}, \citenamefont {Wessels}, \citenamefont {de~Kruiff},
  \citenamefont {Kuipers}, \citenamefont {Ying}, \citenamefont {S{\'a}nchez},
  \citenamefont {Vadgama}, \citenamefont {B{\"o}kman},\ and\ \citenamefont
  {Bekkers}}]{islam2025platonic}%
  \BibitemOpen
  \bibfield  {author} {\bibinfo {author} {\bibfnamefont {M.~M.}\ \bibnamefont
  {Islam}}, \bibinfo {author} {\bibfnamefont {R.}~\bibnamefont {Anand}},
  \bibinfo {author} {\bibfnamefont {D.~R.}\ \bibnamefont {Wessels}}, \bibinfo
  {author} {\bibfnamefont {F.}~\bibnamefont {de~Kruiff}}, \bibinfo {author}
  {\bibfnamefont {T.~P.}\ \bibnamefont {Kuipers}}, \bibinfo {author}
  {\bibfnamefont {R.}~\bibnamefont {Ying}}, \bibinfo {author} {\bibfnamefont
  {C.~I.}\ \bibnamefont {S{\'a}nchez}}, \bibinfo {author} {\bibfnamefont
  {S.}~\bibnamefont {Vadgama}}, \bibinfo {author} {\bibfnamefont
  {G.}~\bibnamefont {B{\"o}kman}},\ and\ \bibinfo {author} {\bibfnamefont
  {E.~J.}\ \bibnamefont {Bekkers}},\ }\bibfield  {title} {\bibinfo {title}
  {Platonic transformers: A solid choice for equivariance},\ }\href@noop {}
  {\bibfield  {journal} {\bibinfo  {journal} {arXiv preprint arXiv:2510.03511}\
  } (\bibinfo {year} {2025})}\BibitemShut {NoStop}%
\bibitem [{\citenamefont {Thomas}\ \emph {et~al.}(2018)\citenamefont {Thomas},
  \citenamefont {Smidt}, \citenamefont {Kearnes}, \citenamefont {Yang},
  \citenamefont {Li}, \citenamefont {Kohlhoff},\ and\ \citenamefont
  {Riley}}]{thomas2018tensor}%
  \BibitemOpen
  \bibfield  {author} {\bibinfo {author} {\bibfnamefont {N.}~\bibnamefont
  {Thomas}}, \bibinfo {author} {\bibfnamefont {T.}~\bibnamefont {Smidt}},
  \bibinfo {author} {\bibfnamefont {S.}~\bibnamefont {Kearnes}}, \bibinfo
  {author} {\bibfnamefont {L.}~\bibnamefont {Yang}}, \bibinfo {author}
  {\bibfnamefont {L.}~\bibnamefont {Li}}, \bibinfo {author} {\bibfnamefont
  {K.}~\bibnamefont {Kohlhoff}},\ and\ \bibinfo {author} {\bibfnamefont
  {P.}~\bibnamefont {Riley}},\ }\href@noop {} {\bibinfo {title} {Tensor field
  networks: Rotation- and translation-equivariant neural networks for 3d point
  clouds}},\ \bibinfo {howpublished} {arXiv:1802.08219 [cs.LG]} (\bibinfo
  {year} {2018}),\ \Eprint {https://arxiv.org/abs/arXiv:1802.08219}
  {arXiv:1802.08219 [cs.LG]} \BibitemShut {NoStop}%
\bibitem [{\citenamefont {Weiler}\ \emph {et~al.}(2018)\citenamefont {Weiler},
  \citenamefont {Geiger}, \citenamefont {Welling}, \citenamefont {Boomsma},\
  and\ \citenamefont {Cohen}}]{weiler20183d}%
  \BibitemOpen
  \bibfield  {author} {\bibinfo {author} {\bibfnamefont {M.}~\bibnamefont
  {Weiler}}, \bibinfo {author} {\bibfnamefont {M.}~\bibnamefont {Geiger}},
  \bibinfo {author} {\bibfnamefont {M.}~\bibnamefont {Welling}}, \bibinfo
  {author} {\bibfnamefont {W.}~\bibnamefont {Boomsma}},\ and\ \bibinfo {author}
  {\bibfnamefont {T.~S.}\ \bibnamefont {Cohen}},\ }\bibfield  {title} {\bibinfo
  {title} {{3D} steerable cnns: Learning rotationally equivariant features in
  volumetric data},\ }\href@noop {} {\bibfield  {journal} {\bibinfo  {journal}
  {Advances in Neural Information Processing Systems}\ }\textbf {\bibinfo
  {volume} {31}} (\bibinfo {year} {2018})}\BibitemShut {NoStop}%
\bibitem [{\citenamefont {Weiler}\ and\ \citenamefont {Cesa}(2019)}]{e2cnn}%
  \BibitemOpen
  \bibfield  {author} {\bibinfo {author} {\bibfnamefont {M.}~\bibnamefont
  {Weiler}}\ and\ \bibinfo {author} {\bibfnamefont {G.}~\bibnamefont {Cesa}},\
  }\bibfield  {title} {\bibinfo {title} {{General E(2)-Equivariant Steerable
  CNNs}},\ }in\ \href@noop {} {\emph {\bibinfo {booktitle} {Conference on
  Neural Information Processing Systems (NeurIPS)}}}\ (\bibinfo {year}
  {2019})\BibitemShut {NoStop}%
\bibitem [{\citenamefont {Brehmer}\ \emph {et~al.}(2023)\citenamefont
  {Brehmer}, \citenamefont {de~Haan}, \citenamefont {Behrends},\ and\
  \citenamefont {Cohen}}]{brehmer2023geometric}%
  \BibitemOpen
  \bibfield  {author} {\bibinfo {author} {\bibfnamefont {J.}~\bibnamefont
  {Brehmer}}, \bibinfo {author} {\bibfnamefont {P.}~\bibnamefont {de~Haan}},
  \bibinfo {author} {\bibfnamefont {S.}~\bibnamefont {Behrends}},\ and\
  \bibinfo {author} {\bibfnamefont {T.~S.}\ \bibnamefont {Cohen}},\ }\bibfield
  {title} {\bibinfo {title} {Geometric algebra transformer},\ }in\ \href
  {https://proceedings.neurips.cc/paper_files/paper/2023/file/6f6dd92b03ff9be7468a6104611c9187-Paper-Conference.pdf}
  {\emph {\bibinfo {booktitle} {Advances in Neural Information Processing
  Systems}}},\ Vol.~\bibinfo {volume} {36}\ (\bibinfo {year} {2023})\ pp.\
  \bibinfo {pages} {35472--35496}\BibitemShut {NoStop}%
\bibitem [{\citenamefont {Ruhe}\ \emph {et~al.}(2023)\citenamefont {Ruhe},
  \citenamefont {Brandstetter},\ and\ \citenamefont
  {Forr\'{e}}}]{ruhe2024clifford}%
  \BibitemOpen
  \bibfield  {author} {\bibinfo {author} {\bibfnamefont {D.}~\bibnamefont
  {Ruhe}}, \bibinfo {author} {\bibfnamefont {J.}~\bibnamefont {Brandstetter}},\
  and\ \bibinfo {author} {\bibfnamefont {P.}~\bibnamefont {Forr\'{e}}},\
  }\bibfield  {title} {\bibinfo {title} {Clifford group equivariant neural
  networks},\ }in\ \href
  {https://proceedings.neurips.cc/paper_files/paper/2023/file/c6e0125e14ea3d1a3de3c33fd2d49fc4-Paper-Conference.pdf}
  {\emph {\bibinfo {booktitle} {Advances in Neural Information Processing
  Systems}}},\ Vol.~\bibinfo {volume} {36}\ (\bibinfo {year} {2023})\ pp.\
  \bibinfo {pages} {62922--62990}\BibitemShut {NoStop}%
\bibitem [{\citenamefont {Deng}\ \emph {et~al.}(2021)\citenamefont {Deng},
  \citenamefont {Litany}, \citenamefont {Duan}, \citenamefont {Poulenard},
  \citenamefont {Tagliasacchi},\ and\ \citenamefont {Guibas}}]{deng2021vector}%
  \BibitemOpen
  \bibfield  {author} {\bibinfo {author} {\bibfnamefont {C.}~\bibnamefont
  {Deng}}, \bibinfo {author} {\bibfnamefont {O.}~\bibnamefont {Litany}},
  \bibinfo {author} {\bibfnamefont {Y.}~\bibnamefont {Duan}}, \bibinfo {author}
  {\bibfnamefont {A.}~\bibnamefont {Poulenard}}, \bibinfo {author}
  {\bibfnamefont {A.}~\bibnamefont {Tagliasacchi}},\ and\ \bibinfo {author}
  {\bibfnamefont {L.~J.}\ \bibnamefont {Guibas}},\ }\bibfield  {title}
  {\bibinfo {title} {Vector neurons: A general framework for
  {SO(3)}-equivariant networks},\ }in\ \href@noop {} {\emph {\bibinfo
  {booktitle} {Proceedings of the IEEE/CVF International Conference on Computer
  Vision}}}\ (\bibinfo {year} {2021})\ pp.\ \bibinfo {pages}
  {12200--12209}\BibitemShut {NoStop}%
\bibitem [{\citenamefont {Satorras}\ \emph {et~al.}(2021)\citenamefont
  {Satorras}, \citenamefont {Hoogeboom},\ and\ \citenamefont
  {Welling}}]{satorras2021n}%
  \BibitemOpen
  \bibfield  {author} {\bibinfo {author} {\bibfnamefont {V.~G.}\ \bibnamefont
  {Satorras}}, \bibinfo {author} {\bibfnamefont {E.}~\bibnamefont
  {Hoogeboom}},\ and\ \bibinfo {author} {\bibfnamefont {M.}~\bibnamefont
  {Welling}},\ }\bibfield  {title} {\bibinfo {title} {E (n) equivariant graph
  neural networks},\ }in\ \href@noop {} {\emph {\bibinfo {booktitle}
  {International conference on machine learning}}}\ (\bibinfo {organization}
  {PMLR},\ \bibinfo {year} {2021})\ pp.\ \bibinfo {pages}
  {9323--9332}\BibitemShut {NoStop}%
\bibitem [{\citenamefont {Cheng}(2024)}]{cheng2024cartesian}%
  \BibitemOpen
  \bibfield  {author} {\bibinfo {author} {\bibfnamefont {B.}~\bibnamefont
  {Cheng}},\ }\bibfield  {title} {\bibinfo {title} {Cartesian atomic cluster
  expansion for machine learning interatomic potentials},\ }\href@noop {}
  {\bibfield  {journal} {\bibinfo  {journal} {npj Computational Materials}\
  }\textbf {\bibinfo {volume} {10}},\ \bibinfo {pages} {157} (\bibinfo {year}
  {2024})}\BibitemShut {NoStop}%
\bibitem [{\citenamefont {Simeon}\ and\ \citenamefont
  {De~Fabritiis}(2023)}]{simeon2024tensornet}%
  \BibitemOpen
  \bibfield  {author} {\bibinfo {author} {\bibfnamefont {G.}~\bibnamefont
  {Simeon}}\ and\ \bibinfo {author} {\bibfnamefont {G.}~\bibnamefont
  {De~Fabritiis}},\ }\bibfield  {title} {\bibinfo {title} {{TensorNet}:
  Cartesian tensor representations for efficient learning of molecular
  potentials},\ }in\ \href@noop {} {\emph {\bibinfo {booktitle} {Advances in
  Neural Information Processing Systems}}},\ Vol.~\bibinfo {volume} {36}\
  (\bibinfo {year} {2023})\ pp.\ \bibinfo {pages} {37334--37353}\BibitemShut
  {NoStop}%
\bibitem [{\citenamefont {Liu}\ \emph {et~al.}(2019)\citenamefont {Liu},
  \citenamefont {Fan}, \citenamefont {Xiang},\ and\ \citenamefont
  {Pan}}]{liu2019relation}%
  \BibitemOpen
  \bibfield  {author} {\bibinfo {author} {\bibfnamefont {Y.}~\bibnamefont
  {Liu}}, \bibinfo {author} {\bibfnamefont {B.}~\bibnamefont {Fan}}, \bibinfo
  {author} {\bibfnamefont {S.}~\bibnamefont {Xiang}},\ and\ \bibinfo {author}
  {\bibfnamefont {C.}~\bibnamefont {Pan}},\ }\bibfield  {title} {\bibinfo
  {title} {Relation-shape convolutional neural network for point cloud
  analysis},\ }in\ \href@noop {} {\emph {\bibinfo {booktitle} {Proceedings of
  the IEEE/CVF conference on computer vision and pattern recognition}}}\
  (\bibinfo {year} {2019})\ pp.\ \bibinfo {pages} {8895--8904}\BibitemShut
  {NoStop}%
\bibitem [{\citenamefont {Wang}\ \emph {et~al.}(2019)\citenamefont {Wang},
  \citenamefont {Sun}, \citenamefont {Liu}, \citenamefont {Sarma},
  \citenamefont {Bronstein},\ and\ \citenamefont {Solomon}}]{wang2019dynamic}%
  \BibitemOpen
  \bibfield  {author} {\bibinfo {author} {\bibfnamefont {Y.}~\bibnamefont
  {Wang}}, \bibinfo {author} {\bibfnamefont {Y.}~\bibnamefont {Sun}}, \bibinfo
  {author} {\bibfnamefont {Z.}~\bibnamefont {Liu}}, \bibinfo {author}
  {\bibfnamefont {S.~E.}\ \bibnamefont {Sarma}}, \bibinfo {author}
  {\bibfnamefont {M.~M.}\ \bibnamefont {Bronstein}},\ and\ \bibinfo {author}
  {\bibfnamefont {J.~M.}\ \bibnamefont {Solomon}},\ }\bibfield  {title}
  {\bibinfo {title} {Dynamic graph cnn for learning on point clouds},\
  }\href@noop {} {\bibfield  {journal} {\bibinfo  {journal} {ACM Transactions
  on Graphics (tog)}\ }\textbf {\bibinfo {volume} {38}},\ \bibinfo {pages} {1}
  (\bibinfo {year} {2019})}\BibitemShut {NoStop}%
\bibitem [{\citenamefont {Puny}\ \emph {et~al.}(2022)\citenamefont {Puny},
  \citenamefont {Atzmon}, \citenamefont {Smith}, \citenamefont {Misra},
  \citenamefont {Grover}, \citenamefont {Ben-Hamu},\ and\ \citenamefont
  {Lipman}}]{puny2021frame}%
  \BibitemOpen
  \bibfield  {author} {\bibinfo {author} {\bibfnamefont {O.}~\bibnamefont
  {Puny}}, \bibinfo {author} {\bibfnamefont {M.}~\bibnamefont {Atzmon}},
  \bibinfo {author} {\bibfnamefont {E.~J.}\ \bibnamefont {Smith}}, \bibinfo
  {author} {\bibfnamefont {I.}~\bibnamefont {Misra}}, \bibinfo {author}
  {\bibfnamefont {A.}~\bibnamefont {Grover}}, \bibinfo {author} {\bibfnamefont
  {H.}~\bibnamefont {Ben-Hamu}},\ and\ \bibinfo {author} {\bibfnamefont
  {Y.}~\bibnamefont {Lipman}},\ }\bibfield  {title} {\bibinfo {title} {Frame
  averaging for invariant and equivariant network design},\ }in\ \href
  {https://openreview.net/forum?id=zIUyj55nXR} {\emph {\bibinfo {booktitle}
  {International Conference on Learning Representations}}}\ (\bibinfo {year}
  {2022})\BibitemShut {NoStop}%
\bibitem [{\citenamefont {Pozdnyakov}\ and\ \citenamefont
  {Ceriotti}(2023)}]{pozdnyakov2024smooth}%
  \BibitemOpen
  \bibfield  {author} {\bibinfo {author} {\bibfnamefont {S.}~\bibnamefont
  {Pozdnyakov}}\ and\ \bibinfo {author} {\bibfnamefont {M.}~\bibnamefont
  {Ceriotti}},\ }\bibfield  {title} {\bibinfo {title} {Smooth, exact rotational
  symmetrization for deep learning on point clouds},\ }in\ \href
  {https://proceedings.neurips.cc/paper_files/paper/2023/file/fb4a7e3522363907b26a86cc5be627ac-Paper-Conference.pdf}
  {\emph {\bibinfo {booktitle} {Advances in Neural Information Processing
  Systems}}},\ Vol.~\bibinfo {volume} {36}\ (\bibinfo {year} {2023})\ pp.\
  \bibinfo {pages} {79469--79501}\BibitemShut {NoStop}%
\bibitem [{\citenamefont {Kaba}\ \emph {et~al.}(2023)\citenamefont {Kaba},
  \citenamefont {Mondal}, \citenamefont {Zhang}, \citenamefont {Bengio},\ and\
  \citenamefont {Ravanbakhsh}}]{kaba2023equivariance}%
  \BibitemOpen
  \bibfield  {author} {\bibinfo {author} {\bibfnamefont {S.-O.}\ \bibnamefont
  {Kaba}}, \bibinfo {author} {\bibfnamefont {A.~K.}\ \bibnamefont {Mondal}},
  \bibinfo {author} {\bibfnamefont {Y.}~\bibnamefont {Zhang}}, \bibinfo
  {author} {\bibfnamefont {Y.}~\bibnamefont {Bengio}},\ and\ \bibinfo {author}
  {\bibfnamefont {S.}~\bibnamefont {Ravanbakhsh}},\ }\bibfield  {title}
  {\bibinfo {title} {Equivariance with learned canonicalization functions},\
  }in\ \href@noop {} {\emph {\bibinfo {booktitle} {International Conference on
  Machine Learning}}}\ (\bibinfo {organization} {PMLR},\ \bibinfo {year}
  {2023})\ pp.\ \bibinfo {pages} {15546--15566}\BibitemShut {NoStop}%
\bibitem [{\citenamefont {Qi}\ \emph {et~al.}(2017{\natexlab{b}})\citenamefont
  {Qi}, \citenamefont {Su}, \citenamefont {Mo},\ and\ \citenamefont
  {Guibas}}]{qi2017pointnet}%
  \BibitemOpen
  \bibfield  {author} {\bibinfo {author} {\bibfnamefont {C.~R.}\ \bibnamefont
  {Qi}}, \bibinfo {author} {\bibfnamefont {H.}~\bibnamefont {Su}}, \bibinfo
  {author} {\bibfnamefont {K.}~\bibnamefont {Mo}},\ and\ \bibinfo {author}
  {\bibfnamefont {L.~J.}\ \bibnamefont {Guibas}},\ }\bibfield  {title}
  {\bibinfo {title} {Pointnet: Deep learning on point sets for 3d
  classification and segmentation},\ }in\ \href@noop {} {\emph {\bibinfo
  {booktitle} {Proceedings of the IEEE conference on computer vision and
  pattern recognition}}}\ (\bibinfo {year} {2017})\ pp.\ \bibinfo {pages}
  {652--660}\BibitemShut {NoStop}%
\bibitem [{\citenamefont {Li}\ \emph {et~al.}(2021{\natexlab{b}})\citenamefont
  {Li}, \citenamefont {Fujiwara}, \citenamefont {Okura},\ and\ \citenamefont
  {Matsushita}}]{li2021closer}%
  \BibitemOpen
  \bibfield  {author} {\bibinfo {author} {\bibfnamefont {F.}~\bibnamefont
  {Li}}, \bibinfo {author} {\bibfnamefont {K.}~\bibnamefont {Fujiwara}},
  \bibinfo {author} {\bibfnamefont {F.}~\bibnamefont {Okura}},\ and\ \bibinfo
  {author} {\bibfnamefont {Y.}~\bibnamefont {Matsushita}},\ }\bibfield  {title}
  {\bibinfo {title} {A closer look at rotation-invariant deep point cloud
  analysis},\ }in\ \href@noop {} {\emph {\bibinfo {booktitle} {Proceedings of
  the IEEE/CVF International Conference on Computer Vision}}}\ (\bibinfo {year}
  {2021})\ pp.\ \bibinfo {pages} {16218--16227}\BibitemShut {NoStop}%
\bibitem [{\citenamefont {Baker}\ \emph {et~al.}(2024)\citenamefont {Baker},
  \citenamefont {Wang}, \citenamefont {de~Fernex},\ and\ \citenamefont
  {Wang}}]{baker2024explicit}%
  \BibitemOpen
  \bibfield  {author} {\bibinfo {author} {\bibfnamefont {J.}~\bibnamefont
  {Baker}}, \bibinfo {author} {\bibfnamefont {S.-H.}\ \bibnamefont {Wang}},
  \bibinfo {author} {\bibfnamefont {T.}~\bibnamefont {de~Fernex}},\ and\
  \bibinfo {author} {\bibfnamefont {B.}~\bibnamefont {Wang}},\ }\bibfield
  {title} {\bibinfo {title} {An explicit frame construction for normalizing
  {3D} point clouds},\ }in\ \href@noop {} {\emph {\bibinfo {booktitle}
  {Forty-first International Conference on Machine Learning}}}\ (\bibinfo
  {year} {2024})\BibitemShut {NoStop}%
\bibitem [{\citenamefont {Zhao}\ \emph {et~al.}(2020)\citenamefont {Zhao},
  \citenamefont {Birdal}, \citenamefont {Lenssen}, \citenamefont {Menegatti},
  \citenamefont {Guibas},\ and\ \citenamefont {Tombari}}]{zhao2020quaternion}%
  \BibitemOpen
  \bibfield  {author} {\bibinfo {author} {\bibfnamefont {Y.}~\bibnamefont
  {Zhao}}, \bibinfo {author} {\bibfnamefont {T.}~\bibnamefont {Birdal}},
  \bibinfo {author} {\bibfnamefont {J.~E.}\ \bibnamefont {Lenssen}}, \bibinfo
  {author} {\bibfnamefont {E.}~\bibnamefont {Menegatti}}, \bibinfo {author}
  {\bibfnamefont {L.}~\bibnamefont {Guibas}},\ and\ \bibinfo {author}
  {\bibfnamefont {F.}~\bibnamefont {Tombari}},\ }\bibfield  {title} {\bibinfo
  {title} {Quaternion equivariant capsule networks for {3D} point clouds},\
  }in\ \href@noop {} {\emph {\bibinfo {booktitle} {European conference on
  computer vision}}}\ (\bibinfo {organization} {Springer},\ \bibinfo {year}
  {2020})\ pp.\ \bibinfo {pages} {1--19}\BibitemShut {NoStop}%
\bibitem [{\citenamefont {Zhao}\ \emph {et~al.}(2022)\citenamefont {Zhao},
  \citenamefont {Yang}, \citenamefont {Xiong}, \citenamefont {Zhu},
  \citenamefont {Cao},\ and\ \citenamefont {Li}}]{zhao2022rotation}%
  \BibitemOpen
  \bibfield  {author} {\bibinfo {author} {\bibfnamefont {C.}~\bibnamefont
  {Zhao}}, \bibinfo {author} {\bibfnamefont {J.}~\bibnamefont {Yang}}, \bibinfo
  {author} {\bibfnamefont {X.}~\bibnamefont {Xiong}}, \bibinfo {author}
  {\bibfnamefont {A.}~\bibnamefont {Zhu}}, \bibinfo {author} {\bibfnamefont
  {Z.}~\bibnamefont {Cao}},\ and\ \bibinfo {author} {\bibfnamefont
  {X.}~\bibnamefont {Li}},\ }\bibfield  {title} {\bibinfo {title} {Rotation
  invariant point cloud analysis: Where local geometry meets global topology},\
  }\href@noop {} {\bibfield  {journal} {\bibinfo  {journal} {Pattern
  Recognition}\ }\textbf {\bibinfo {volume} {127}},\ \bibinfo {pages} {108626}
  (\bibinfo {year} {2022})}\BibitemShut {NoStop}%
\bibitem [{\citenamefont {Zhang}\ \emph {et~al.}(2020)\citenamefont {Zhang},
  \citenamefont {Hua}, \citenamefont {Chen}, \citenamefont {Tian},\ and\
  \citenamefont {Yeung}}]{zhang2020global}%
  \BibitemOpen
  \bibfield  {author} {\bibinfo {author} {\bibfnamefont {Z.}~\bibnamefont
  {Zhang}}, \bibinfo {author} {\bibfnamefont {B.-S.}\ \bibnamefont {Hua}},
  \bibinfo {author} {\bibfnamefont {W.}~\bibnamefont {Chen}}, \bibinfo {author}
  {\bibfnamefont {Y.}~\bibnamefont {Tian}},\ and\ \bibinfo {author}
  {\bibfnamefont {S.-K.}\ \bibnamefont {Yeung}},\ }\bibfield  {title} {\bibinfo
  {title} {Global context aware convolutions for {3D} point cloud
  understanding},\ }in\ \href@noop {} {\emph {\bibinfo {booktitle} {2020
  International Conference on 3D Vision}}}\ (\bibinfo {organization} {IEEE},\
  \bibinfo {year} {2020})\ pp.\ \bibinfo {pages} {210--219}\BibitemShut
  {NoStop}%
\bibitem [{\citenamefont {Wang}\ and\ \citenamefont
  {Zhang}(2022)}]{wang2022graph}%
  \BibitemOpen
  \bibfield  {author} {\bibinfo {author} {\bibfnamefont {X.}~\bibnamefont
  {Wang}}\ and\ \bibinfo {author} {\bibfnamefont {M.}~\bibnamefont {Zhang}},\
  }\bibfield  {title} {\bibinfo {title} {Graph neural network with local frame
  for molecular potential energy surface},\ }in\ \href@noop {} {\emph {\bibinfo
  {booktitle} {Proceedings of the First Learning on Graphs Conference}}}\
  (\bibinfo {organization} {PMLR},\ \bibinfo {year} {2022})\ pp.\ \bibinfo
  {pages} {19:1--19:30}\BibitemShut {NoStop}%
\bibitem [{\citenamefont {Luo}\ \emph {et~al.}(2022)\citenamefont {Luo},
  \citenamefont {Li}, \citenamefont {Guan}, \citenamefont {Su}, \citenamefont
  {Cheng}, \citenamefont {Peng},\ and\ \citenamefont
  {Ma}}]{luo2022equivariant}%
  \BibitemOpen
  \bibfield  {author} {\bibinfo {author} {\bibfnamefont {S.}~\bibnamefont
  {Luo}}, \bibinfo {author} {\bibfnamefont {J.}~\bibnamefont {Li}}, \bibinfo
  {author} {\bibfnamefont {J.}~\bibnamefont {Guan}}, \bibinfo {author}
  {\bibfnamefont {Y.}~\bibnamefont {Su}}, \bibinfo {author} {\bibfnamefont
  {C.}~\bibnamefont {Cheng}}, \bibinfo {author} {\bibfnamefont
  {J.}~\bibnamefont {Peng}},\ and\ \bibinfo {author} {\bibfnamefont
  {J.}~\bibnamefont {Ma}},\ }\bibfield  {title} {\bibinfo {title} {Equivariant
  point cloud analysis via learning orientations for message passing},\ }in\
  \href@noop {} {\emph {\bibinfo {booktitle} {Proceedings of the IEEE/CVF
  Conference on Computer Vision and Pattern Recognition}}}\ (\bibinfo {year}
  {2022})\ pp.\ \bibinfo {pages} {18932--18941}\BibitemShut {NoStop}%
\bibitem [{\citenamefont {Du}\ \emph {et~al.}(2022)\citenamefont {Du},
  \citenamefont {Zhang}, \citenamefont {Du}, \citenamefont {Meng},
  \citenamefont {Chen}, \citenamefont {Zheng}, \citenamefont {Shao},\ and\
  \citenamefont {Liu}}]{du2022se}%
  \BibitemOpen
  \bibfield  {author} {\bibinfo {author} {\bibfnamefont {W.}~\bibnamefont
  {Du}}, \bibinfo {author} {\bibfnamefont {H.}~\bibnamefont {Zhang}}, \bibinfo
  {author} {\bibfnamefont {Y.}~\bibnamefont {Du}}, \bibinfo {author}
  {\bibfnamefont {Q.}~\bibnamefont {Meng}}, \bibinfo {author} {\bibfnamefont
  {W.}~\bibnamefont {Chen}}, \bibinfo {author} {\bibfnamefont {N.}~\bibnamefont
  {Zheng}}, \bibinfo {author} {\bibfnamefont {B.}~\bibnamefont {Shao}},\ and\
  \bibinfo {author} {\bibfnamefont {T.-Y.}\ \bibnamefont {Liu}},\ }\bibfield
  {title} {\bibinfo {title} {Se (3) equivariant graph neural networks with
  complete local frames},\ }in\ \href@noop {} {\emph {\bibinfo {booktitle}
  {International Conference on Machine Learning}}}\ (\bibinfo {organization}
  {PMLR},\ \bibinfo {year} {2022})\ pp.\ \bibinfo {pages}
  {5583--5608}\BibitemShut {NoStop}%
\bibitem [{\citenamefont {Gerhartz}\ \emph {et~al.}(2025)\citenamefont
  {Gerhartz}, \citenamefont {Lippmann},\ and\ \citenamefont
  {Hamprecht}}]{gerhartz2025equivariance}%
  \BibitemOpen
  \bibfield  {author} {\bibinfo {author} {\bibfnamefont {G.}~\bibnamefont
  {Gerhartz}}, \bibinfo {author} {\bibfnamefont {P.}~\bibnamefont {Lippmann}},\
  and\ \bibinfo {author} {\bibfnamefont {F.~A.}\ \bibnamefont {Hamprecht}},\
  }\bibfield  {title} {\bibinfo {title} {Equivariance by local
  canonicalization: A matter of representation},\ }in\ \href
  {https://openreview.net/forum?id=0SifqnYaBe} {\emph {\bibinfo {booktitle}
  {NeurIPS 2025 Workshop on Symmetry and Geometry in Neural Representations}}}\
  (\bibinfo {year} {2025})\BibitemShut {NoStop}%
\bibitem [{\citenamefont {Spinner}\ \emph {et~al.}(2025)\citenamefont
  {Spinner}, \citenamefont {Favaro}, \citenamefont {Lippmann}, \citenamefont
  {Pitz}, \citenamefont {Gerhartz}, \citenamefont {Plehn},\ and\ \citenamefont
  {Hamprecht}}]{spinner2025lloca}%
  \BibitemOpen
  \bibfield  {author} {\bibinfo {author} {\bibfnamefont {J.}~\bibnamefont
  {Spinner}}, \bibinfo {author} {\bibfnamefont {L.}~\bibnamefont {Favaro}},
  \bibinfo {author} {\bibfnamefont {P.}~\bibnamefont {Lippmann}}, \bibinfo
  {author} {\bibfnamefont {S.}~\bibnamefont {Pitz}}, \bibinfo {author}
  {\bibfnamefont {G.}~\bibnamefont {Gerhartz}}, \bibinfo {author}
  {\bibfnamefont {T.}~\bibnamefont {Plehn}},\ and\ \bibinfo {author}
  {\bibfnamefont {F.~A.}\ \bibnamefont {Hamprecht}},\ }\bibfield  {title}
  {\bibinfo {title} {Lorentz local canonicalization: How to make any network
  lorentz-equivariant},\ }in\ \href
  {https://openreview.net/forum?id=2xeYuWd7WZ} {\emph {\bibinfo {booktitle}
  {The Thirty-ninth Annual Conference on Neural Information Processing
  Systems}}}\ (\bibinfo {year} {2025})\BibitemShut {NoStop}%
\bibitem [{\citenamefont {Dym}\ \emph {et~al.}(2024)\citenamefont {Dym},
  \citenamefont {Lawrence},\ and\ \citenamefont {Siegel}}]{dym2024equivariant}%
  \BibitemOpen
  \bibfield  {author} {\bibinfo {author} {\bibfnamefont {N.}~\bibnamefont
  {Dym}}, \bibinfo {author} {\bibfnamefont {H.}~\bibnamefont {Lawrence}},\ and\
  \bibinfo {author} {\bibfnamefont {J.~W.}\ \bibnamefont {Siegel}},\ }\bibfield
   {title} {\bibinfo {title} {Equivariant frames and the impossibility of
  continuous canonicalization},\ }in\ \href@noop {} {\emph {\bibinfo
  {booktitle} {International Conference on Machine Learning}}}\ (\bibinfo
  {organization} {PMLR},\ \bibinfo {year} {2024})\ pp.\ \bibinfo {pages}
  {12228--12267}\BibitemShut {NoStop}%
\bibitem [{\citenamefont {Cohen}\ \emph
  {et~al.}(2019{\natexlab{b}})\citenamefont {Cohen}, \citenamefont {Weiler},
  \citenamefont {Kicanaoglu},\ and\ \citenamefont {Welling}}]{cohen2019gauge}%
  \BibitemOpen
  \bibfield  {author} {\bibinfo {author} {\bibfnamefont {T.}~\bibnamefont
  {Cohen}}, \bibinfo {author} {\bibfnamefont {M.}~\bibnamefont {Weiler}},
  \bibinfo {author} {\bibfnamefont {B.}~\bibnamefont {Kicanaoglu}},\ and\
  \bibinfo {author} {\bibfnamefont {M.}~\bibnamefont {Welling}},\ }\bibfield
  {title} {\bibinfo {title} {Gauge equivariant convolutional networks and the
  icosahedral cnn},\ }in\ \href@noop {} {\emph {\bibinfo {booktitle}
  {International conference on Machine learning}}}\ (\bibinfo {organization}
  {PMLR},\ \bibinfo {year} {2019})\ pp.\ \bibinfo {pages}
  {1321--1330}\BibitemShut {NoStop}%
\bibitem [{\citenamefont {De~Haan}\ \emph {et~al.}(2020)\citenamefont
  {De~Haan}, \citenamefont {Weiler}, \citenamefont {Cohen},\ and\ \citenamefont
  {Welling}}]{de2020gauge}%
  \BibitemOpen
  \bibfield  {author} {\bibinfo {author} {\bibfnamefont {P.}~\bibnamefont
  {De~Haan}}, \bibinfo {author} {\bibfnamefont {M.}~\bibnamefont {Weiler}},
  \bibinfo {author} {\bibfnamefont {T.}~\bibnamefont {Cohen}},\ and\ \bibinfo
  {author} {\bibfnamefont {M.}~\bibnamefont {Welling}},\ }\bibfield  {title}
  {\bibinfo {title} {Gauge equivariant mesh cnns: Anisotropic convolutions on
  geometric graphs},\ }\href@noop {} {\bibfield  {journal} {\bibinfo  {journal}
  {arXiv preprint arXiv:2003.05425}\ } (\bibinfo {year} {2020})}\BibitemShut
  {NoStop}%
\bibitem [{\citenamefont {Wang}\ \emph {et~al.}(2023)\citenamefont {Wang},
  \citenamefont {Elhag}, \citenamefont {Jaitly}, \citenamefont {Susskind},\
  and\ \citenamefont {Bautista}}]{wang2023swallowing}%
  \BibitemOpen
  \bibfield  {author} {\bibinfo {author} {\bibfnamefont {Y.}~\bibnamefont
  {Wang}}, \bibinfo {author} {\bibfnamefont {A.~A.}\ \bibnamefont {Elhag}},
  \bibinfo {author} {\bibfnamefont {N.}~\bibnamefont {Jaitly}}, \bibinfo
  {author} {\bibfnamefont {J.~M.}\ \bibnamefont {Susskind}},\ and\ \bibinfo
  {author} {\bibfnamefont {M.~A.}\ \bibnamefont {Bautista}},\ }\bibfield
  {title} {\bibinfo {title} {Swallowing the bitter pill: Simplified scalable
  conformer generation},\ }\href@noop {} {\bibfield  {journal} {\bibinfo
  {journal} {arXiv preprint arXiv:2311.17932}\ } (\bibinfo {year}
  {2023})}\BibitemShut {NoStop}%
\bibitem [{\citenamefont {Pertigkiozoglou}\ \emph {et~al.}(2024)\citenamefont
  {Pertigkiozoglou}, \citenamefont {Chatzipantazis}, \citenamefont {Trivedi},\
  and\ \citenamefont {Daniilidis}}]{pertigkiozoglou2024improving}%
  \BibitemOpen
  \bibfield  {author} {\bibinfo {author} {\bibfnamefont {S.}~\bibnamefont
  {Pertigkiozoglou}}, \bibinfo {author} {\bibfnamefont {E.}~\bibnamefont
  {Chatzipantazis}}, \bibinfo {author} {\bibfnamefont {S.}~\bibnamefont
  {Trivedi}},\ and\ \bibinfo {author} {\bibfnamefont {K.}~\bibnamefont
  {Daniilidis}},\ }\bibfield  {title} {\bibinfo {title} {Improving equivariant
  model training via constraint relaxation},\ }\href@noop {} {\bibfield
  {journal} {\bibinfo  {journal} {Advances in Neural Information Processing
  Systems}\ }\textbf {\bibinfo {volume} {37}},\ \bibinfo {pages} {83497}
  (\bibinfo {year} {2024})}\BibitemShut {NoStop}%
\bibitem [{\citenamefont {Elhag}\ \emph {et~al.}(2024)\citenamefont {Elhag},
  \citenamefont {Rusch}, \citenamefont {Di~Giovanni},\ and\ \citenamefont
  {Bronstein}}]{elhag2024relaxed}%
  \BibitemOpen
  \bibfield  {author} {\bibinfo {author} {\bibfnamefont {A.~A.}\ \bibnamefont
  {Elhag}}, \bibinfo {author} {\bibfnamefont {T.~K.}\ \bibnamefont {Rusch}},
  \bibinfo {author} {\bibfnamefont {F.}~\bibnamefont {Di~Giovanni}},\ and\
  \bibinfo {author} {\bibfnamefont {M.}~\bibnamefont {Bronstein}},\ }\bibfield
  {title} {\bibinfo {title} {Relaxed equivariance via multitask learning},\
  }\href@noop {} {\bibfield  {journal} {\bibinfo  {journal} {arXiv preprint
  arXiv:2410.17878}\ } (\bibinfo {year} {2024})}\BibitemShut {NoStop}%
\bibitem [{\citenamefont {Qu}\ and\ \citenamefont
  {Krishnapriyan}(2024)}]{qu2024importance}%
  \BibitemOpen
  \bibfield  {author} {\bibinfo {author} {\bibfnamefont {E.}~\bibnamefont
  {Qu}}\ and\ \bibinfo {author} {\bibfnamefont {A.}~\bibnamefont
  {Krishnapriyan}},\ }\bibfield  {title} {\bibinfo {title} {The importance of
  being scalable: Improving the speed and accuracy of neural network
  interatomic potentials across chemical domains},\ }\href@noop {} {\bibfield
  {journal} {\bibinfo  {journal} {Advances in Neural Information Processing
  Systems}\ }\textbf {\bibinfo {volume} {37}},\ \bibinfo {pages} {139030}
  (\bibinfo {year} {2024})}\BibitemShut {NoStop}%
\bibitem [{\citenamefont {Langer}\ \emph {et~al.}(2024)\citenamefont {Langer},
  \citenamefont {Pozdnyakov},\ and\ \citenamefont
  {Ceriotti}}]{langer2024probing}%
  \BibitemOpen
  \bibfield  {author} {\bibinfo {author} {\bibfnamefont {M.~F.}\ \bibnamefont
  {Langer}}, \bibinfo {author} {\bibfnamefont {S.~N.}\ \bibnamefont
  {Pozdnyakov}},\ and\ \bibinfo {author} {\bibfnamefont {M.}~\bibnamefont
  {Ceriotti}},\ }\bibfield  {title} {\bibinfo {title} {Probing the effects of
  broken symmetries in machine learning},\ }\href@noop {} {\bibfield  {journal}
  {\bibinfo  {journal} {arXiv preprint arXiv:2406.17747}\ } (\bibinfo {year}
  {2024})}\BibitemShut {NoStop}%
\bibitem [{\citenamefont {Eissler}\ \emph {et~al.}(2025)\citenamefont
  {Eissler}, \citenamefont {Korjakow}, \citenamefont {Ganscha}, \citenamefont
  {Unke}, \citenamefont {M{\~A}{\v{z}}ller},\ and\ \citenamefont
  {Gugler}}]{eissler2025simple}%
  \BibitemOpen
  \bibfield  {author} {\bibinfo {author} {\bibfnamefont {M.}~\bibnamefont
  {Eissler}}, \bibinfo {author} {\bibfnamefont {T.}~\bibnamefont {Korjakow}},
  \bibinfo {author} {\bibfnamefont {S.}~\bibnamefont {Ganscha}}, \bibinfo
  {author} {\bibfnamefont {O.~T.}\ \bibnamefont {Unke}}, \bibinfo {author}
  {\bibfnamefont {K.-R.}\ \bibnamefont {M{\~A}{\v{z}}ller}},\ and\ \bibinfo
  {author} {\bibfnamefont {S.}~\bibnamefont {Gugler}},\ }\bibfield  {title}
  {\bibinfo {title} {How simple can you go? an off-the-shelf transformer
  approach to molecular dynamics},\ }\href@noop {} {\bibfield  {journal}
  {\bibinfo  {journal} {arXiv preprint arXiv:2503.01431}\ } (\bibinfo {year}
  {2025})}\BibitemShut {NoStop}%
\bibitem [{\citenamefont {Manolache}\ \emph {et~al.}(2025)\citenamefont
  {Manolache}, \citenamefont {Chamon},\ and\ \citenamefont
  {Niepert}}]{manolache2025learning}%
  \BibitemOpen
  \bibfield  {author} {\bibinfo {author} {\bibfnamefont {A.}~\bibnamefont
  {Manolache}}, \bibinfo {author} {\bibfnamefont {L.~F.~O.}\ \bibnamefont
  {Chamon}},\ and\ \bibinfo {author} {\bibfnamefont {M.}~\bibnamefont
  {Niepert}},\ }\bibfield  {title} {\bibinfo {title} {Learning (approximately)
  equivariant networks via constrained optimization},\ }in\ \href
  {https://openreview.net/forum?id=NM4emKloy6} {\emph {\bibinfo {booktitle}
  {The Thirty-ninth Annual Conference on Neural Information Processing
  Systems}}}\ (\bibinfo {year} {2025})\BibitemShut {NoStop}%
\bibitem [{\citenamefont {Berndt}\ and\ \citenamefont
  {St{\"u}hmer}(2026)}]{berndt2026approximate}%
  \BibitemOpen
  \bibfield  {author} {\bibinfo {author} {\bibfnamefont {T.}~\bibnamefont
  {Berndt}}\ and\ \bibinfo {author} {\bibfnamefont {J.}~\bibnamefont
  {St{\"u}hmer}},\ }\bibfield  {title} {\bibinfo {title} {Approximate
  equivariance via projection-based regularisation},\ }\href@noop {} {\bibfield
   {journal} {\bibinfo  {journal} {arXiv preprint arXiv:2601.05028}\ }
  (\bibinfo {year} {2026})}\BibitemShut {NoStop}%
\bibitem [{\citenamefont {Bigi}\ \emph {et~al.}(2026)\citenamefont {Bigi},
  \citenamefont {Pegolo}, \citenamefont {Mazitov},\ and\ \citenamefont
  {Ceriotti}}]{bigi2026pushing}%
  \BibitemOpen
  \bibfield  {author} {\bibinfo {author} {\bibfnamefont {F.}~\bibnamefont
  {Bigi}}, \bibinfo {author} {\bibfnamefont {P.}~\bibnamefont {Pegolo}},
  \bibinfo {author} {\bibfnamefont {A.}~\bibnamefont {Mazitov}},\ and\ \bibinfo
  {author} {\bibfnamefont {M.}~\bibnamefont {Ceriotti}},\ }\bibfield  {title}
  {\bibinfo {title} {Pushing the limits of unconstrained machine-learned
  interatomic potentials},\ }\href@noop {} {\bibfield  {journal} {\bibinfo
  {journal} {arXiv preprint arXiv:2601.16195}\ } (\bibinfo {year}
  {2026})}\BibitemShut {NoStop}%
\bibitem [{\citenamefont {Sutton}(2019)}]{sutton2019bitter}%
  \BibitemOpen
  \bibfield  {author} {\bibinfo {author} {\bibfnamefont {R.}~\bibnamefont
  {Sutton}},\ }\bibfield  {title} {\bibinfo {title} {The bitter lesson},\
  }\href@noop {} {\bibfield  {journal} {\bibinfo  {journal} {Incomplete Ideas
  (blog)}\ }\textbf {\bibinfo {volume} {13}},\ \bibinfo {pages} {38} (\bibinfo
  {year} {2019})}\BibitemShut {NoStop}%
\bibitem [{\citenamefont {Jacobs}\ \emph {et~al.}(1991)\citenamefont {Jacobs},
  \citenamefont {Jordan}, \citenamefont {Nowlan},\ and\ \citenamefont
  {Hinton}}]{jacobs1991adaptive}%
  \BibitemOpen
  \bibfield  {author} {\bibinfo {author} {\bibfnamefont {R.~A.}\ \bibnamefont
  {Jacobs}}, \bibinfo {author} {\bibfnamefont {M.~I.}\ \bibnamefont {Jordan}},
  \bibinfo {author} {\bibfnamefont {S.~J.}\ \bibnamefont {Nowlan}},\ and\
  \bibinfo {author} {\bibfnamefont {G.~E.}\ \bibnamefont {Hinton}},\ }\bibfield
   {title} {\bibinfo {title} {Adaptive mixtures of local experts},\ }\href@noop
  {} {\bibfield  {journal} {\bibinfo  {journal} {Neural computation}\ }\textbf
  {\bibinfo {volume} {3}},\ \bibinfo {pages} {79} (\bibinfo {year}
  {1991})}\BibitemShut {NoStop}%
\bibitem [{\citenamefont {Fedus}\ \emph {et~al.}(2022)\citenamefont {Fedus},
  \citenamefont {Zoph},\ and\ \citenamefont {Shazeer}}]{fedus2022switch}%
  \BibitemOpen
  \bibfield  {author} {\bibinfo {author} {\bibfnamefont {W.}~\bibnamefont
  {Fedus}}, \bibinfo {author} {\bibfnamefont {B.}~\bibnamefont {Zoph}},\ and\
  \bibinfo {author} {\bibfnamefont {N.}~\bibnamefont {Shazeer}},\ }\bibfield
  {title} {\bibinfo {title} {Switch transformers: Scaling to trillion parameter
  models with simple and efficient sparsity},\ }\href@noop {} {\bibfield
  {journal} {\bibinfo  {journal} {Journal of Machine Learning Research}\
  }\textbf {\bibinfo {volume} {23}},\ \bibinfo {pages} {1} (\bibinfo {year}
  {2022})}\BibitemShut {NoStop}%
\bibitem [{\citenamefont {Brehmer}\ \emph {et~al.}(2024)\citenamefont
  {Brehmer}, \citenamefont {Behrends}, \citenamefont {de~Haan},\ and\
  \citenamefont {Cohen}}]{brehmer2024does}%
  \BibitemOpen
  \bibfield  {author} {\bibinfo {author} {\bibfnamefont {J.}~\bibnamefont
  {Brehmer}}, \bibinfo {author} {\bibfnamefont {S.}~\bibnamefont {Behrends}},
  \bibinfo {author} {\bibfnamefont {P.}~\bibnamefont {de~Haan}},\ and\ \bibinfo
  {author} {\bibfnamefont {T.}~\bibnamefont {Cohen}},\ }\bibfield  {title}
  {\bibinfo {title} {Does equivariance matter at scale?},\ }\href@noop {}
  {\bibfield  {journal} {\bibinfo  {journal} {arXiv preprint arXiv:2410.23179}\
  } (\bibinfo {year} {2024})}\BibitemShut {NoStop}%
\end{thebibliography}%

\end{document}